\documentclass[12pt,a4paper]{report}
\usepackage{titlesec}
\usepackage{amsmath}
\usepackage{graphicx}
\usepackage{amsfonts}
\usepackage{footnote}
\usepackage{booktabs} 

\usepackage{amssymb,amsmath,amsfonts}
\usepackage{array}
\usepackage{url}
\usepackage{multirow}
\usepackage{tabularx,ragged2e}

\DeclareMathOperator*{\argmax}{arg\,max}
\usepackage{longtable}

\usepackage{Packages/epsfig}
\usepackage{Packages/fancyheadings}
\usepackage{Packages/setspace}

\usepackage{color}
\usepackage{amssymb}
\usepackage{graphicx}
\usepackage{subfigure}
\usepackage{geometry}
\usepackage[lined,ruled,vlined,linesnumbered]{algorithm2e}
\usepackage{bmpsize}
\usepackage{capt-of}

\renewcommand{\theequation}{\mbox{\rm \thechapter.\arabic{equation}}}

\newcounter{proposition}

\newcounter{theorem}

\newcounter{assumption}

\newcounter{lemma}

\newcounter{corollary}

\newcounter{definition}

\renewcommand{\thefigure}{\thechapter.\arabic{figure}}

\def\LEQ.#1.#2.#3{#1\!\leqslant\!#2\!\leqslant\!#3}
\def\GEQ.#1.#2.#3{#1\!\geqslant\!#2\!\geqslant\!#3}
\def\emptyset{\varnothing}
\renewcommand{\theenumi}{\rm \roman{enumi}}

\renewcommand{\theenumii}{\rm \alph{enumii}}

\begin{document}

\title{\sc
\resizebox*{0.7\textwidth}{!}{\includegraphics{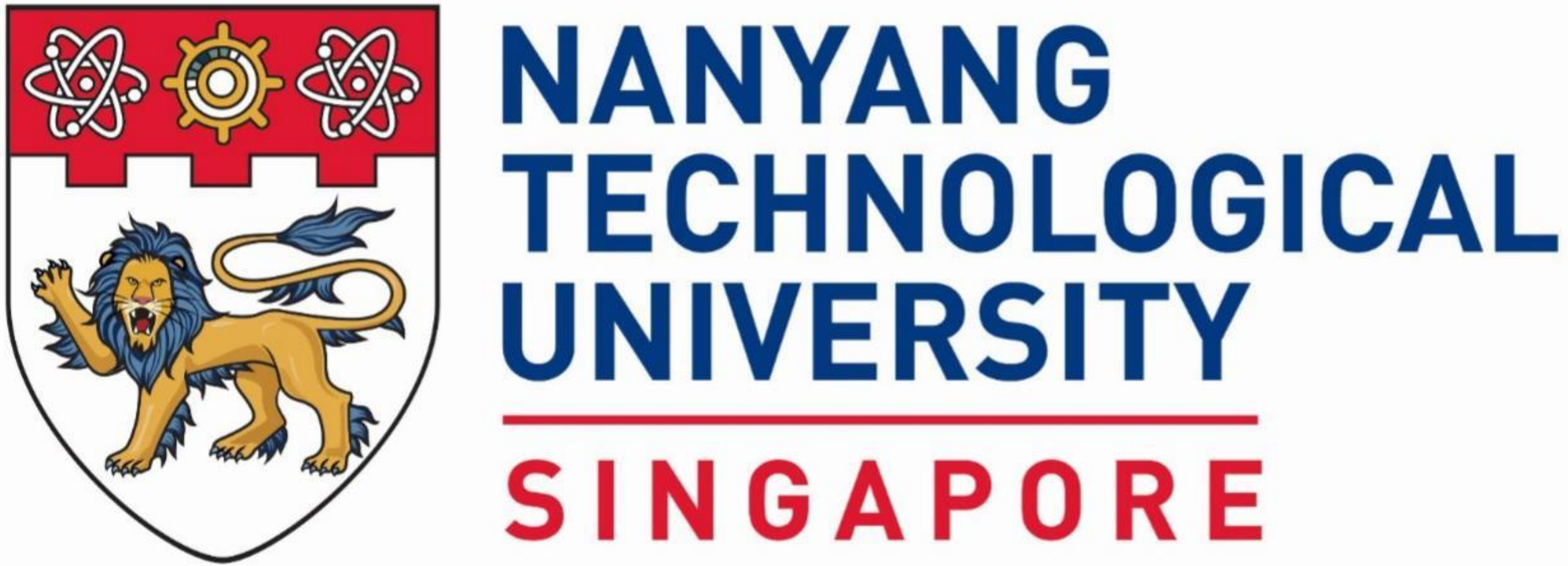}}\\[1em]
\vspace{0.2in}{\huge\bf Concept-Based Embeddings\\for Natural Language Processing}\\[1em]}

\author{
Ph.D Thesis\\[1em]
By\\[1em]
{\rm\bf Yukun Ma}\\[1em]
{\rm\bf Supervisor: Prof. Erik Cambria}\\[4em]
School of Computer Science and Engineering\\[1em]
A thesis submitted to the Nanyang Technological University \\in fulfillment of the requirement for the degree of
Doctor of Philosophy}

\date{January, 2018}
\maketitle
\thispagestyle{empty}        


%

\setlength{\baselineskip}{20pt plus 1pt minus 1pt} 
\pagenumbering{roman}    

\setcounter{page}{0}

\chapter* {Acknowledgments}
\addcontentsline{toc}{section}{\numberline{}\hspace{-.35in}{\bf Acknowledgments}} 

First and foremost, I am deeply grateful to my supervisor Dr. Erik Cambria who has been extremely supportive during the last two years of my Ph.D. Without his guidance and assistance, it would be impossible for me to come to this final stage. 

I also would like to extend my thanks to Dr. Jungjae Kim, my former supervisor, for bringing to me this program and helping me to set up the initial research profile.  

I would like to thank Rolls-Royce and National Research Funding for providing  financial support for my research as well as my living in Singapore. I also would like to thank San Linn, the manager of our research project, for his effective coordination and kind support throughout the whole process.

Special thanks go to Dr. Benjamin Bigot, Dr. Tahir Khan, who are my colleagues and friends in the Rolls-Royce Corp Lab, for continuously encouraging and inspiring me with great thinkings.  

I also have greatly benefited from the friendship of Dr. Luu Ann Tuan, Dr. Tran Ha Nguyen,  Sa Gao, Haiyun Peng, Wen Song, Guanghao Zhang, and Zehong Hu. Thank Patrick and Dr. Chieh Lu who are great friends of us and had stayed by my side during my toughest days in the small and hopeless student hostel. Thanks to all of them, my Ph.D. could end up as a pleasant and rewarding journey.

I would like to give extra credit to my parents, Jie Yang and Jianguo Ma, for that all the achievements in this thesis are results of their wise advice on selecting my undergraduate program. I owe enormous gratitude to my parents-in-law, Dongrong Tong and Xiaoying Xu, who have always supported our decisions and life both financially and emotionally. Finally, as the most valuable asset of my life and my wife, Qiying Tong definitely deserves my lifelong gratitude and love for her bearing with my endless stupidity and dullness since our first day. This thesis is fully dedicated to them.  


%
%
\setlength{\baselineskip}{20pt plus 1pt minus 1pt} 

\setcounter{secnumdepth}{3} 
\setcounter{tocdepth}{2}  

\tableofcontents      

\newpage
\addcontentsline{toc}{section}{\numberline{}\hspace{-.35in}{\bf List of Figures}}
\listoffigures       

\newpage
\addcontentsline{toc}{section}{\numberline{}\hspace{-.35in}{\bf List of Tables}}
\listoftables       

\def\listofnotations {\input symbols.tex \clearpage}
\def\addnotation #1: #2#3 {\parbox{6in}{$#1$ \hfill \parbox{5in}{#2 \dotfill \pageref{#3}}\\}}
\def\newnot#1{\label{#1}}

\newcolumntype{L}[1]{>{\raggedright\let\newline\\\arraybackslash\hspace{0pt}}m{#1}}
\newcolumntype{C}[1]{>{\centering\let\newline\\\arraybackslash\hspace{0pt}}m{#1}}
\newcolumntype{R}[1]{>{\raggedleft\let\newline\\\arraybackslash\hspace{0pt}}m{#1}}

\newenvironment{megaalgorithm}[1][htb]
 {\renewcommand{\algorithmcfname}{Algorithm}
  \begin{algorithm}[#1]%
 }{\end{algorithm}}

\addtolength{\headheight}{3pt} \pagestyle{fancy} \setlength{\headrulewidth}{0.1pt}
\renewcommand{\chaptermark}[1]{\markboth{\chaptername\ \thechapter. #1}{}}
\lhead{\fancyplain{}{\bfseries\footnotesize\sc\leftmark}} \rhead{} \addtolength{\headsep}{-0.1in}
\cfoot{\fancyplain{}{\bfseries\rm\thepage}}

\newpage
\pagenumbering {arabic}   

\setlength{\baselineskip}{24pt plus 1pt minus 1pt} 

\setlength{\belowcaptionskip}{5pt} 
\renewcommand\arraystretch{1} 

\chapter*{List of Abbreviations}
\pagenumbering{gobble}

\begin{longtable}{p{2.7cm}p{14.5cm}}

AI & Artificial Intelligence \\
NLP & Natural Language Processing \\
ABSA & Aspect-based Sentiment Analysis\\
ASR & Automatic Speech Recognition\\
DNN & Deep Neural Network\\
RNN & Recurrent Neural Network\\
NER & Named Entity Recognition\\
FNET& Fine-grained Named Entity Typing\\
RBM& Restricted Boltzman Machine \\
LSTM & Long Short-Term Memory \\
Bi-LSTM&Bidirectional LSTM\\
TDLSTM& Target Dependent LSTM\\
TCLSTM& Target Connection LSTM\\
TSA & Targeted Sentiment Analysis\\
CRF & Conditional Random Field\\
PMI & Pointwise Mutual Information\\
\end{longtable}

\chapter* {Abstract} 

Concepts are critical semantics capturing the high-level knowledge of human language. As a way to go beyond the word-level analysis, representing and leveraging the concept-level information is an important add-on to existing natural language processing (NLP) systems. More specifically, the concepts are critical for understanding opinions of people. For example, people express their opinion towards particular entities such as products or sentiment aspects in online reviews, where these entities are mentions of concepts rather than just words. 

As compared with words, the mentions of abstract concepts may be compounded phrases (either consecutive or non-consecutive) that are likely to form a large vocabulary. Furthermore, there might be semantic properties (e.g., relations or attributes) attached to the concepts, which increases the dimensionality of concepts. In short, using concepts is faced with the curse of dimensionality. On the other hand, information from only a single level does not suffice for a thorough understanding of human language, and meaningful representation is required at any point to encode the correlation and dependency between abstract concepts and words. In this thesis, we thus focus on effectively leveraging and integrating information from concept-level as well as word-level via projecting concepts and words into a lower dimensional space while retaining most critical semantics. In a broad context of opinion understanding system, we investigate the use of the fused embedding for several core NLP tasks: named entity detection and classification, automatic speech recognition reranking, and targeted sentiment analysis. 

We first propose a novel method to inject the entity-based information into a word embedding space. The word embeddings are learned from a set of named entity features instead of merely contextual words. We demonstrate that the new word embedding is a better feature representation for detecting and classifying named entities from the stream of telephone conversations. 

Apart from learning input feature embeddings, we then explore encoding the entity types (i.e., concept categories) in a label embedding space. Our label embeddings mainly leverage two types of information: label hierarchy and label prototype. Since our label embedding is computed prior to the training process, it has exactly the same computation complexity at run-time.  We evaluate the resulting label embeddings on multiple large-scale datasets built for the task of fine-grained named entity typing. As compared with the state-of-the-art methods, our label embedding method can achieve superior performance.

Next, we demonstrate that a binary embedding of the named entities can help reranking the speech-to-text hypothesis. Named entities are encoded using a Restricted Boltzmann Machine (RBM) and used as a prior knowledge in the discriminative reranking model. We also extend the training of RBM to work with speech recognition hypothesis. 

Finally, we investigate the problem of using embeddings of commonsense concepts for the task of targeted sentiment analysis. The task is also entity-centered. Namely, given a targeted entity in a sentence, the task is to resolve the correct aspects categories and corresponding sentiment polarity of the target. We propose a new computation structure of Long Short-Term Memory (LSTM) that can more effectively incorporate the embeddings of commonsense knowledge. 

In summary, this thesis proposes novel solutions of representing and leveraging concept-level and word-level information in a series of NLP tasks that are key to understanding the opinion of people.

\addcontentsline{toc}{section}{\numberline{}\hspace{-0.35in}{\bf Abstract}} 

\chapter{Introduction}
\label{tag:chap1}
\pagenumbering{arabic}

The understanding of human language involves processing information beyond the words. A complete process should be capable of handling information from various levels such as concepts or thoughts. Concepts, as one of the most basic units of language, refer to aspects of the world that are used by people in their language. Concepts are constituents including mostly phrases such as noun phrases, verb phrases, adjective phrases and so on and so forth. In addition, the definition of concepts is broad enough to include mentions of named entities which can be deemed as concepts belonging to certain categories, e.g., location, organization, or person. 

Recently, a variety of knowledge resources such as ConceptNet have been built from the huge unstructured information thanks to the rapid growth of online communities and social network. These knowledge resources t are based on concepts rather than single words and are in the form of either a directed or undirected graph with concepts being the nodes and their relations being the edges. Additional attributes or properties might also be incorporated in order to better represent the semantics of concepts. These concept-based resources might benefit many NLP systems via providing accesses to external knowledge.

On the other hand, concepts are playing a core role in understanding the opinions of people. For example, people tend to express their sentiment towards a particular targeted entity (e.g., purchased products) in online reviews. By our definition of concepts, these entities are actually mentions of concepts. Therefore, the opinion understanding system would benefit from the use of concept-level information. However, due to the complexity of natural language, a complete analysis of people's opinions requires performing a series of subtasks such as concept extraction (or named entity recognition) and sentiment analysis, and it is hard to train a single end-to-end trained model that can perform well on all the subtasks of an opinion understanding system. Instead, it typically needs to build a pipeline composed of a series of natural language processing modules. In this thesis, we choose to extend the use of concepts to three important modules in the whole context: named entity recognition, targeted sentiment analysis and ASR hypotheses reranking. These tasks are core components of a multimodal opinion understanding system which is capable of processing not only text data but also audios. One of the natural obstacle hindering concepts from being employed in large scale is that concepts are generally of high dimensionality. Hence, our solutions focus on generating and leveraging concept-based embeddings for solving concept-centered NLP problems.

\section{Preliminaries}
We first introduce the definition of concepts as well as two knowledge resources that have been widely adopted for NLP tasks. We then give a brief introduction of concept-centered NLP tasks that we are interested in addressing in this thesis: named entity recognition, fine-grained named entity typing, targeted sentiment analysis, and ASR hypotheses reranking. 

\subsection{Concepts}
\begin{itemize}
\item \textbf{Concepts} are used by people to describe the objects or actions perceived. Concepts might have different forms in different contexts. Admittedly, the surface form of concepts are composed of single or multiple words. For example, the concept \textbf{small\_room} is composed of two words \textbf{small} and \textbf{room}. Inflections of concepts are prevalent in natural language. Namely, the same objects or action might be referred to as different concepts due to many reasons including cultural differences, the existence of synonyms and acronyms, and so on. We could deem a concept as an abstract term having a canonical form as well as several surface forms or mentions. For example, the concept of which the surface is ``New York'' might have mentions like ``New York city'' or ``big apple".

\item Named entities~\cite{muc7,conll:2003} can also be deemed as concepts.  In the very first challenge of named entity recognition~\cite{muc7}, named entities are defined as noun phrases in its surface form and refer to objects belonging to categories such as person, location, and organization. Although different challenges such as CoNLL 2003 shared tasks have extended the definition to involve other concepts, a standard definition of named entities has been constrained to consider only flat categories (i.e., non-hierarchical). Notably,  a recent study on fine-grained named entities has further extended the concern with named entities to involve a hierarchical structure where IsA relation between concepts is taken into consideration.
\end{itemize}

\subsection{Concept-based Knowledge Resources}

Recently, knowledge bases such as ConceptNet~\cite{concept:2013} have linked concepts to more structural semantic information, e.g., relations and attributes. These knowledge bases are built automatically by employing rule-based and machine learning techniques to extract facts from Web texts. Therefore, the coverage of concepts could keep growing over the time. 

\subsubsection{ConceptNet}
ConceptNet is automatically built as a combination and filtering of many existing and ever-growing multilingual knowledge resources such as Open Mind. ConceptNet keeps updating itself according to the source knowledge bases. One notable feature of ConceptNet is that it contains a large number of relational knowledge. Namely, we could be able to know the relations between concepts. For example, from ConceptNet, it can be known that \textbf{the white house} (argument 1) is \textbf{located} (relation) in \textbf{america} (argument 2) and \textbf{a type of} (relation) \textbf{building} (argument 3).

\subsubsection{SenticNet}

Same as ConceptNet, SenticNet~\cite{camnt5} merges the information in many existing knowledge resources. At the core of SenticNet is AffecNet where each concept is attached to a set of affective properties~\cite{camsen}.

\subsection{Concept-centered NLP for Opinion Understanding}

\begin{figure}
\centering
\includegraphics[width=\textwidth]{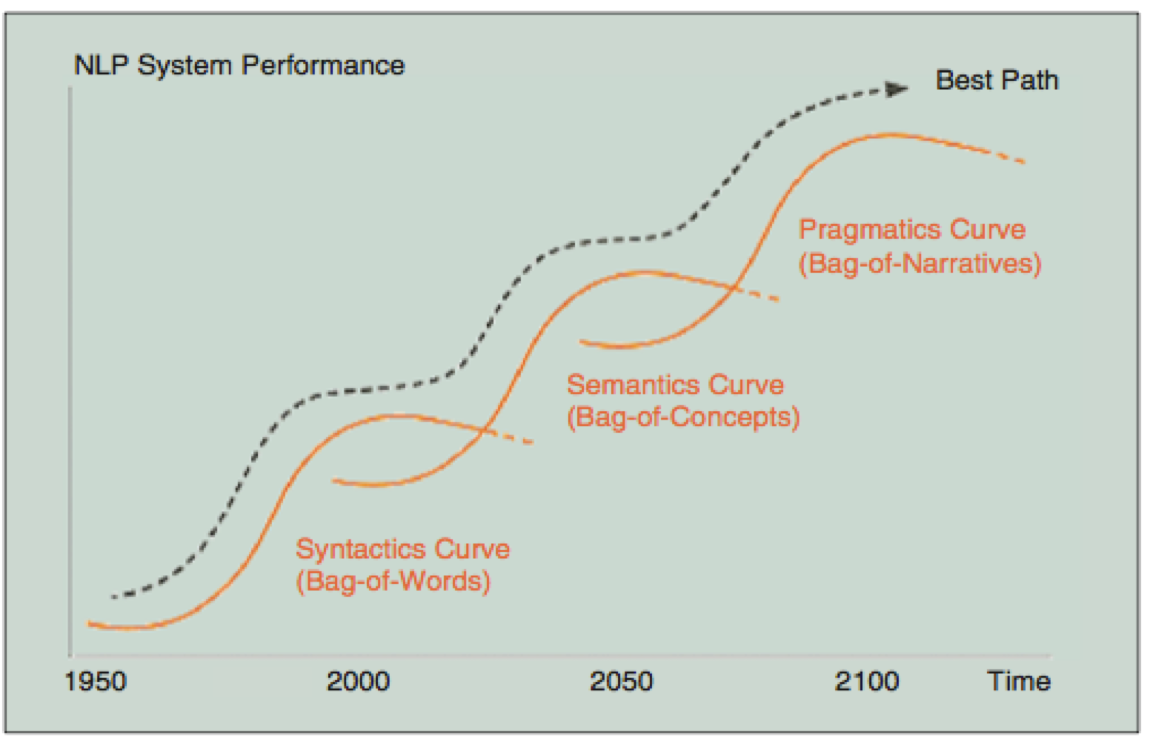}
\caption{NLP curve for the evolution of NLP techniques by Cambria et al.~\cite{cambria2014jumping} }
\label{fig:nlpcurv}
\end{figure}

The evolution of NLP techniques clears shows a trend moving from word-level to concept-level (as illustrated in Fig.~\ref{fig:nlpcurv}). Merely relying on word-level analysis of natural language is limited in the power of expressiveness since the natural language is composed of different levels of abstraction. It is, therefore, advantageous to use concepts for NLP tasks for that it allows for access to the rich semantic features of particular concepts. For example, collecting information about battery life and price is helpful when comparing the two products ``iPhone X'' and ``Samsung S8''. 

The concept-level analysis has been adopted for NLP tasks such as sentiment analysis and named entity recognition. However, concepts are very sparse due to the overwhelming number of unique concepts as well as concept inflections and features. A more compressed and effective representation is required to encode the concepts. On the other hand, a dense and lower dimensional representation also facilitate the exploration of deep learning models that have recently gained great successes in many NLP tasks~\cite{yourec}. With the access to knowledge bases, we are able to augment NLP solutions with concept-level semantics. 

\begin{figure}
\centering
\includegraphics[width=10cm]{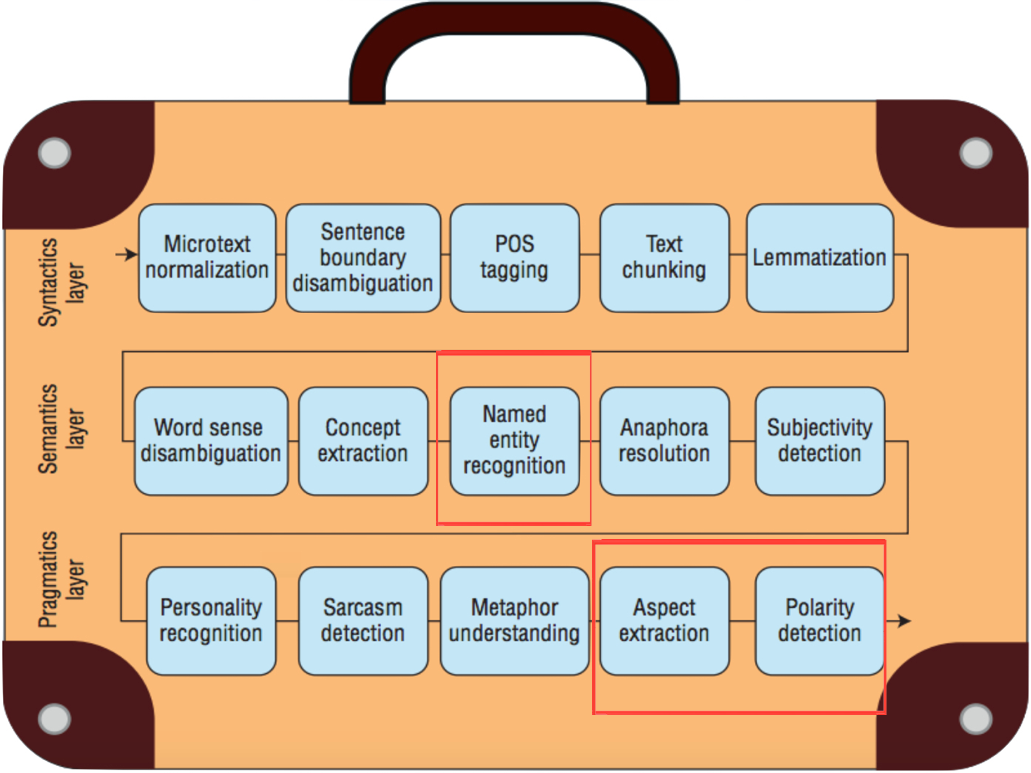}
\caption{A pipeline system~\cite{camsui} illustrating the key components of an opinion understanding system. Tasks selected as the focus of this thesis are highlighted in red boxes.}
\label{fig:pipeline}
\end{figure}
On the other hand, understanding the opinions expressed by human beings is one of the ultimate objectives of NLP. As shown in Fig.~\ref{fig:pipeline}, an opinion understanding system requires the composition of multiple NLP modules involving natural language concepts, from which we identify the following tasks that either requires performing concept-level analysis or can benefit from using concept-level information:

\begin{itemize}

\item \textbf{Named entity recognition (NER)}. Named entities are typically belonging to several semantic classes defined according to the particular interest of downstream applications.  For example, a flight booking assistant system might require recognizing the location names corresponding to departure and arrival cities is the most critical step for retrieving the best matched flight. A typical setting of NER is composed of both tasks of detection and classification. In general, an NER model might take as input a sequence of tokens as well as a predefined set of entity categories, and then outputs the boundaries and semantic types of named entities presenting in the input. 

\item \textbf{Fine-grained named entity typing (FNET)}. Typical setting of NER models aims at resolving only the coarse-grained types of named entities. As an extension of the classification task in NER, FNET aims at recognizing the fine-grained types of a named entity, which are typically organized in a tree-shape hierarchy~\cite{ling:2012}. A natural formulation of FNET is a multi-label multi-class classification problem of which one entity is assigned with multiple categories.

\item \textbf{Targeted sentiment analysis (TSA)}. The targeted sentiment analysis is another concept-centered NLP task resembling the task of FNET. The input to TSA is identical to FNET in which the region of the targeted entity is presumed known, and the ultimate goal is mapping the targeted entity to its sentiment aspects and polarity labels (e.g., positive and negative). More specifically, the task of TSA is composed of two subtasks: 1) aspect categorization and 2) polarity classification. For example, the desired outputs for a product review \textit{Pixel 2 Xl is cheaper than Iphone X} are the sentiment aspect, \textbf{price}, and the corresponding sentiment polarity label, \textbf{negative}, given the target is \textbf{Iphone X}. 
\end{itemize}

Apart from these concept-centered NLP tasks, we also extend the use of concept-based information to automatic speech recognition (ASR) candidate re-ranking, which could augment the opinion understanding system with the capability to perform on data of multi-modality. We assume that concepts such as named entities deserve more attention than functional words for their critical roles for downstream tasks such as spoken language understanding systems or a multimodal sentiment analysis system. Therefore, given a set of ASR decoding candidates, we hope the entity-based information could help us select the candidate the makes fewer mistakes in recognizing important keywords such as person names and location names.

\section{Challenges and Problems Faced by Concept-centered NLP}
In the past decade, great progresses have been made on concept-centered tasks, especially for tasks such as named entity recognition that have been studied for decades. In general, whenever concepts are concerned in these NLP tasks, it is naturally desirable to have an effective representation as well as the corresponding computational structure to leverage the concept-level information.

For knowledge base such as ConceptNet that extracts conceptual facts from the Web automatically, the number of concepts will grow over time. On the other hand, the number of semantic properties (or relations) associated with each concept might also be an obstacle for employing conceptual analysis. 
 
Therefore, a better representation of concepts basically requires tackling two problems: 1) producing a condensed representation of the original feature space, and 2) retain the semantics encoded in the raw data. Ideally, the compressed representation minimizes the dimensionality and information loss at the same time. However, it is hard to fulfill in practice, and a trade-off must be taken in such case.

Apart from an efficient and effective dense representation of concepts, leveraging conceptual information also needs well-designed computation structure in order to be integrated with information flowing from other levels (e.g., word level). 

Most importantly, we believe using only information from a single-level does not suffice to achieve a complete analysis of the natural language. Therefore, the fusion of multi-level information is required for advancing the NLP research.

\section{Contribution}

The focus of this thesis is to leverage concept embeddings for concept-centered natural language processing tasks discussed in the last section and investigating on how to fuse the information of multiple levels.

\begin{itemize}
\item We investigate several novel solutions on fusing the concept-level and word-level information. We demonstrate the effectiveness of such fusion with multiple concept-centered NLP tasks.

\item We propose a novel method to inject concept-based information into the word embedding space which then being used as effective features to recognize concepts/entities from input sentences.

(\textbf{Yukun Ma}, Jung-jae Kim, Benjamin Bigot, and Tahir Muhammad Khan. ``Feature-enriched word embeddings for named entity recognition in open-domain conversations." In Acoustics, Speech and Signal Processing (ICASSP), 2016 IEEE International Conference on, pp. 6055-6059. IEEE, 2016.)

\item We explore to use the pre-trained representation of concept categories (entity labels) in the task of fine-grained named entity typing. We demonstrate that the concept category embedding can help resolve the fine-grained categories of entities. 
(\textbf{Yukun Ma}, Erik Cambria, and Sa Gao. ``Label Embedding for Zero-shot Fine-grained Named Entity Typing." In COLING, pp. 171-180. 2016.)
\item We propose a binary embedding of named entities to help ranking hypotheses of speech recognition.
(\textbf{Yukun Ma}, Erik Cambria, and Benjamin Bigot. ``ASR Hypothesis Reranking using Prior-informed Restricted Boltzmann Machine." In International Conference on
Computational Linguistics and Intelligent Text Processing (CiCLing), 2017)

\item We propose an extension of the Long short-term memory model that could leverage both concept-level and word-level information with more effectiveness.
(\textbf{Yukun Ma}, Haiyun Peng, and Erik Cambria. ``Targeted Aspect-Based Sentiment Analysis via Embedding Commonsense Knowledge into an Attentive LSTM", In the Thirty-Second AAAI Conference on Artificial Intelligence (AAAI-18), pp. 5876-5883, 2018)

\end{itemize}

\section{Organization}
The rest of this thesis is organized as follows: Chapter~\ref{tag:chap2} surveys the background and existing work on concept embeddings and concept-centered NLP tasks; Chapter~\ref{tag:chap3} describes our method on injecting entity-based information into word embedding space to help recognize named entities from telephone conversations; Chapter~\ref{tag:chap4} describes the concept-category embedding used for resolve fine-grained categories of named entities; Chapter~\ref{tag:chap5} presents our binary entity embedding for reranking ASR hypotheses; finally, Chapter~\ref{tag:chap6} presents our work on leveraging commonsense concepts for analyzing sentiment with respect to targeted entities.


\chapter{Background}
\label{tag:chap2}
\pagenumbering{arabic}

In this chapter, we review the related work on NER, ASR reranking, TSA, as well as recent progress on learning concept or word embeddings.

\section{Named Entity Recognition}
Recognizing named entities from texts such as news articles is a concept-level NLP task has been studied by the community. Typically, recognition of named entity requires jointly resolving the location and category of a named entity in its context. In other words, the task could also be split into two tasks: detection and classification (or called Entity Typing). We start with introducing the related work on NER ranging from rule-based approaches of early days to most recent deep neural networks. We then extend our interest to fine-grained entity typing that has recently attracted increasing attention.

\subsection{Rule-based NER}
Early attempts to build NER systems mainly rely on hand-written grammar and rules designed by domain experts for NER challenges, e.g., MUC-7~\cite{bikel:1999}. The context of named entities is critical for rule-based approaches for it characterizing the mention of an entity. For example, one rule can be that any single word following ``Mr" should be tagged as ``PER". Such rules encode information such as graphical information (capitalization, punctuation), syntactic (Part-of-speech tag, syntactic structure) or even semantics (word meanings) related to the current word itself as well as its context. 

Synsets (country, person etc.) in WordNet are used as the reference of named entity types~\cite{alfonseca:2002}. All the words in the same synset are sent as queries to a search engine to retrieve relevant documents. They then collected words appearing in the context of the query words and calculated the frequency. For an input word along with its context, they compared the context words with the contexts of synsets. The word will be assigned with the corresponding named entity type of the synset whose context is most similar to that of the given word.

Evans et al.~\cite{evans:2003} built a document-specific typology for named entities using manually designed patterns, e.g., ``Y such as X" or ``Y like X". Then, they performed clustering on the set of hypernyms in order to identify potential named entity types. Afterwards, a classification was trained with the typology.

Rule-based approaches rely on hand-crafted rules that are expensive to obtain and require additional efforts when adapted to new data or domain. In addition, rule-based approaches are also vulnerable to noises and variations, even though some of the rules can be encoded as Finite-State-Transducer which increases its capability of being generalized.

\subsection{Learning-based NER Model}

Rather than using manual rules encoding the knowledge of experts, machine learning approaches derive knowledge in a data-driven manner and have been widely used for the task of NER. Machine learning approaches typically model the two subtasks of NER as a sequence prediction task using machine learning based approach: given a word sequence $W$ = $w_0$ $w_1$ $w_2$ $\dots$ $w_n$, the objective of NER is to find the optimal tag sequence $T$ = $t_0$ $t_1$ $t_2$ $\dots$ $t_n$, where the tag set is pre-defined. One of the most popular tagging schema used by NER approaches is the BIO encoding. The set of BIO tags includes \textbf{B}eginning, \textbf{I}nside, and \textbf{O}utside tags. For example, the BIO scheme for person names includes $B_{Pers}$, $I_{Pers}$ and $O$, which indicate the beginning of a person name, an internal word of a person name and the word belonging to none of the named entity types. Table \ref{tab:muc7} shows a more specific example of MUC7 where the start and end of \textit{Martin Puris} is indicated by ``B-PER" and ``I-PER" respectively. Given the tagging schema and word sequence, the sequence labeling model searches for the best tag sequence with the parameters learned from training process. In the following, we review the popular sequence labeling models having been used for NER. 

\begin{table}[!htb]
\centering
\begin{tabular}{|L{12cm}|}
\hline
Mr.(O) Dooner(B-PER) met(O) with(O) Martin(B-PER) Puris(I-PER)
,(O) president(O) and(O) chief(O) executive(O) officer(O) of(O)
Ammirati(B-ORG) \&(I-ORG) Puris(I-ORG) ,(O) about(O) McCann(B-ORG)
’s(O) acquiring(O) the(O) agency(O) with(O) billings(O) of(O)
\$400(B-MONEY) million(I-MONEY) ,(O) but(O) nothing(O) has(O)
materialized(O) .(O)\\
\hline

\end{tabular}
\caption{Example BIO tagging for a sample of MUC-7~\cite{muc7}}
\label{tab:muc7}
\end{table}
Charniak et al.~\cite{charniak:1993} proposes a Hidden Markov Model(HMM) for Part-of-speech tagging. Given the word sequence $W$ = $w_0$ $w_1$ $w_2$ $\dots$ $w_n$, the best tagging sequence $T$ = $t_0$ $t_1$ $t_2$ $\dots$ $t_n$ is defined as
	\[\hat{T_{1...n}} = \argmax_{T_{1..n}} P(T_{1...n},W_{1...n}) \]
HMM model assumes that the current word $w_i$ is generated from a multinomial distribution conditioned on current tag while current tag is only conditioned on previous tags. In a first order HMM, the above joint probability can be expanded to
	\[ P(T_{1...n},W_{1...n}) \approx \coprod_{i=1}^{n}P(t_i|t_{i-1})P(w_i|t_i) \]
However, the assumption made by HMM is too strong for NER as the current tag depends on not only the previous tags but also the other types of characteristics in context. For example, in Table \ref{tab:muc7}, HMM model assumes that the tag \textbf{B-ORG} for word \textbf{Ammirati} only depends on the previous tag which is \textbf{O}. This makes less sense than assuming \textbf{B-ORG} depends on the previous word \textbf{officer}. Classic HMM does not allow such dependency between tags and words. Many variations of HMM model have been proposed for addressing this problem, most of which relax the Markov assumption to allow the dependency between previous tags and observations.

HMM models are not flexible enough to train with a rich feature set as the dependency between words and tags are strictly constrained. In order to overcome this shortcoming, Maximum Entropy (ME) model is proposed as an alternative to HMM. It has been widely applied to tagging tasks such as POS tagging for it has the freedom of using arbitrary features.

Similar to HMM-based NER, ME-based model assigns each word in the input with either a tag corresponding to the named entity type and indicating if it is the beginning, or a tag indicating it is not part of an named entity.


 Given the word sequence $W_{1...n}$, the probability of tag sequence $T_{1..n}$ is calculated as:
\[P(T_{1...n}|W_{1...n}) = \frac{\exp^{\sum_{j}^m \alpha_j \dot \phi_j(T_{1...n},W_{1...n})}}{Z(T_{1...n})}\]
where $Z(T_{1...n})$ is the partition function defined as:
\[Z(T_{1...n}) = \sum_T \exp^{\sum_{j}^m \alpha_j \dot \phi_j(T_{1...n},W_{1...n})} \]
The objective of learning process of ME model is to find the optimal setting of parameters $alpha_j$, and during the inference phase, a Viterbi decoding algorithm can be used for finding the most likely tagging sequence for the given input. 

\begin{figure}[!htb]
\centering
  \includegraphics[width=0.7\textwidth]{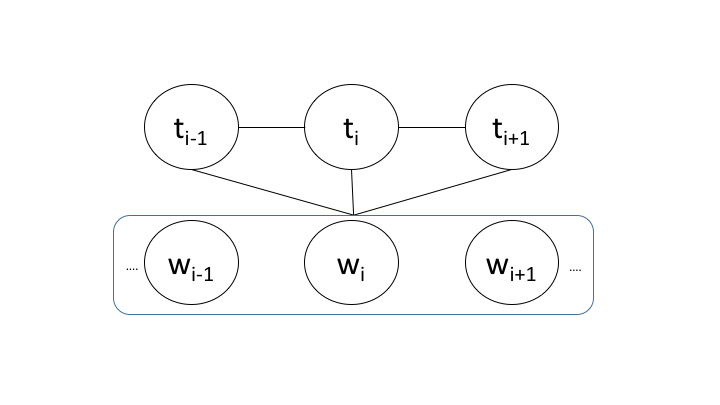}
  \caption{Structure of linear-chain CRF~\cite{lafferty:2001}}
  \label{fig:crf}
\end{figure}

One problem with ME model is known as the \textit{label bias} problem. When there is no alternative transition from one certain state to another, the model will ignore the observation. In other words, when estimating $P(t_i|t_{i-1},w_i)$, the observed word $w_i$ will not have any influence on the model if $t_i$ is the only outgoing transition from $t_{i-1}$. As an alternative to ME model, Conditional Random Field (CRF)~\cite{lafferty:2001} is proposed in order to avoid such label bias problem by estimating $P(t_i|t_{i-1},w_i)$ based on the complete sequence of tags. CRF shares the same exponential probabilistic form as ME model except that CRF model is an undirected graph model. Both HMM and ME are directed models while CRF uses an undirected structure which allows the dependency between words and tags to go in both directions.
 However, long-range dependency becomes intractable as the size of possible features grows exponentially with the number of words in the sequence. To significantly reduce the possible searching space, a simplified version of CRF models allowing only dependency of tags between consecutive time slots, i.e., $t_{i-1}$ and $t_i$, which is called linear-chain CRF (as shown in Figure~\ref{fig:crf}).

As defined by Lafferty et al.~\cite{lafferty:2001}, linear-chain CRF defines the conditional probability of $T$ given the word sequence $W$:
\[P(T|W) = \frac{\exp^{\sum_i^n \sum_{j}^m \alpha_j \dot \phi_j(t_i,t_{i-1},W)}}{Z(T)}\]
while $Z$ is the partition function

\[Z(T) = \sum_t \exp^{\sum_i^n \sum_{j}^m \alpha_j \dot \phi_j(t_i,t_{i-1},W)} \]

, where $T$ is the tag sequence and $W$ is the word sequence; $\alpha_j $ is the weight associated with the $j^{th}$ feature; and $phi_j$ is the $j^{th}$ feature function. Based on the above definition, one of the advantages of CRF is that it dynamically models the dependency between possible tags. In addition, it allows us to define arbitrary feature functions for $T$ and $W$.

Like ME model, CRF model provides the freedom for defining arbitrary features on the sequence of words and tags. However, note that this does not mean that there are no constrains on the dependency. Long range dependency will result in intractable $Z$ as the size of possible tagging sequence is exponential to the length of the word sequence. Similar to first order HMM or second order HMM, only previous two tags are included in the feature set for tractable computation.

\subsection{Deep Neural NER Models}
Most recently, sequence modeling has also benefited from the series of successes achieved by deep neural models. In general, deep neural models relieve the requirement for manually designed features and thus can be effortlessly adapted to problems sharing similar settings. For example, Collobert et al.~\cite{collobert:2011} builds a deep multi-layer network that can be applied to both POS tagging and NER without the requirement of any adaptation. From a different point of view, deep neural models could serve as better feature extractors for discriminative models such as CRF and MaxEnt. In practice, as have been discussed before, many discriminative models such as MaxEnt and CRF have difficulties in using global features and usually have to resort to window-based simplification (e.g., using n-gram-based features). In comparison, deep neural nets such as Recurrent Neural Network (RNN) are capable of encoding the information of the whole sequence thanks to its computational structure (e.g., recurrent connection).

Recurrent neural networks are powerful in modeling sequential data. A RNN takes as input a sequence of words $x_1, x_2, \cdots,x_n$, and outputs a sequence of hidden states $h_1,h_2, \cdots, h_n$, with each $h_i$ representing the subsequence from the beginning to $i$th time step. Despite its theoretical capability of encoding the long term dependency, it turns out that the vanilla setting of RNN tends to capture only the most recent input. Besides, the vanilla setting of RNN is also faced with a serious problem known as gradient vanishment. Namely, the gradient would either vanish or explode when being propagated through many time steps. To overcome this shortcoming, a variant of RNN, called LSTM~\cite{hochreiter1997long}, is proposed with a memory cell as well as several gates controlling the information flow from time to time. Another issue with a single RNN is that it encodes only the unidirectional sequential information. This is suboptimal in the sense that the hidden output at each timestep provides no access to the information of the incoming word sequence. Therefore, a simple yet effective extension is to simultaneously use the outputs of a forward RNN and a backward RNN, which is referred to as bidirectional RNN. 
\begin{figure}[!htb]
\centering
\includegraphics[width=0.7\linewidth]{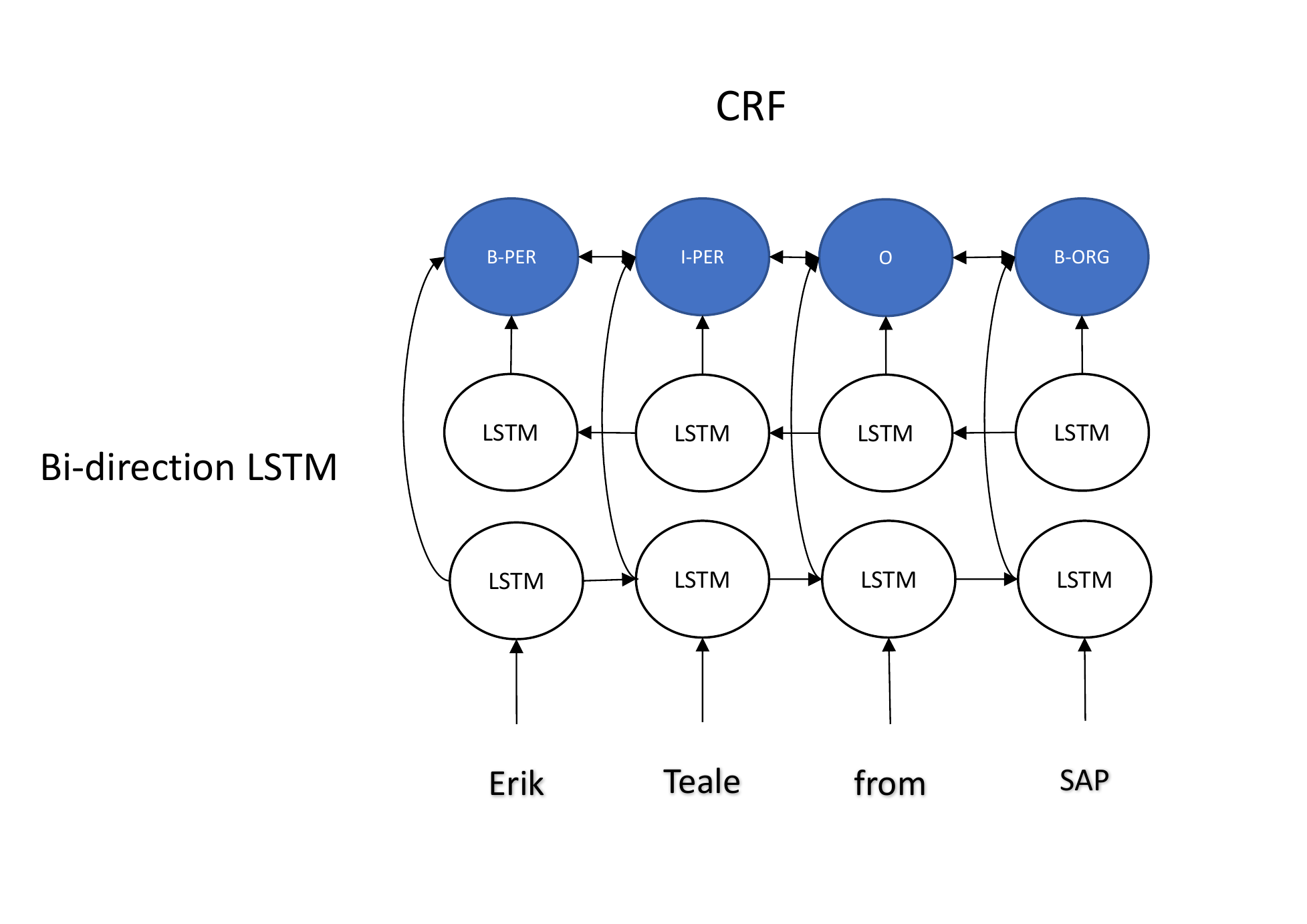}
\caption{Neural architecture of LSTM-CRF used in Guillaume et al.~\cite{lstm-ner}}
\label{fig:ner-arch}
\end{figure}

RNNs became natural choices for modeling NER problem as soon as they are invented. As compared with previous work such as CRF or HMM, RNNs are shown better at leveraging global information lying with the whole sequence. But, RNNs fail to model the dependency between outputs of different time steps which can be easily achieved by models like CRF or HMM. 
Most recently, NER models~\cite{lstmcnncrf,lstm-ner} using a neural structure combining the LSTM with CRF components (as shown in Figure~\ref{fig:ner-arch}) have also been proposed. In the new architecture, a bidirectional LSTM is used as a feature extractor, while a linear-chain CRF component is used for modeling the dependency between the output tags of neighboring timesteps.

\subsection{NER Features}

All the aforementioned supervised NER models rely on a feature set to represent the inputting sequence of words. Therefore, it is critical to design and use features capturing the most discriminative information marking the presence and categories of a named entity. These features basically fall into two categories: manual designed features and automatically learned features. 
The following are a subset of features used by Collin et al.~\cite{collins:2002},

\begin{center}
\begin{tabular}{l}
Features\\
\hline
Current word\\
Previous word\\
Next word\\
Previous tag\\
Tag bi-grams (previous two tags)\\
If the word is the first word of a sentence\\
If the word appears as uppercase more frequently than lower case\\
Type of the initial word, i.e., uppercase, lower case, digit, etc.\\
Word shape, i.e., each character in the word is mapped to its type\\
\end{tabular}
\end{center}
It can be found that some of the manually designed features shares the same intuition of the hand-crafted rules used by a rule-based system. Basically, the feature set consists of grapheme features of the word at the current step as well as previous and next steps as contextual markers of the entities.

The manually designed features require human knowledge and are hard to be generalized to new domains or new datasets. In comparison, the second class of features are data-driven and can be automatically learned without the need for intervention. Most notably, NER models~\cite{collobert:2011,lstm-ner,lstmcnncrf} using pre-trained word embeddings (or phrase embedding) as features have successfully achieved state-of-the-art performance. Word embedding is a type of vector representation of words that is typically learned via auxiliary tasks such as language modeling~\cite{mikolov2013distributed}. We will cover more reviews on the technique of learning embeddings of words or concepts in section\ref{intro:embd}.



\section{Fine-grained NER}
Fine-grained named entity recognition refers to the task of recognizing named entities belonging to a fine-grained set of entity categories. As compared with coarse-grained NER, the fine-grained entity types are organized into a hierarchy. 
In contrast to coarse-grained NER, the fine-grained NER is generally modeled in two separate steps: 1) segmentation, which is to detect the boundaries of named entity mentions; and 2) classification, which is to classify the detected named entity into fine-grained types. It is also likely that the classifier might have accesses to pre-existing entity boundaries rather than reading the outputs of the detection model. In such case, classification of entity mentions becomes the only focus of Fine-grained NER, which is also known as the task of Fine-grained Entity Typing (FNET).

One of the biggest challenges faced by fine-grained named entity recognition is the shortage of training data. Since the number of fine-grained types is much bigger than coarse-grained of named entities, it is harder to have sufficient training data for each category. Fortunately, structural information in public resources such as Wikipedia and Freebase can be used for automatically building annotation of fine-grained named entity types. In this section, we review the most recent work on FINER or FNET: FIGER~\cite{ling:2012}, HYENA~\cite{hyena:2012}, GFT~\cite{googlefner:2014}, the embedding based method{yogatama:2015}, as well as methods based on deep neural networks~\cite{shimaoka:2017,abhishek:2017}.
\subsection{Fine-grained Labels}

\begin{figure}[!htb]
\centering
\includegraphics[width=\linewidth]{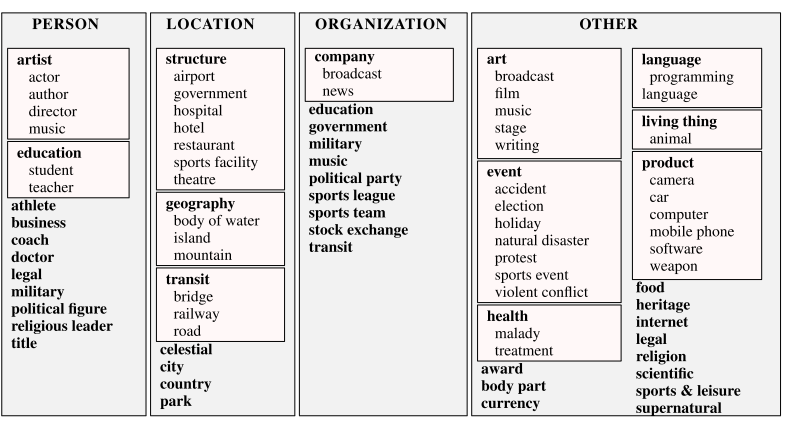}
\caption{Hierarchical categories used in Ling et al.~\cite{googlefner:2014}}
\label{fig:hier}
\end{figure}

The first step to creating a benchmark for fine-grained NER is identifying the set of labels (or classes) of named entities. Different from coarse-grained NER, whose label set is typically limited to a few categories such as Organization, Location, and Person, there's actually no standards for fine-grained types. However, previous works have used online knowledge base for building the label set as well as their hierarchy. FIGER derives a set of 112 named entity types from Freebase, an online knowledge base built on Wikipedia and WordNet, even though the hierarchy only has two levels. HYENA uses a much larger tagset composed of 505 entity types by selecting top 100 subtypes for each top-level types of YAGO knowledge base~\cite{yago:2007,yago2:2013}. Additionally, the labels used in their approach are organized as a taxonomic hierarchy. More recently, GFT manually refined and expanded the label hierarchy used by FIGER into three levels. As shown in Fig.~\ref{fig:hier}, the labels are organized into a three-level hierarchy, where the first level contains only four categories, i.e., PERSON, LOCATION, ORGANIZATION, and OTHER (or MISC), which are typically used in coarse NER.

\subsection{Automatic Generation of Training Data}
\begin{figure}[!htb]
\centering
\includegraphics[width=0.6\linewidth]{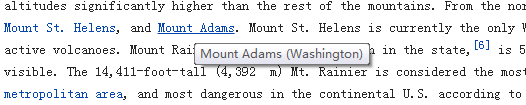}
\caption{Example of anchor text on a Wikipedia page.}
\label{fig:figer_data}
\end{figure}

Training classifiers for fine-grained entity types requires a considerable number of labeled data, which is expensive and time-consuming. Instead, previous methods have been trying to use online resources such as Wikipedia to automatically construct labeled data. Most of the times, anchor text in Wikipedia is in fact marking a mention of named entity. The link of anchor text redirects to the Wiki page corresponding to the classic form of that given named entity mention (as shown in Fig~\ref{fig:figer_data}). Despite that the Wikipedia page has already the categories of that named entity, they are very noisy as many of them are automatically given by the system of Wikipedia. Instead of using the Wikipedia categories, Ling et al.~\cite{ling:2012} propose to use the freebase categories which contains fewer noises. By creating the mappings from the classic form to a freebase category and from freebase to manually refined labels, a large annotated corpus (approx. 3 million sentences) is built from the anchor texts of Wikipedia pages and Freebase. 

However, the labels in the annotated data built by~\cite{ling:2012} still contain noises, because the mapping from classic forms to freebase concepts do not consider the impacts of context during tagging. For example, $ishington$ can be a $military\_leader$ or a $city$ according to Freebase, but in FIGER data, no matter what the contexts is, $Washington$ is always annotated with both the two labels. To overcome this shortcoming, ~\cite{googlefner:2014} used the context of named entities to refine the labels with the help of a set of heuristic rules which takes into account the contexts of named entities:
\textbf{Sibling}, using the parent label, if the mention is tagged with multiple labels having the same parents;
\textbf{Coarse-type}, using softmax classifiers trained on ACE corpus, which classify named entity mentions into top-level labels based on their local context, and removing fine-grained labels that are inconsistent with the results of coarse-grained classification;
\textbf{Minimum count}, removing the labels appearing less than a certain number of times within the current documents. These heuristic rules actually result in training instances with higher confidence as compared to FIGER.


\subsection{Features}
Fine-grained NER methods convert the input into a vector of features. Most of the features used in fine-grained NER are shared across existing methods. In Table ~\ref{tab:fine-feat}, we list the features have been used in all of the four systems. We observed that most of the features are adapted from coarse-grained NER.

\begin{table}[!htb]
\begin{center}
\begin{tabular}{|L{3.4cm}| L{7cm}| L{3.5cm}| }
\hline
Feature&Description&Example\\
\hline\hline

Phrase head& Head word of a named entity mention & ``Obama"\\
Non-head & All the non-head words of the mention&``Barack", ``H"\\
Head cluster& Word cluster ID of the head word& ``59"\\
Character n-gram& character n-grams of the head word&``:ob",``oba",...\\
Word Shape & The word shape of the mention& ``Aa A Aa"\\
Role & Dependency label of the head word&``nsubj"\\
Context &context word of the mention& ``B:who", ``A:first"\\
Parent &Lexical parent in a dependency tree& ``picker"\\

\hline
\end{tabular}
\end{center}
\caption{Features shared by all the four related work}
\label{tab:fine-feat}
\end{table}

\subsection{Segmentation and Multi-label Classification}
Depending on the settings of experiments, different methods might use different strategies to obtain the segmentation of named entity mentions. For example, human-annotated segmentation of testing data might be directly used as input~\cite{googlefner:2014,yogatama:2015,shimaoka:2017,abhishek:2017}, even though the assumption that the entity boundaries are pre-existing sometimes does not hold in reality. HYENA uses the segmentation provided by the anchored text of Wikipedia page that is not as reliable as manual annotations. In contrast, the setting of FIGER is the most reasonable and practical one. Their testing set is manually labeled. They train a CRF model with BIO tagging schema for segmentation of text into named entity mentions and non-mention during testing.

The classification of fine-grained NER is typically modeled as a multi-label classification problem. According to Silla et al.~\cite{silla:2011}, most of the existing multi-label classifiers fall into four categories: 1) \textit{flat}, referring to those using a single multi-class classifier; 2) \textit{local}, referring to methods using one-versus-rest binary classifiers for each node in the hierarchy; 3) \textit{local per parent node}, associating a multi-class classifier with each internal node; and 4)~\textit{global}, using a single multi-class classifier with objective function encoding the similarity between labels.
As an example of the first category (\textit{flat}), ~\cite{ling:2012} adapted a simple linear classifier based on single perceptron~\cite{rosenblatt:1958}, which maps an input x to a label $y$ given the feature function $F(x,y) = [f_1(x,y),\dots,f_n(x,y)]$,
$$\hat{y} = argmax_y \sum_{i=1}^n w_i \cdot f_i(x,y)$$
where $w=[w_1,\dots,w_n]$ are the vector of weights learned from training set. During training, $w$ is updated by
$$ w \leftarrow w + \alpha(\sum_{y}F(x,y)-\sum_{\hat{y}}F(x,\hat{y}))$$
where \{$y$\} are the set of true predictions, and \{$\hat{y}$\} are the set of mispredictions, which is discouraged by decreasing the weights associated with the corresponding features. In order to get multiple labels for one input $x$, all the tags $y$ with positive score are selected as the predicted labels. 

HYENA trained one binary SVM for each node in the hierarchy using all the siblings of that node as negative examples, and all instances of children node as positive examples. In this sense, HYENA falls into the \textit{local} category. Since the hierarchy of labels might be incomplete, it creates an additional child node \textit{OTEHR} for each inner node in the hierarchy. HYENA classifies the mention of named entity in a top-down manner, assigning each mention with all labels accepted by the classifier, and stopping at the current level when classifier chooses the OTHER label. GFT experiments with both \textit{local} and \textit{flat} classifiers. They also use SVM as the choice for local classification in a way similar to HYENA, while using a softmax classifier as the local approach. 

\begin{figure}[!htb]
\centering
\includegraphics[width=0.8\linewidth]{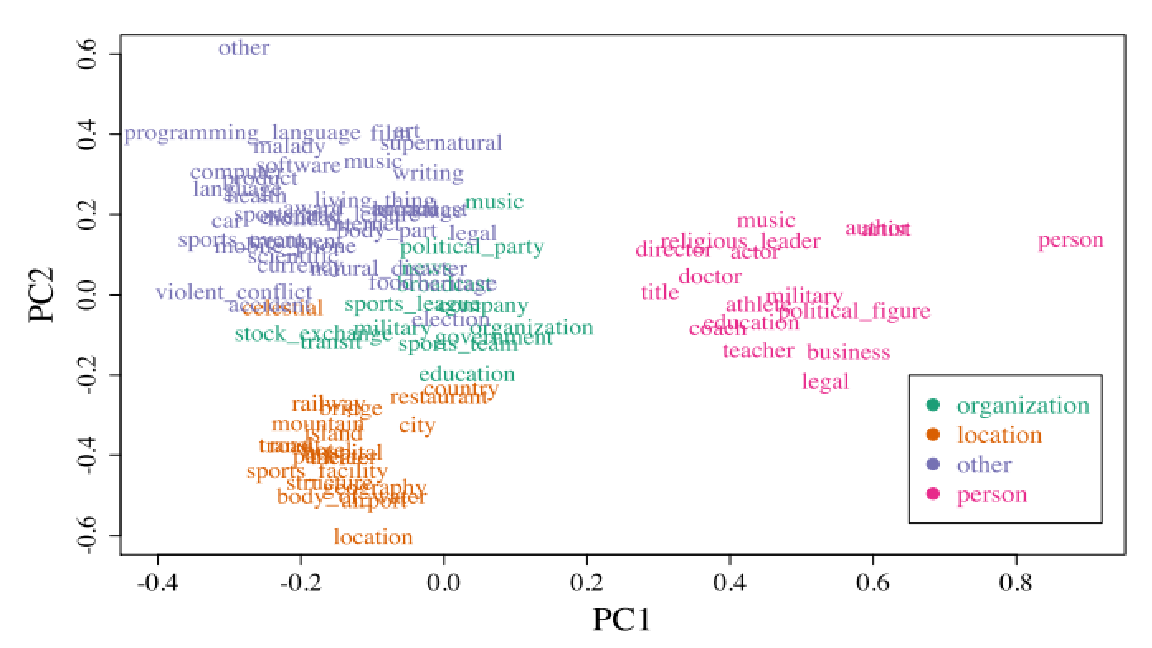}
\caption{Label in the projected space~\cite{yogatama:2015}}
\label{fig:wsabie}
\end{figure}

All the above three methods, i.e., FIGER, HYNNA, and GFT, do not consider the similarity between labels, which would be very useful for handling the data sparseness when a big label set is in use. Instead of using traditional classification methods, WsabieNER used an embedding-based method, called Wsabie~\cite{wasbie:2011}, which maps both labels and features to same lower dimensional space using liner transformations
$$F(x)\rightarrow AF(x)$$
$$y \rightarrow By$$

where $F(x)$ is the vector of features, and $y$ is the vector of labels. The classification is done by scoring a pair $(x,Y)$ with the products of the transformed vectors,
$$s(x,y) = AF(x)By$$
To learn the parameters, $A$ and $B$, from training data, they optimize an objective function called weighted approximate pairwise (WARP)~\cite{wasbie:2011}, which is designed to rank the true prediction $y$ higher than the misprediction $\hat{y}$. The mathematical description of WARP is 
$$\sum_y\sum_{\hat{y}} R(rank(x,y)) max(1-s(x,y)+s(x,\hat{y})),0)$$
where $rank(x,y)$ is margin infused counting function defined as 
\[rank(x,y) = \sum_{\hat{y}} \mathbb{I}(s(x,y)-1<s(x,\hat{y})\]
, and $R(k)=\sum_{i=1}^{k}\frac{1}{i}$ is a function turning the rank into a weight.
By optimizing this objective function, $F(x)$ and $y$ are mapped to lower dimensional space where labels cooccurring with similar feature vector $F(x)$ are close to each other. Fig.\ref{fig:wsabie} shows that in 2-dimension space (after applying PCA, and labels are colored based on their top-level labels), labels sharing parent node or superclass tend to be grouped together. As the prediction is made based on the products of label and feature embeddings, this also suggests that the classifier is less discriminative to labels of a lower level, which tends to be close in new space, while more discriminative to labels of a higher level.
\subsection{Label Noise Reduction}
As mentioned above, data used for training FNET models are weakly tagged. It is unavoidable that the weak tagging process would introduce label noises. More specifically, the noisy instances are mostly over-labeled because of the heuristic rule of weak tagging. For example, in the dataset created by Ling et al.~\cite{ling:2012}, each instance is assigned with all the possible entity types if the surface form matches with multiple canonical forms. 
Some previous studies~\cite{ren:2016} have formulated this problem as a partial label learning problem where the true label set is a subset of the observed label. Namely, 
\[\hat{Y} \cap Y = \hat{Y}\]
where $\hat{Y}$ is the true label set and $Y$ refers to the set of observed labels. The learning process can then be based on loss functions considering only a subset of the observed label set. For example, one of the loss function used in the previous work~\cite{ren:2016} has the form like
\[max\{0,1- max_{y\in Y}s(x,y) + max_{y'\in Y'} s(x,y')\}\]
where $Y'$ refers to the set of negative labels. The intuitive behind this loss function is to learn from only the most confident label of a noisily labeled instance.
\begin{figure}
\centering
\includegraphics[width=0.8\textwidth]{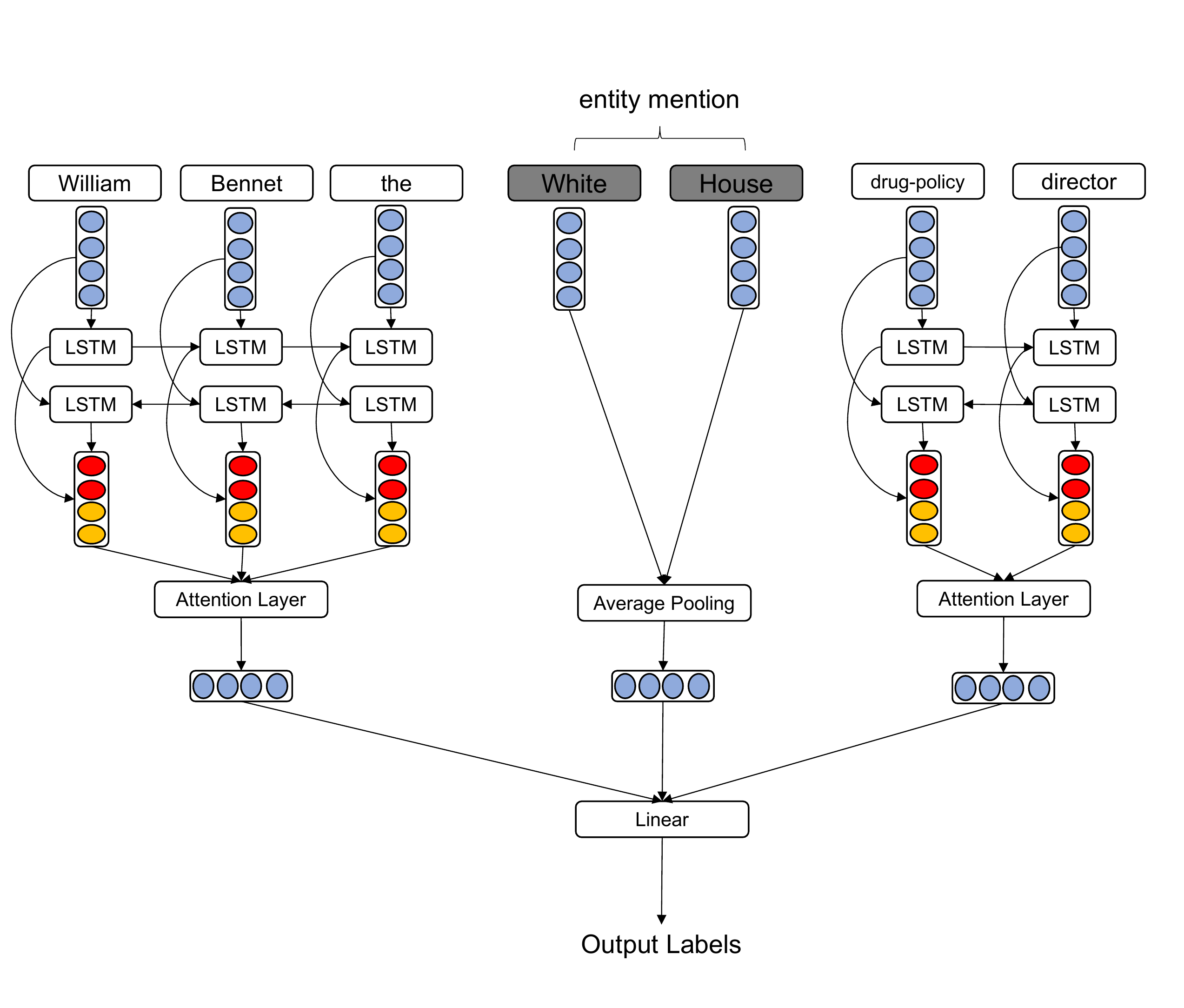}
\caption{Neural Architecture used by Shimaoka et al.~\cite{shimaoka:2017}}
\label{fig:shimaoka}
\end{figure}
\subsection{Neural Architecture for FNET}
Following the trend in adopting deep neural networks for NLP tasks, recent work on FNET has adapted the LSTM to modeling the sequence of contextual words as well as of entity mentions. For example, Shimaoka et al.~\cite{shimaoka:2017}(as illustrated in Fig.~\ref{fig:shimaoka}) proposes to encode the contextual word sequences using bi-directional LSTM. Their framework also incorporates an attention component to model the salience of each contextual word in resolving the types of the center entity mention. A similar neural network is also adopted by work of Abhishek et al.~\cite{abhishek:2017} which replaces the bag-of-words encoder of entity mentions with another uni-directional LSTM encoder. Their work shows that the neural architecture is able to perform comparably than embedding methods using the manually designed features.

One question is how to encode the correlation of labels. In other words, the labels are not created free of semantics. Each label in the hierarchy, in fact, corresponds to a concept in real world. It is, therefore, useful to model the correlation or dependency of labels. In HYENA~\cite{hyena:2012}, although the classification results are passed from parent to its children, each classifier, however, is trained independently. In comparison, the embedding based methods~\cite{yogatama:2015,xiang:2015} captured the similarity between labels by considering the concurrence of labels and features. But, their methods fail to take into consideration the existing hierarchy that has encoded useful information about the dependency between labels. To address these issues of existing approach, one solution is to consider the Bayesian hierarchical classifier~\cite{gopal:2012}, which models the dependency between labels using a hierarchical prior. Their experiments have shown improved performance over $flat$ and $local$ classifiers on a range of tasks and datasets and is scalable to work with large-scale training data.


\section{Targeted Sentiment Analysis}

In this section, we survey multiple research areas related to the proposed framework, namely: Aspect-based Sentiment Analysis (ABSA), targeted sentiment analysis, targeted ABSA, and finally works on incorporating external knowledge into deep neural models.

\subsection{Aspect-Based Sentiment Analysis}
ABSA is the task of classifying sentiment polarity with respect to a set of aspects. The biggest challenge faced by ABSA is how to effectively represent the aspect-specific sentiment information of the whole sentence. Early works on ABSA have mainly relied on feature-engineering to characterize sentences~\cite{wagner2014dcu,kiritchenko2014nrc}. Motivated by the success of deep learning in representation learning, many recent works~\cite{dong2014adaptive,lakkaraju2014aspect,nguyen2015phrasernn,wangattention} utilize deep neural networks to generate sentence embeddings (dense vector representation of sentences) which are then fed to a classifier as a low-dimensional feature vector. Moreover, the representation can be enhanced by using the attention mechanism~\cite{wangattention}, which is typically a multi-layer neural network taking as input the word sequence and aspects. For each word of the sentence, the attention vector quantifies its sentiment salience as well as the relevance to the given aspect. The resulting sentiment representation benefits from the attention mechanism for it overcomes the shortcoming of recurrent neural networks (RNNs), which suffer from information loss when only one single output (e.g., the output at the end of the sequence) is used by the classifier. 

\subsection{Targeted Sentiment Analysis}
Targeted sentiment analysis aims to analyze sentiment with respect to targeted entities in the sentence. It is thus critical for targeted sentiment analysis methods, e.g., the target-dependent LSTM (TDLSTM) and target connection LSTM (TCLSTM)~\cite{tang2015effective}, to model the interaction between sentiment targets and the whole sentence. In order to obtain the target-dependent sentence representation, TDLSTM directly uses the hidden outputs of a bidirectional-LSTM sentence encoders panning the target mentions, while TCLSTM extends TDLSTM by concatenating each input word vector with a target vector. Similar to ABSA, attention models are also applicable to targeted sentiment analysis. Rather than using a single level of attention, deep memory networks~\cite{tang2016aspect} and recurrent attention models~\cite{chen:2017} have achieved superior performance by learning a deep attention over the single-level attention, as multiple passes (or hops) over the input sequence could refine the attended words again and again to find the most important words.
All existing approaches have either ignored the problem of multiple target instances (or words) or simply used an averaging vector over target expressions~\cite{tang2016aspect,wang2017td}. Unlike such approaches, our method weights each target word with an attention weight so that a given target is represented by its most informative components. 

\subsection{Targeted Aspect-Based Sentiment Analysis}
 Two baseline systems~\cite{saeidi:2016} are proposed together with SentiHood: a feature-based logistic regression model and an LSTM-based model. The feature-based logistic regression model uses feature templates including n-grams tokens and POS tags extracted from the context of instances. The LSTM baseline can be seen as an adaptation of TDLSTM that simply uses the hidden outputs at the position of target instances assuming that all target instances are equally important. 
 
\subsection{Incorporating External Knowledge}
External knowledge bases have been typically used as a source of features~\cite{ratinov2009design,rahman2011coreference,nakashole2015knowledge}. Most recently, neural sequential models~\cite{ahn2016neural,yang-mitchell:2017} leverage the lower-dimensional continuous representation of knowledge concepts as additional inputs. However, these approaches have treated the computation of neural sequential models as a black-box without tight integration of knowledge and computational structure. The proposed model, termed Sentic LSTM, is inspired by~\cite{xu2016incorporating}, which adds a knowledge recall gate to the cell state of LSTM. However, our method differs from~\cite{xu2016incorporating} in the way of using external knowledge to generate the hidden outputs and controlling the information flow. 

\section{ASR Hypoehses Reranking}
Another research line to be reviewed is ASR hypotheses reranking. It is a technique to improve the quality of texts automatically transcribed by a speech recognition system. ASR hypoehteses reranking is critical for understanding opinions expressed via not only texts but also audios. 
ASR hypotheses reranking might sometimes be referred to as discriminative language modeling in contrast to generative models such as the n-gram model used by~\cite{Hatmi2013:jmm}.

Typically, ASR systems consist of two modules:1) acoustic model and 2) language model. The best hypothesis of an ASR system is induced with 

$$\argmax_{W} P(W|A) = \argmax_{W} \log(P_a(A|W)) + \alpha \log(P_l(W))$$
\noindent
where $A$ is the acoustic representation of an utterance, and $W$ is the hypothesized word sequence, respectively. $P_a(s|a)$ is the acoustic model, and $P_l(s)$ the language model with $\alpha$ controlling the relative importance of the language model.

Speech transcripts contain disfluencies such as repetitions, false starts and fillers, which can interrupt the consecutive n-gram patterns. The conventional n-gram language model, therefore, is unsatisfactory for modelling spoken language, especially for conversational speech which is spontaneously incurred. In order to address the problems caused by the informal nature of speech, some works have modeled the long distance context using syntactic parse trees. Approaches using syntactic information for language modeling either incorporate the parsing process into language modeling or maintain a separate process by including features generated from syntactic parsing. As an example of the first type of approaches, ~\cite{chelba:1998} used a shift-reduce parser to maintain a set of possible parses of a given sentence. The probability of a word given its previous context is then defined as the sum of probabilities of candidate parses. As for the second type, ~\cite{Collins:2005dsl,Lambert:2013dp} used single perceptron model with features extracted from syntactic parse trees and dependency tree. Their experiments show that, when combined with the ASR baseline model, discriminative reranking could further reduce the word error rate, even though by a small margin. A feature function $F(W)$ is defined for a given word sequence W, and used for rescoring the hypothesis of ASR with a modified log probability
$$\log(P(W,A)) = \log(P_a(A|W)) + \alpha \log(P_l(W)) +\sum_i^N f_i(W) \cdot w_i $$
where $f_i$ is the $i^{th}$ feature in $F(W)$, and $w_i$ is its weight. 
The syntactic features used by~\cite{Collins:2005dsl} are derived from the parse trees produced by a syntactic parser, including head features such as the syntactic head of a detected noun phrase, and head-to-head dependencies. Instead of using syntactic parsing based features, ~\cite{Lambert:2013dp} extracted features from dependency parses, which are mainly composed of a set of dependency tuples. For example, for a sentence

 \textit{I WENT TO THE AN- THE DOCTER}
 
Stanford parser produces the set of tuples as shown in Table.\ref{tab:tuples}. Each tuple represents a dependency link in the parsing result, including two nodes and their syntactic relation such as noun subject(\textit{nsubj}) between a verb and subject. All the tuples generated for the given sentence are added to the feature set.

\begin{table}
\centering
\begin{tabular}{|l|}
\hline
nsubj(went-2, I-1)\\
root(ROOT-0, went-2)\\
case(docter-7, to-3)\\
det:predet(docter-7, the-4)\\
det(docter-7, an-5)\\
amod(docter-7, --6)\\
nmod(went-2, docter-7)\\
\hline
\end{tabular}
\caption{The dependency tuples for sentence \textit{I WENT TO THE AN- THE DOCTER}}
\label{tab:tuples}
\end{table}
One interesting finding made by Lambert et al.~\cite{Lambert:2013dp} is that a large number of features are actually redundant for reranking. After performing feature selection, they found that the reranker can still achieve comparable results using only 0.5\% of the original feature set. In addition, their results also show that approximate 80\% features are extracted less than 5 times, which suggests a problem of data sparseness. It is worth a further investigation on whether reranking can benefit from reducing the data sparseness.

\section{Embedding Models}
\label{intro:embd}
As discussed in previous sections, many NLP models transform the input into feature vectors of very high dimensionality. Computation on such high-dimensional space is very expensive and sometimes even intractable. To address this issue, dimensionality reduction tricks have been adopted to convert the original feature vector into lower dimensional representations which are also known as embeddings. As compared with one-hot representation whose dimensionality is typically same as the size of feature dictionary, embedding approaches could significantly reduce the dimensionality without suffering a significant loss of information. Depending on the tasks, embeddings can be generated for not only words but also concepts or phrases. 

One of the fundamental intuition of learning embeddings for words or concepts is distributional semantics which assumes the distribution of contexts characterizes the semantic meaning of a central word. In other words, two words (or concepts) are close in semantics if having the similar distribution of contexts. Recently, embedding based on distributional semantics have been used as effective features for many NLP tasks, e.g., named entity recognition~\cite{lstm-ner}, dependency parsing~\cite{bansal:2014}, semantic relation classification~\cite{hashimoto:2015}, antonym detection~\cite{masataka:2015}, sentiment analysis~\cite{li2017learning}, and spoken language understanding~\cite{tasos:2014}. 

According to the classification by Baroni et al.~\cite{baroni2014don} and Levy et al.~\cite{levy:2015}, existing word embedding methods based on distributional semantics can be grouped into two categories: count-based and prediction-based. The count-based methods can be dated back to methods such as Latent Semantic Analysis (or Latent Semantic Indexing)~\cite{deerwester1990indexing} that are followed by lines of research on decomposing matrices of co-occurrence counts. 

\subsection{Count-based Models}
It is a long history that the co-occurrence counts and reweighting tricks have been used for obtaining embeddings. Here, we are only able to review a small subset of them that have recently attracted enough attention. 

 First of all, the task of learning embeddings from co-occurrence counts can be formulated as a matrix factorization problem. Let $M_{ij}$ be an entry of $M$. Given the training dataset $D= \{w_1,w_2,\cdots,w_n\}$, the value of $M_{ij}$ can be either original or weighted counts of co-occurrence of word $w_i$ and a context $c_j$. For example, a popular weighting schema is Pointwise Mutual Information (PMI)~\cite{church1990word}.

 We then formulate the process of deriving word embeddings as a mapping from co-occurrence matrix $M \in R^{n\times m}$ to embedding matrix $W^{n\times d}$, with each row of $W$ corresponding to a $d$-dimensional vector representation of the word $w_i$.

Next, we will review two of the count-based methods that are recently studied for learning embeddings of words or concepts.

\subsubsection{Single Value Decomposition}
A natural choice to decompose the co-occurrence matrix to lower-dimension vectors is to use traditional dimension reduction methods such as Single Value Decomposition (SVD). 

SVD generates an optimal low-rank approximation of the original counts matrix $M$ by factorizing it as
\[ M = U\Sigma V^\top\]
where $U \in R^{n\times n}$ and $V\in R^{m\times m}$ are unitary matrices, and $\Sigma \in R^{n\times M}$ being a diagonal matrix. Each diagonal entry of $\Sigma$ is a singular value of $M$. 
To get a $d$-dimensional vector representation, we just take the dot product of $U*$ and $\Sigma*$, i.e.,	
			\[W = U^*\Sigma^*\]
where $\Sigma^*$ is the same matrix as $\Sigma$ except that it contains only the top $d$ values, and $U^*$ being the corresponding truncated matrix of $U$.

\subsubsection{Random Projection}
\label{sec:rp}
One problem with SVD is that it is hard to scale up, especially when the original matrix is sparse and too high-dimensional~\cite{balduzzi2013randomized}. 

Cambria et al.~\cite{camaf2} propose to a data oblivious method, called random projection, to map the original high-dimensional space to a lower dimensional space. The adopted random projection of concept into the new space is done via a Gaussian matrix. According to Johnson and Lindenstrauss (JL) Lemma~\cite{johnson1986extensions}, points distributed in a $m$-dimensional space can be safely projected to a $d$-dimensional space, where $d$ is logarithmic to $m$, and pairwise distances being preserved with high probability. It is, therefore, possible to use a random matrix to transform each row of original co-occurrence matrix to a lower-dimensional row vector. 

Cambria et al.~\cite{camaf2} propose to use a fast random projection technique, called Subsampled Randomized Hadamard Transform (SRHT)~\cite{lu2013faster} a random matrix $\Phi$, which can be factorized by

\[\Phi = \sqrt{\frac{m}{d}}RHD\] 
where $R$ is a $d\times m$ with rows being sampled from an uniform distribution, and $H$ is a $m\times m$ normalized Walsh-Hadamard matrix, and $D$ is $m \times m$ random matrix. 	

After computing $\Phi$, the new embedding matrix $W$ can be obtained by linearly transforming $M$ using the transpose of $\Phi$, i.e., 
\[W= M\Phi^{\top}\]
As shown by Cambria et al.~\cite{camaf2}, the random projection method could project the very high-dimensional concept matrix to a low-dimensional space with high efficiency. 

\subsection{Prediction-based Models}
Apart from working with co-occurrence counts, the past few years have seen a trend of learning embedding by maximizing the probability of contextual terms. The family of embedding learning methods~\cite{collobert:2011,huang:2012,mikolov:distributed2013} are, therefore, classified as prediction-based methods. Next, we will review one of the most widely-used prediction-based methods, called Skip-gram model, as well as a variant of it, called CBOW model.
\subsubsection{Skip-gram Models}
\label{sec:skipgram}
\begin{figure}
\centering
  \includegraphics[width=150mm,scale=3.0]{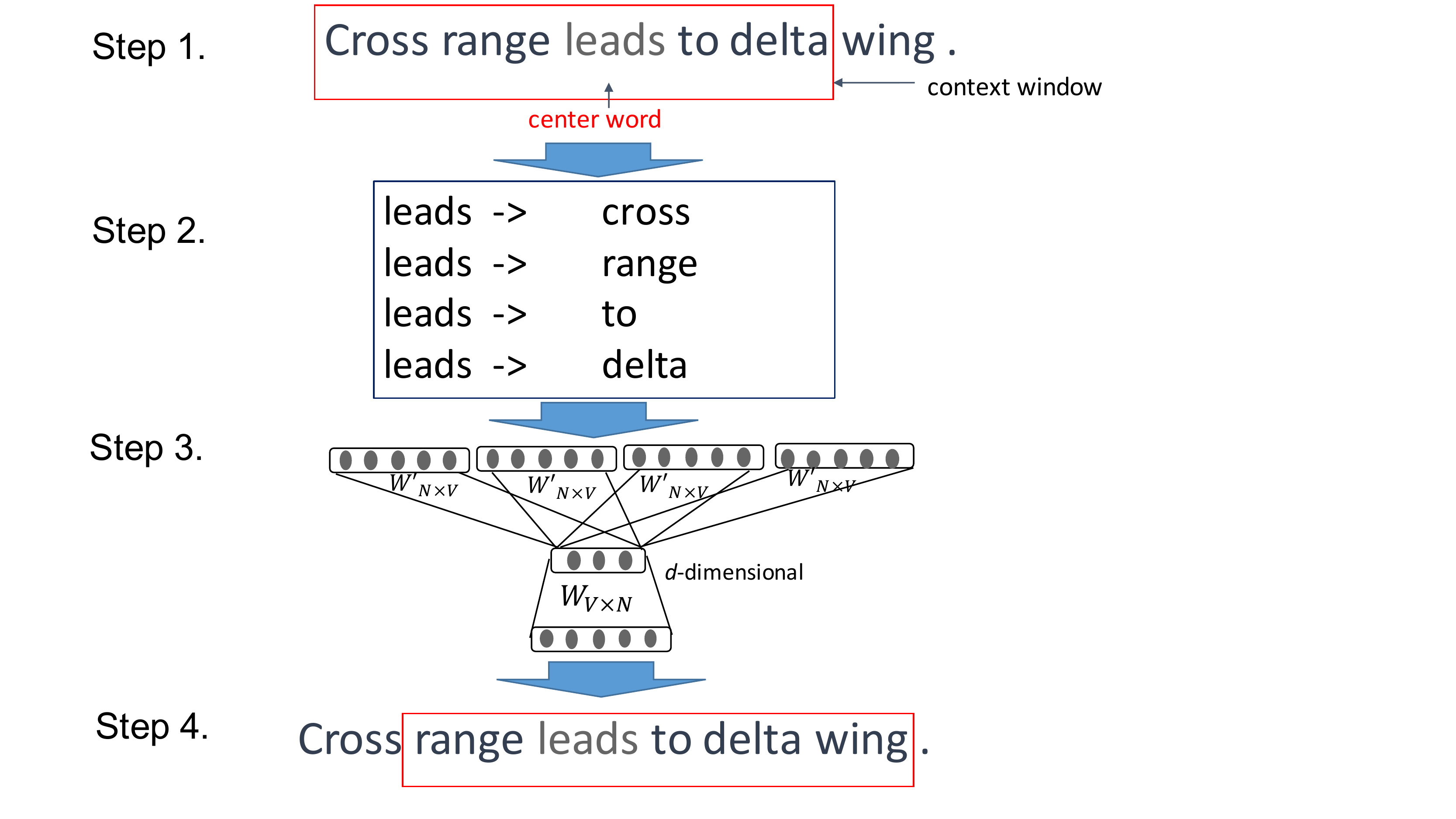}
  \caption{Training process of Skip-gram model.}
  \label{fig:skip}
\end{figure}

Skip-gram model~\cite{mikolov:distributed2013} is a neural network model that can be efficiently learned from a large-scale corpus with billions of words. Figure.\ref{fig:skip} shows the training process of Skip-gram model. The model first takes a window of the local context of the given word to generate the word pairs. Then, it updates the parameters of the neural network using each word pair. Afterwards, it moves to next position and loop over Step1-4 again until the end of the corpus.

Its objective function is the sum of log probabilities $p(w_{i+j}|w_i)$ over the whole corpus,
$$ \sum_{i=1}^{N}\sum_{j=-k}^{k}\log p(w_{i+j}|w_i)$$
where ${w_{i+j}}$ indicates a neighbor word of $w_i$, $k$ the size of the context window and $N$ is the length of the dataset. And, the basic Skip-gram formulation defines $p(w_{i+j}|w_i)$ using the softmax function as follows:
$$p(w_{i+j}|w_i) =\frac{ \exp(v'^\top_{w_{i+j}}v_{w_i})}{\sum_{w \in W}\exp({v'^\top_{w}v_{w_i})}}$$
where $W$ is the vocabulary of words, $v'_{w_{i+j}}$ and $v_{w_i}$ are the context and word embeddings, respectively.
\subsubsection{Continuous Bag-of-Words (CBOW) Model}

The training objective CBOW model~\cite{mikolov:distributed2013} is to maximize the sum of probability $p(w_i|w_{i-k},\dots,w_{i+k})$, which is similar to Skip-gram model.
$$ \sum_{i=1}^{N}p(w_i|w_{i-k},\dots,w_{i+k}))$$
The probability of $w_i$ given contexts words \{$w_i+j$\} is defined on top of $v'_{w_i}$, which is the context embedding of $w_i$ and $h$ which is defined as the average vector of words in the context
$$h = \frac{1}{2k}\sum_{j=-k}^{k} v_{w_i+j}$$
Same as Skip-gram model, the probability $p(w_i|w_{i-k},\dots,w_{i+k})$ is normalized using softmax function
$$p(w_i|w_{i-k},\dots,w_{i+k}) =\frac{ \exp(v'^\top_{w_i} h)}{\sum_{w'\in W}\exp({v'^\top_{w'} h)}}$$
\section{Summary}
In this chapter, we have reviewed recent progress in several tasks that are core to understanding human opinions. We also survey on techniques aiming to learn embedding for words and concepts. All existing solutions have their own technical limits as well as unsolved issues. However, it lacks investigations on leveraging the concepts for specific tasks. In addition, the fusion of information from concept-level and other levels (e.g., word-level) has been missed in existing studies.

\chapter{Learning Word Embedding with Multi-task Entity-based Features}
\label{tag:chap3}
\setcounter{equation}{0}

\section{Introduction}
\label{sec:intro}
Named entity recognition (NER) has been studied and applied successfully to formal~\cite{ner:2007} and informal texts~\cite{chenliang2015} for many years. Nevertheless the adaptation of NER methods to conversational speech remains challenging due to, for example, case insensitivity, lack of punctuations, un-grammatical structure, repetition, and presence of disfluencies inherent to conversations. In addition, there is not much spoken data annotated with named entities to cover the huge variety of named entity instances likely occurring in speech, and simply increasing the amount of manual annotation is not realistic for reasons of cost, evolution of new spoken terms and diversity.

Several works on NER from spoken contents have already explored the use of external resources like online gazetteers~\cite{surdeanu:2005} and Wikipedia~\cite{Bechet:2010} to overcome the lack of annotations.
Gazetteers, for instance, have successfully boosted NER performance for given entities (e.g., Location), but do not convey the information related to the context words surrounding the entity names that are also important for NER.
Other lexical resources such as WordNet provide semantic relations like synonymy among common English words, but remain limited for names.
A second category of approaches~\cite{collobert:2008,turian:2010,yu:2013,guo:2014,glove:2014} tackling the sparseness of NER training data use unlabeled data to learn low dimensional vector representation of words, called word embeddings. Used either as continuous~\cite{collobert:2008,turian:2010,glove:2014} or discrete features~\cite{yu:2013,guo:2014}, word embeddings have been shown effective in improving the generalization of NER.

For the last two years, an increasing number of studies have suggested that injecting application-specific information into the neural networks used to train word embeddings can further improve the performance of down-stream applications, e.g., dependency parsing~\cite{bansal:2014}, semantic relation classification~\cite{hashimoto:2015}, antonym detection~\cite{masataka:2015}, spoken language understanding~\cite{tasos:2014}. The task-specific information used by these methods are injected into the training process of word embeddings mainly by expanding or replacing the input or output of the neural network. Passos {\it et al.}~\cite{passos:2014}, to our knowledge, is the only work trying to learn word embeddings for NER by leveraging lexicons related to named entities.  Our proposal is inspired by their approach and can be seen as its generalization in the sense that our word embedding is trained with multiple relevant tasks while their model is trained on only one additional task that predicts whether the current word belonging to a predefined set of semantic classes.

The key to learning word embeddings for the particular problem of NER is to encode entity-related information into the word embedding space. As the first step to investigate a tight fusion of concept-level\footnote{Named entity is a subclass of concepts.} and word-level information, we propose to augment the learning process of word embedding with entity-related features. More specifically, we introduce an NER framework including a modified word embedding method based on Skip-gram models~\cite{mikolov:distributed2013}, in which we generalize the training objective to integrate features explicitly designed for NER task and grouped by types. From a different point of view, we extend the training of Skip-gram model to a multi-tasks setting where each feature type corresponds a particular task. The multi-task learning of word embeddings allow the embedding to form a thorough representation of the word by incorporating information from different information sources. We prove the effectiveness of our method in the context of NER. Namely, we find that injecting such NER-specific information as part-of-speech tags, taxonomic relations, and self-training features~\cite{Bechet:2010} yields improvements over baseline (i.e., without using word embeddings), when evaluated against conversational speech transcripts (i.e., Switchboard Corpus). Furthermore, the results show that our proposed word embeddings also outperform the task-independent word embeddings.

\section{Methods}
\label{sec:format}
\begin{figure}[!t]
\includegraphics[width=\linewidth]{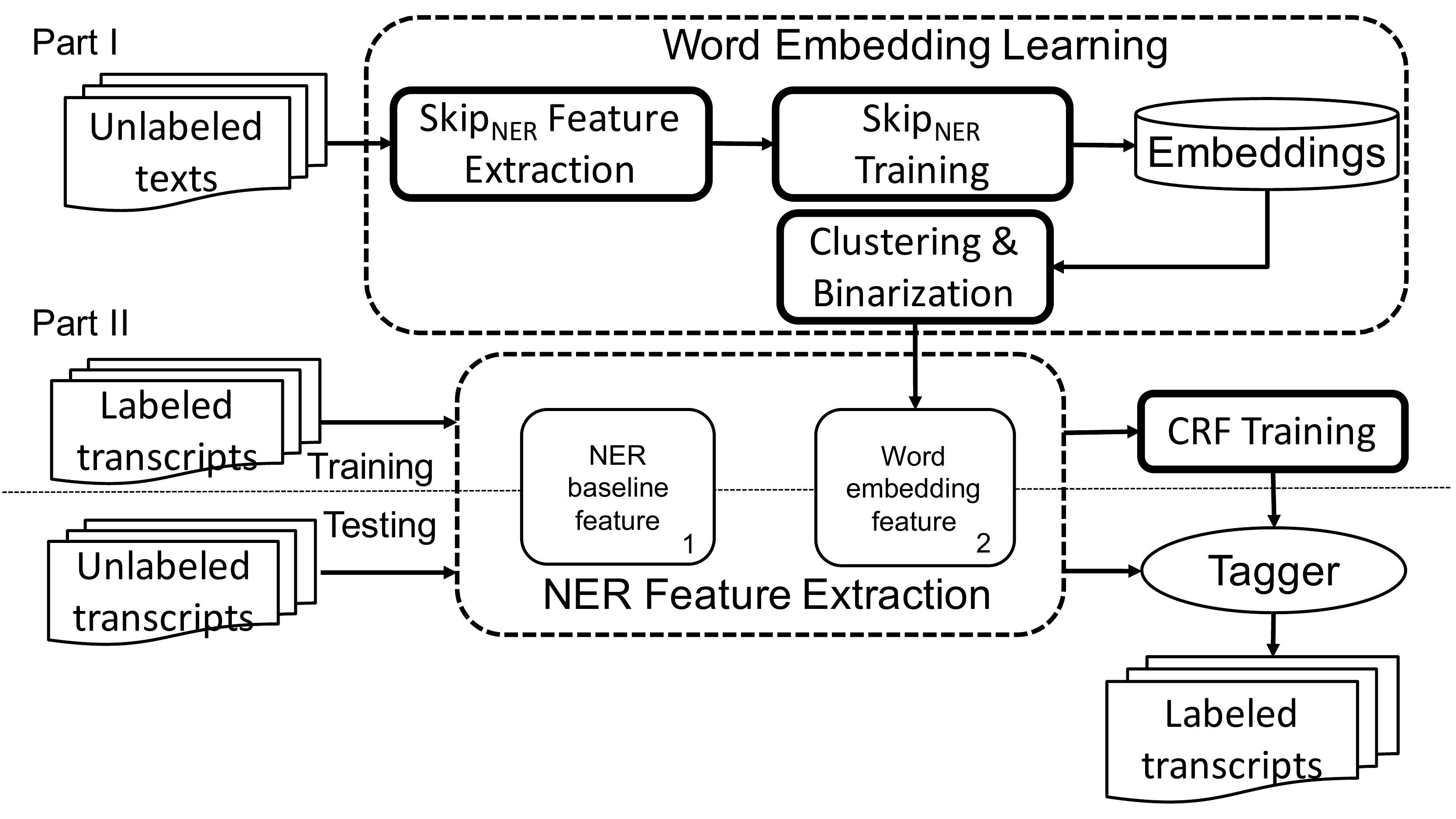} 
\caption{NER Framework with NER-specific word embeddings}
\label{fig:overview}
\end{figure}

The proposed NER framework illustrated in Figure~\ref{fig:overview} is composed of two parts: 1) a task-specific word embedding learning integrating features specific to NER (Part I in Figure~\ref{fig:overview}.), using not only unlabeled data but also resources like knowledge base and baseline NER tagger; and 2) a supervised model of linear-chain CRF trained with the NER features extracted from the labeled corpus and the resultant word embeddings (Part II in Figure~\ref{fig:overview}). We denote the proposed word embedding method as Skip$_{\text{NER}}$. 

\subsection{Baseline NER method}
\label{sec:conventional-features}

In this section, we describe the baseline NER method, a conventional linear-chain CRF with BIO encodings. 
The features extracted from every word $w_i$ of the training data and used to train the CRF are summarized in Table~\ref{tab:feat}. 
The feature set consists of n-grams, part-of-speech (POS) tags, affixes, and the BIO tag of the $i^{th}$ word (designated as $y_i$).
\begin{table}[!htb]
\centering
\begin{tabular}{|l l|}
\hline
\multicolumn{2}{|l|}{Context Features ($CF$)}\\
\hline
Unigram 	& $w_{i+k}$, \, $-2\leq k \leq 2$\\
Bigram 		& $w_{i+k} \wedge w_{i+k+1}$, \, $-2\leq k \leq 1 $\\
POS 			& $t_{i+k}$, \, $-2\leq k \leq 2$\\
POS bigram & $t_{i+k} \wedge t_{i+k+1}$, \, $-2\leq k \leq 1 $\\
Prefix 		& $Pre(w_{i+k},l)$,\, $-2\leq k \leq 2,\, 0 \leq l \leq 4$\\
Suffix 		& $Suf(w_{i+k},l)$,\, $-2\leq k \leq 2,\, 0 \leq l \leq 4$\\
\hline
\hline
Tag Feature& \\
\hline
Tag\&Context & $y_i \wedge c$	,\quad $c \in CF$\\
Tag Bigram & $y_{i-1}y_i$ \\
\hline
\end{tabular}
\caption{Templates of features used in the CRF baseline}
\label{tab:feat}
\end{table}
\subsection{Skip-gram model}

Skip-gram model is a neural language model that can be efficiently trained with corpus of billions of words. It assumes that the user would only be concerned about the quality of word embeddings but not the capability of language modeling. The objective of Skip-gram model is to find the representation of word that is useful for predicting the surrounding words instead of just next words. Each word is linked with a $d$-dimensional vector $v_w$, and each context $c$ is assigned a vector $v'_c$ which is also of $d$ dimensions, i.e., the objective function is to maximize the log probability
\[ \sum_{t=1}^{T}\sum_{j=-k}^{k}\log p(c_j=w_{t+j}|w_t)\]
where $k$ is size of context window which are a certain number of words surrounding the central word. The probability $p(c|w)$ is given as

\[p(c|w) =\frac{ \exp(v'^\top_{c}v_{w})}{\sum_{c'}\exp({v'^\top_{c'}v_{w})}}\]

However, as computing the gradient of $\log p(c|w)$ requires iteration over all pairs of $(w,c)$, optimizing this objective function is intractable when the size of context vocabulary is large. To approach this problem, ~\cite{ mikolov:distributed2013} uses negative sampling as an alternative training objective. The objective function can be rewritten as

\[\sum_{t=1}^{T}\sum_{j=-k}^{k}(\log \sigma(v'^T_{c}v_{w_t})+ \sum_k^N\mathbb{E}_{c '\sim P_{w}} (\log(\sigma (-v'^T_{c'}v_{w_t}))))\]
where $\sigma(x)=1/(1+exp(-x))$. For each $(w,c)$ pair appearing in the corpus, N pairs of $(w,c_k)$ are randomly sampled from a unigram distribution $P_w$ as negative training instances.

\subsection{Feature extraction for word embeddings}
\label{sec:skip-ner}
First of all, we extract four types of features (i.e., neighboring words, part of speech (POS), taxonomic, self-trained) from the unlabeled data to train word embeddings. In principal, they are mostly features used for training the baseline NER model, and one of the intuition of our model is  to learn a condensed version of these features from unlabelled data to achieve higher capability of generalization.

\textbf{Neighboring words and POS tags}: 
They are acknowledged to be efficient for NER~\cite{surdeanu:2005,Bechet:2010}. In fact, we use these features not only for training our word embeddings, but also for training the baseline model of NER (Section \ref{sec:conventional-features}). Their formal definitions can be found in the first 4 lines of Table~\ref{tab:feat}. In terms of feature type, we define two separate types for words and POS tags respectively.

\textbf{Taxonomic features}: The generalization of similar entities within a common category (or concept) in a taxonomy may be useful for NER to capture contexts shared by similar entities. We select 280 concepts with at least 100 instances from ConceptNet, which is a large, automatically created semantic graph including taxonomic relations. For each concept $g$, we add binary taxonomic features, which are defined as ${\text{TX}_g(w_{i+k}), 2 \leq k \leq 2}$, where $i$ is the index of current word, and $\text{TX}_g(w)$ indicates if the word $w$ belongs to the concept $g$ or not.

\textbf{Self-trained features}: As in~\cite{Qi:2014}, these features are generated by automatically labeling the training data using a baseline NER tagger (Section \ref{sec:conventional-features}).
We use the resultant named entities hypotheses as additional features for word embeddings.
The features are defined as $T_{i+k}$, with $T_{j}$ the NER label of the $j^{th}$ word.

\subsection{ Learning word embeddings from entity-based features}
\label{sec:feature-grouping}
Our proposal is based-on the Skip-gram model~\cite{mikolov:distributed2013}, a neural language model that can be efficiently trained on a large corpus of billions of words.
Its objective function is the sum of log probabilities $p(w_{i+j}|w_i)$ over the whole corpus,
$$ \sum_{i=1}^{N}\sum_{j=-k}^{k}\log p(w_{i+j}|w_i)$$
where ${w_{i+j}}$ indicates a neighbor word of $w_i$, $k$ the size of the context window and $N$ is the length of the dataset. And, the basic Skip-gram formulation defines $p(w_{i+j}|w_i)$ using the softmax function as follows:
$$p(w_{i+j}|w_i) =\frac{ \exp(v'^\top_{w_{i+j}}v_{w_i})}{\sum_{w \in W}\exp({v'^\top_{w}v_{w_i})}}$$
where $W$ is the vocabulary of words, $v'_{w_{i+j}}$ and $v_{w_i}$ are the embeddings of context word and current word, respectively.
We modify the Skip-gram model to predict the set of features $F(w_i)$ extracted for the given word $w_i$ at the $i^{th}$ position of a corpus. The objective function can be rewritten:
$$\sum_{i=1}^{N}\sum_{f \in F(w_i)} \log p(f|w_i)$$

To estimate $p(w_{i+j}|w_i)$, the basic Skip-gram model assumes a single distribution over all the words. However, for $p(f|w_{i})$, since the features we use for training application-specific word embeddings are heterogeneous, they may have different distributions. 
We thus split the whole set of features ($S$) into subsets ($S_X$), where $X$ indicates one of the four feature types aforementioned and the relative position to the center word. For example, the subset $S_{pos:-1}$ includes all the features that tell the POS tags of the previous word. We define $C_S(f)$ as the function that returns the subset of $S$ that contains the feature $f$.
We define the probability of extracting $f$ for a given word $w$ from the training data as follows:
$$p(f|w) =\frac{ \exp(v^\top_{f}v_{w})}{\sum_{f'\in C_S(f)}\exp({v^\top_{f'}v_{w})}}$$
where $v_{w}$ and $v_f$ are the vectors associated with center word $w$ and feature $f$ respectively, and both are the parameters to be learned. Note that, with the new definition, the objective function becomes a linear combination of the objective functions of multiple classifiers with equal weights.
We optimize this objective function using stochastic gradient descent and negative sampling method previously proposed for Skip-gram~\cite{mikolov:distributed2013}, which rewrites the objective function as: 
$$ \sum_{i=1}^{N}\sum_{f \in F(w_i)}\left(\log \sigma(v^T_{f}v_{w_i})+\sum_{f'\in Z(f)} \log\left(\sigma (-v^T_{f'}v_{w_i})\right)\right)$$
In this expression $\sigma(x)=1/(1+\text{exp}(-x))$, and the set of features $Z(f)$ is created by randomly selecting $n$ negative samples from a unigram distribution over features in $C_S(f)$

\subsection{Multi-task interpretation of the proposed method}
Since each type of features are heterogeneous to one another, we could also view the prediction of each feature type as performing a separate task. Therefore, the iteration over the set of feature type (i.e., $F$) can be seen as performing multi-task learning over a set of tasks. For example, predicting neighboring words and POS tags corresponds to the tasks of language modeling and POS tagging respectively.

\subsection{Using word embeddings as NER features}

\label{sec:word-embeddings-features-for-CRF}
We convert word embeddings to additional features of the CRF model as in~\cite{yu:2013,guo:2014} by binarizing vector elements of word embeddings, and clustering of words based on similarity of word embeddings. The vectors are binarized using the following rules, 

$$
D_{mn} = 
\begin{cases} 
  1    & \text{if } W_{mn} \geq \overline{W^+}_{m\cdot} \\
  -1    & \text{if } W_{mn} \leq \overline{W^-}_{m\cdot}\\
  0    & \text{else}
 \end{cases}
$$
where $W$ is the original word embedding matrix, and $D$ the binarized matrix, $\overline{W^{+}}_{m\cdot}$ the mean of all positive values of the $m^{th}$ dimension, and $\overline{W^-}_{m\cdot}$ the mean of all negative values. Only non-zero values are added to the feature set. $\overrightarrow{V_D}_w$ denotes the column of $D$ corresponding to the word $w$. Then we cluster all words based on Euclidean distance using K-Means~\cite{guo:2014,yu:2013}.
We use different numbers of clusters, $K$, to let the clustering reflect different levels of granularity. 
$C_K(w)$ is the cluster of the word $w$, where $K$ indicates the number of clusters in the K-Means (500$\leq$K$\leq$3000).
The features learned from word embeddings are summarized in Table~\ref{tab:featc}.
\begin{table}[!h]
\centering
\begin{tabular}{|L{5cm} L{5cm} L{2.5cm}| }
\hline
Binarized vector	& $\overrightarrow{V_D}_{w_{i+k}}$	& $-2\leq k \leq 2$\\[3ex]
\hline
Cluster (Unigram)& $C^K{w_{i+k}}$& $-2\leq k \leq 2$\\
Cluster (Bigram)	& $C^K_{w_{i+k}} \wedge C^K_{w_{i+k+1}}$ & $-2\leq k \leq 1$\\
Cluster (Disjunct)	& $C^K_{w_{i-1}} \wedge C^K_{w_{i+1}}$& \\
\hline
\end{tabular}
\caption{Features learned from word embeddings for NER}
\label{tab:featc}
\end{table}

\section{Experiments}
\label{sec:typestyle}

\subsection{Corpora}
\label{sec:data}
We have used the \textbf{ukWaC corpus}~\cite{ukwac} as unlabeled text dataset to train word embedding models. This is one of the largest English text corpus (about $1.8$ billion tokens), crawled from the Web on the \textit{co.uk} domains, and has been tagged with POS\footnote{http://www.cis.uni-muenchen.de/$\sim$schmid/tools/TreeTagger/}. 

The labeled open-conversation dataset is the Switchboard corpus~\cite{swb}. It is a large collection of 2-speaker conversational telephonic speech recordings.
We use the NER annotations of the corpus including four classes -- location (LOC), person (PER), organization (ORG) and miscellaneous (MISC), and the train/test partitions provided by Surdeanu et al.~\cite{surdeanu:2005}. 
Detailed numbers are presented in Table~\ref{tab:SWB}.

\begin{table}[!htb]
\centering
\begin{tabular}{|l||cccc|} 
\hline &LOC&PER&ORG & MISC \\\hline\hline
Train& 14,397&4,257&5,311&8,484\\\hline
Test& 631&342&296&687\\
\hline
\end{tabular}
\caption{Named entity distribution in Switchboard}
\label{tab:SWB}
\end{table}







We compare the proposed Skip-gram model (Skip$_{\text{NER}}$) with non-task specific word embeddings models previously used for NER. We retrained the word embeddings for Skip-gram, CBOW~\cite{mikolov:distributed2013}, and Glove~\cite{glove:2014} with the ukWaC corpus, using two-word context windows for each side of the current word.
We used the pre-trained word embeddings models\footnote{http://ai.stanford.edu/$\sim$ehhuang/, http://ronan.collobert.com/senna/} of HUANG~\cite{huang:2012} and SENNA~\cite{collobert:2008}, because their training is slow.
Table~\ref{tab:corpus} shows some statistics of the data for the models.
Following~\cite{turian:2010,collobert:2008,guo:2014,glove:2014}, the number of vector dimensions is set to 50 for all experiments. 


\begin{table}[!htb]
 \renewcommand{\arraystretch}{1} 
\centering
\begin{tabular}{|l||c|c|c|} \hline 
\textbf{Word embeddings} &\textbf{Tokens} &\textbf{Vocab.} &\textbf{Dataset}\\ \hline
Skip$_{\text{NER}}$, Skip-gram	& 1.8B	&	167K	&	ukWaC \\ 
 Glove, CBOW		& 1.8B	&	167K	&	ukWaC \\ 
HUANG						& 1.8B	&	100K	&	Wikipedia\\
SENNA						& 1.8B	&	130K	&	Wikipedia\\ \hline
\end{tabular}
\caption{Details on baseline Word Embeddings preparation}
\label{tab:corpus}
\end{table}

 
\subsection{Evaluations}
We report in Table \ref{tab:perform} the performance of the proposed NER method (Skip$_{\text{NER}}$) and the baseline method with the other word embeddings when tested against the manual transcripts of open-domain conversations. 
Our method outperforms all the other methods and achieves 2\% absolute improvement of F-score in comparison to the original Skip-gram model.



\begin{table}[!htb]
 \renewcommand{\arraystretch}{1} 
\centering
\begin{tabular}{|ll|ll|} 
\hline \textbf{System}&\textbf{\footnotesize F-score}&\textbf{System}&\textbf{\footnotesize F-score}\\
 \hline
\hline Baseline&66.51&& \\ 
Baseline + Skip&67.89 &Baseline + Huang &67.88\\
Baseline + CBOW &68.45&Baseline + Senna&68.29\\
Baseline + Glove &67.44&
Baseline + Skip$_{\text{NER}}$ &\textbf{70.19} \\
\hline
\end{tabular}
\caption{Performance of NER methods in terms of F-score}
\label{tab:perform}
\end{table}

\begin{figure}[!h]
\centering
\includegraphics[width=0.8\linewidth]{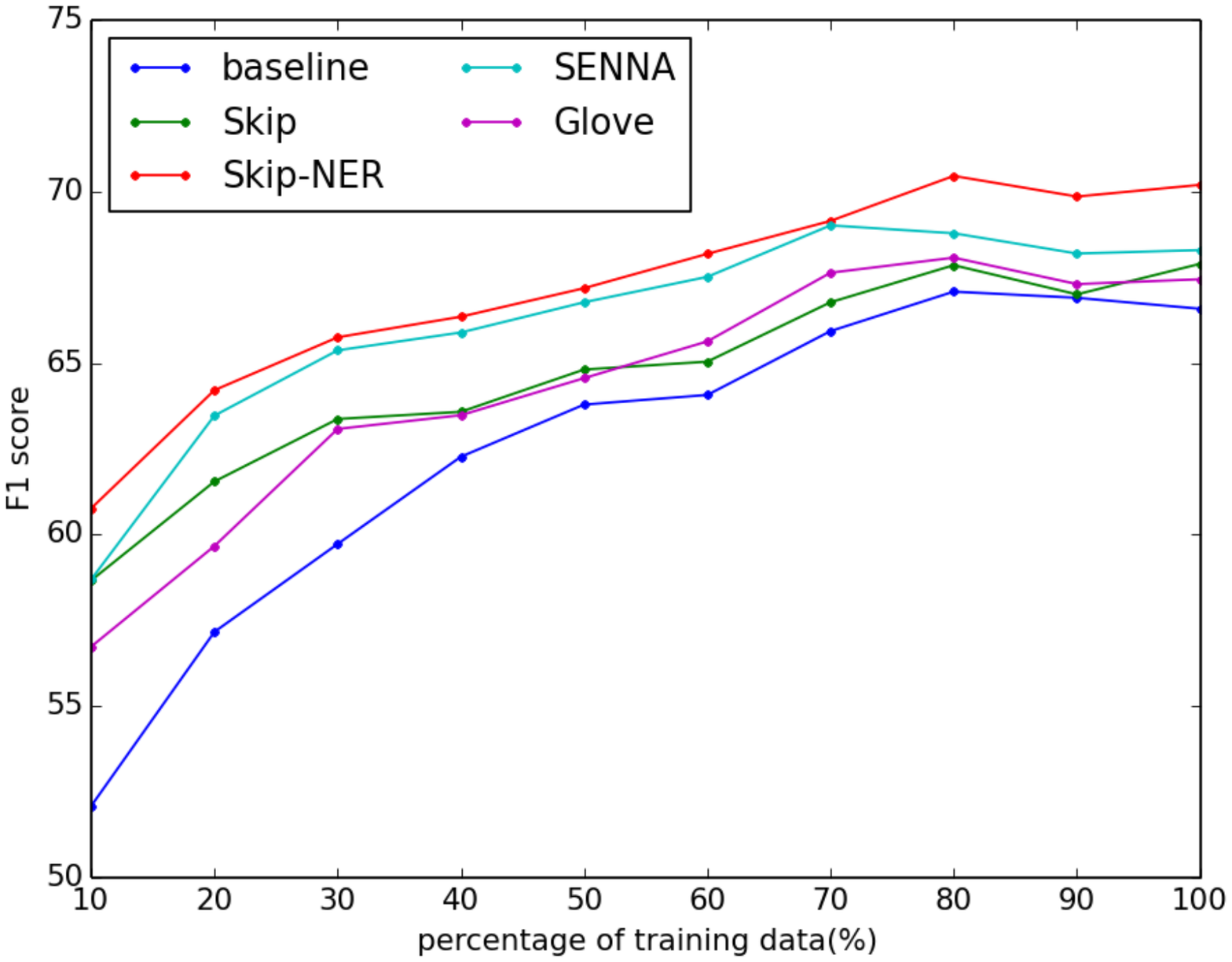}
\caption{Performance with varying size of NER training data}
\label{fig:perform}
\end{figure}

In the second experiment, we investigate the impact of the amount of labeled training data used to train the NER model.
As illustrated in Figure \ref{fig:perform}, the Skip$_{\text{NER}}$-based NER system always outperforms the other systems, independent of the amount of training data. 
Also, the NER systems with the generic (or non task-specific) word embeddings also perform better than the baseline. 
It suggests that the generalization brought by word embeddings consistently helps NER models in dealing with data sparsity. 
As more training data are used (80\%-100\% of training data), the gap of performance between all the baseline systems reduces. However, the gap between the baseline and Skip$_{\text{NER}}$ further increases even using more training data, which may mean that our method robustly handles data sparsity and extracts more discriminative information when more data are available.

We also study the impact of each subset of features used to train the Skip$_{\text{NER}}$ model.
As reported in Table \ref{tab:femb}, we find that using the whole set of features outperforms each of the feature subsets used alone. This result may indicate that the feature subsets are complementary to each other, thus supporting the proposed method of integrating NER-specific features for training word embeddings. Also, note that the Words only subset corresponds to the skip-gram model.
Location, and Person names are particularly well recognized by using the POS subset, while self-trained features and taxonomic relations seem more effective to discriminate the class ORG.

\begin{table}[!htb]
\centering
\begin{tabular}{|l||cccc|c|}
\hline
Skip$_{\text{NER}}$ Feature&PER&LOC& ORG&MISC & All\\\hline
Words			&	$80.6$	&	$79.3$	&	$51.7$	&	$ 57.0$				&	$68.5$\\
POS				&	$\mathbf{82.3 }$	&	$80.6$	&	$51.1$	&	$57.8$		&	$69.4$\\
self-trained	&	$80.8$	&	$79.8$	&	$53.6$	&	$57.2$					&	$68.9$\\
taxonomic		&	$81.0$	&	$80.0$	&	$54.0$	&	$57.4$					&	$69.3$ \\\hline
All 				&$81.7$	&$\mathbf{80.9}$	&	$\mathbf{55.7}$	&$\mathbf{57.9}$	&	 $\mathbf{70.2}$\\
\hline
\end{tabular}
\caption{Impact of feature subsets on Skip$_{\text{NER}}$~\cite{ma2016feature}}
\label{tab:femb}
\end{table}

\begin{table}[!htb]
\centering
\begin{tabular}{l||c|c|c}
\hline
Word							&	texas 			& cowboys			&	batman		\\\hline
								&	lousiana			&	texans		&	superman	\\
								&	spokane			&	cowboy		&	superhero	\\
Skip-gram					&	kansas			&	bandits		&	remake		\\
								&	sacramento	&	cowgirls		&	spiderman	\\
								&	biloxi				& 	impersonators	&	catwoman	\\\hline
								&	kentucky		&	cheerleaders	&	superman	\\
								&	kansas			&	texans		&	shrek			\\
Skip$_{\text{NER}}$	&	lousiana			&	yankees		&	starsky		\\
								&	florida			&	redskins		&	scooby-doo\\
								&	minnesota		&	broncos		&	spiderman	\\\hline
\end{tabular}
\caption{Top similar words returned by Skip-gram \& Skip$_{\text{NER}}$}
\label{tab:most}
\end{table}

Table \ref{tab:most} lists the five words most similar to each of three example words (i.e., cowboys, texas, batman), which are computed using the Skip-gram and Skip$_{\text{NER}}$.
For instance of the keyword \textit{texas}, both Skip$_{\text{NER}}$ and Skip-gram models return locations in the United States (US), but Skip$_{\text{NER}}$ seems to produce more `relevant' results than Skip-gram since \textit{texas} and all its five most similar words (e.g., \textit{kentucky}) are the names of states in US, while the results of Skip-gram include city names (e.g., \textit{spokane}, \textit{sacramento}), thus of different granularity.
For ambiguous words, Skip$_{\text{NER}}$ seems to focus on the semantics of the words related to the NER task. For instance of the keyword \textit{cowboys}, it may be the plural form of the noun `cowboy' or may indicate the American football team ``Dallas Cowboys''. While its similar words from Skip-gram reflect the ambiguity, all the top results of Skip$_{\text{NER}}$ are related to the latter meaning of the keyword. We observe a similar difference in the results for the keyword \textit{batman}.
These observations suggest Skip$_{\text{NER}}$ is able to capture more discriminative information for NER compared to generic word embeddings.

\subsection{Summary}
In this work, we propose a novel method to train task-specific word embeddings for NER by incorporating NER related multi-task learning. Through several experiments, we have shown that, on the manual transcripts of open-domain conversations, how our proposed feature-enriched word embeddings can out- perform baseline NER method and systems using task inde- pendent word embeddings.


\chapter{Label Embedding for Fine-grained Named Entity Typing}
\label{tag:chap4}
\setcounter{equation}{0}

\section{Introduction}
\label{intro}
Named entity typing (NET) is the task of inferring types of named entity mentions in text. NET is a useful pre-processing step for many NLP tasks, e.g., auto-categorization and sentiment analysis. Named entity linking, for instance, can use NET to refine entity candidates of a given mention~\cite{ling:2012}. Besides, NET is capable of supporting applications based on a deeper understanding of natural language, e.g., knowledge completion~\cite{dong:2014} and question answering~\cite{lin:2012,fader:2014}. Standard NET approaches or sometime known as named entity recognition~\cite{muc7,conll:2003,ace:2004} are concerned with coarse-grained types (e.g., person, location, organization) that are flat in structure. In comparison, fine-grained named entity typing (FNET)~\cite{ling:2012}, which has been studied as an extension of standard NET task, uses a tree-structured taxonomy including not only coarse-grained types but also fine-grained types of named entities. For instance, given ``[\textit{Intel}]\textit{ said that over the past decade}", standard NET only classifies \textit{Intel} as \textit{organization}, whereas FNET further classifies it as \textit{organization/corporation}. 

FNET is faced with two major challenges: growing type set and label noises. Since the type hierarchy of entities is typically built from knowledge bases such as DBpedia, which is regularly updated with new types (especially fine-grained types) and entities, it is natural to assume that the type hierarchy is growing rather than fixed over time. However, current FNET systems are impeded from handling a growing type set for that information learned from training set cannot be transferred to unseen types. Another problem with FNET is that the weakly supervised tagging process used for automatically generating labeled data inevitably introduces label noises. Current solutions rely on heuristic rules~\cite{gillick:2014} or embedding method~\cite{ren:2016} to remove noises prior to training the multi-label classifier. In order to address these two problems at the same time, we propose a simple yet effective method for learning prototype-driven label embeddings that works for both seen and unseen types and is robust to the label noises. The label embeddings, which might also deemed as concept embeddings, are inferred from a pre-trained word embedding space via a set of prototypes. From the perspective of fusing information from multiple levels, we made an attempt to fuse word-level information (i.e., word embedding of prototypes) to represent the canonical form of concepts (i.e., concept types). Another contribution of this work is that we combine prototypical and hierarchical information for learning label embeddings.

The remainder of this chapter is organized as follows: Section~\ref{relwk} proposes a survey of prior works related to FNET; Section~\ref{background} introduces the embedding-based FNET method and its zero-shot extension; Section~\ref{labelembd}
describes our label embedding method; Section~\ref{experiment} illustrates experiments and analysis for both few-shot and zero-shot settings; finally, Section~\ref{conclusion} concludes the paper and discusses future work. 

\section{Background}
\label{relwk}
There is little related work specifically on zero-shot FNET but several research lines are considered related to this work: fine-grained named entity recognition, prototype-driven learning, and multi-label classification models based on embeddings.
As FNET works with a much larger type set as compared with standard NET, it becomes difficult to have sufficient training data for every entity type if relying on only manual annotation. Fortunately, semi-structural data such as Wikipedia pages~\cite{ling:2012} turns out to be a good source for automatically generating training data.This auto-annotating practice has been followed by later work on FNET~\cite{hyena:2012,yogatama:2015,ren:2016}. However, since the automated tagging process is not perfect all the time, a number of label noises are propagated to supervised training and have consequently a negative impact on the classification performance.

The starting point of this work is the embedding method, WSABIE~\cite{wasbie:2011}, which have been adapted by~\cite{yogatama:2015} to FNET. WSABIE maps input features and labels to a joint space, where information is shared among correlated labels. However, the joint embedding method stills suffers from label noises. In addition, WSABIE fails to handle unseen types for it uses labeled training set as the only source learning label embeddings. 

In addition to WSABIE, this work is also inspired by recent progress on image annotation based on embedding methdos. For example, DeViSE~\cite{devise:2013} is proposed for annotating image with words or phrases, where embedding of single words, e.g., fruit, that that are pre-learned from textual data using Skip-gram word embeddings (see Section~\ref{sec:skipgram} for more details)~\cite{mikolov:distributed2013} are directly used for representing labels. In addition to label itself, existing work has also attempted to derive label embeddings from side information such as attributes~\cite{akata:2013}, manually-written descriptions~\cite{larochelle:2008}, taxonomy of types~\cite{weinberger:2009,akata:2013,akata:2015}, and so on and so forth.

Another related line of research is prototype-driven learning. In~\cite{Haghighi:2006}, a sequence labeling model uses prototypes as features and is applied to NLP tasks such as part-of-speech (POS) tagging. Prototype-based features~\cite{guo:2014} are then adapted for coarse-grained named entity recognition task. Even though we select prototypes in the same way as~\cite{guo:2014}, we use prototypes in a very different manner: we consider prototypes as the basis for representing labels, whereas prototypes are mainly used as additional features in prior works~\cite{Haghighi:2006,guo:2014}. In other words, prototypes are previously used on the input side, while we use them on the label side.

\section{Embedding Methods for FNET}
\label{background}
In this section, we introduce the embedding method for FNET proposed by~\cite{yogatama:2015} and its extension to zero-shot entity typing.
\subsection{Joint Embedding Model}
Each entity mention $m$ is represented as a feature vector $x\in \mathbb{R}^{V}$; and each label $y \in Y$ is a one-hot vector, where $Y$ is the set of true labels associated with $x$. $\bar{Y}$ denotes the set of false labels of the given entity mention. 
The bi-linear scoring function for a given pair of $x$ and $y$ is defined as follows:
$$f(x,y,W) = x'W y$$

where $W \in \mathbb{R}^{M\times N}$ matrix with $M$ the dimension of feature vector and $N$ the number of types.\\


Instead of using a single compatibility matrix, WSABIE~\cite{wasbie:2011,yogatama:2015} considers an alternate low-rank decomposition of $W$, i.e., $W=A^{\top}B$, in order to reduce the number of parameters. WSABIE rewrites the scoring function as
$$ f(x,y,A,B) = \phi(x,A)\cdot \theta(y,B) = x'A^{\top}By$$

,which maps feature vector $x$ and label vector $y$ to a joint space. Note that it actually defines feature embeddings and label embeddings as 
\begin{align*}
 &\phi(x,A):x \rightarrow Ax,\\
&\theta(y,B): y\rightarrow By,
\end{align*}

where $A \in \mathbb{R}^{D\times M} $ and $B \in \mathbb{R}^{D\times N}$ are matrices corresponding to lookup tables of feature embeddings and label embeddings, respectively. The embedding matrices $A$ and $B$ are the only parameters to be learned from supervised training process. In~\cite{wasbie:2011}, the learning is formulated as a learning-to-rank problem using weighted approximate-rank pairwise (WARP) loss, 
$$\sum_{y\in Y}\sum_{y'\in\bar{Y}}L(rank(x,y))\max(1-f(x,y,A,B)+f(x,y',A,B),0),$$

where the ranking function $rank(x,y) = \sum_{y'\in\bar{Y}}\mathbb{I}(1+f(x,y',A,B)>f(x,y,A,B))$, and $L(k)=\sum_{i=1}^{k} \frac{1}{i}$ which maps the ranking to a floating-point weight.

\subsection{Zero-shot FNET Extension}
\label{bg:zero}
A zero-shot extension of above WSABIE method can be done by introducing pre-trained label embeddings into the framework. The pre-trained label embeddings are learned from additional resources, e.g., text corpora, to encode semantic relation and dependency between labels. Similar to~\cite{akata:2013}, we use two different methods for incorporating pre-trained label embeddings. The first one is to fully trust pre-trained label embeddings. Namely, we fix $B$ as the pre-trained $\tilde{B}$ and only learn $A$ in an iterative process.
The second method is to use pre-trained label embedding as prior knowledge while adjusting both $A$ and $B$ according to the labeled data, i.e., adding a regularizer to the WARP loss function,
$$\sum_{y\in Y}\sum_{y'\in\bar{Y}}L(rank(x,y))\max(1-f(x,y,A,B)+f(x,y',A,B),0)+\lambda||B-\tilde{B}||_F^2,$$

where $||\cdot||_F$ is the Frobenius norm, and $\lambda$ is the trade-off parameter.

\section{Methods}
\label{labelembd}

\subsection{Prototype-driven Label Embedding}
Joint embedding methods such as WSABIE learn label embeddings from the whole training set including noisy labeled instances resulting from weak supervision. It is inevitable that the resulting label embeddings are affected by noisy labels and fail to accurately capture the semantic correlation between types. Another issue is that zero-shot frameworks such as DeViSE are not directly applicable to FNET as conceptually complex types, e.g., \textit{GPE} (Geo-political Entity) cannot be simply mapped to a single natural word or phrase. 

To address this issue, we propose a simple yet effective solution which is referred to as prototype-driven label embedding (ProtoLE), and henceforth we use $\tilde{B}^P$ to denote the label embedding matrix learned by ProtoLE.
The first step is to learn a set of prototypes for each type in the type set. ProtoLE does not fully rely on training data to generate label embeddings. Instead, it selects a subset of entity mentions as the prototypes of each type. These prototypes are less ambiguous and noisy compared to the rest of the full set. 

Even though it is already far less labor-intensive to manually select prototypes than annotating entity mentions one by one, we consider an alternative automated process using Normalized Point-wise Mutual Information (NPMI) as the particular criterion for prototype selection. The NPMI between a label and an entity mention is computed as:
$$\mathtt{NPMI}(y,m) = \frac{\mathtt{PMI}(y,m)}{-\ln p(y,m)},$$

where $\mathtt{NPMI}(\cdot,\cdot)$ is the point-wise mutual information computed as follows:
$$\mathtt{PMI}(y,m) = \log{\frac{p(y,m)}{p(y)p(m)}},$$

where $p(y)$, $p(m)$ and $p(y,m)$ are the probability of entity mention $m$, label $y$ and their joint probability. For each label, NPMI is computed for all the entity mentions and only a list of top $k$ mentions are selected as prototypes. Note that NPMI is not applicable to unseen labels. In such case, it is necessary to combine manual selection and NPMI.

Word embeddings methods such as Skip-gram model~\cite{mikolov:distributed2013} are shown capable of learning distributional semantics of words from unlabeled text corpora. To further avoid affected by label noises, we use pre-trained word embeddings as the source to compute prototype-driven label embeddings. For each label $y_i$, we compute its label embedding as the average of pre-trained word embeddings of the head words of prototypes, i.e.,
	$$\tilde{B}^P_i = \frac{1}{k}\sum_{j=1}^k {v_{m_{ik}}},$$

where $v_{m_{ik}}$ denotes the word embedding of $k$th word in the prototype list of label $y_i$. In the case of using phrase embeddings, the full strings of multi-word prototypes could be used directly. 
\subsection{Hierarchical Label Embedding}
Another side information that is available for generating label embeddings is the label hierarchy. We adapt the Hierarchical Label Embeddings (HLE)~\cite{akata:2013} to FNET task. Unlike~\cite{akata:2013}, which uses the WordNet hierarchy, FNET systems typically have direct access to predefined tree hierarchy of type set. We denote the label embedding matrix resulting from label hierarchy as $\tilde{B}^H$. Each row in $\tilde{B}^H$ corresponds to a binary label embedding and has a dimension equal to the size of label set. For each label, the sets $\tilde{B}^H_{ij}$ to 1 when $y_j$ is the parent of $y_i$ or $i=j$, and 0 to the remainder,
	\[\tilde{B}^H_{ij} = 
 \begin{cases} 
  1 & \text{if } i=j \text{ or } y_j \in Parent(y_i) \\
  0    & \text{otherwise } 
 \end{cases}.\]
HLE explicitly encodes the hierarchical dependency between labels by scoring a type $y_i$ given $m$ using not only $y_i$ but also its parent type $Parent(y_i)$. The underlying intuition is that recognition of a child type should be also based on the recognition of its parent. 
\subsection{Prototype-driven Hierarchical Label Embedding}
One shortcoming of HLE is that it is too sparse. A natural solution is combining HLE with ProtoLE, which is denoted as Proto-HLE. Since $\tilde{B}^H \in \mathbb{R}^{N\times N}$ and $\tilde{B}^P \in \mathbb{R}^{D\times N}$, the combined embedding matrix $\tilde{B}^{HP}$ can be obtained by simply multiplying $\tilde{B}^H$ by $\tilde{B}^P$, i.e.,
	$$\tilde{B}^{HP} = \tilde{B}^P \tilde{B}^{H\top}.$$
Note that $\tilde{B}^{HP}$ has the same shape as $\tilde{B}^P$, and it is actually representing the child label as a linear combination of the ProtoLE vectors of its parent and itself. 

\subsection{Type Inference}
Having computed the scoring function for each label given a feature vector of the mention, we conduct type inference to refine the top $k$ type candidates. In the setting of few-shots FNET, $k$ is typically set to the maximum depth of type hierarchy, while different values for $k$ may be used for a better prediction of unseen labels in zero-shot typing. 
For top $k$ type candidates, we greedily remove the labels that conflict with others. However, unlike~\cite{yogatama:2015}, we use a relative threshold $t$ to decide whether the selected type should remain in the final results, which is more consistent with the margin-infused objective function than a global threshold. Namely, a type candidate will be passed to type inference only if the difference of score from the 1-best is less than a threshold. 
\label{infer}

\section{Experiments}
\label{experiment}
\subsection{Experiment Setup}

Our method uses feature templates similar to what have been used by state-of-the-art FNET methods~\cite{ling:2012,gillick:2014,yogatama:2015,xiang:2015}. Table~\ref{tab:features} illustrates the full set of feature templates used in this work. We evaluate the performance of our methods on three benchmark datasets that have been used for the FNET task: BBN dataset~\cite{bbn}, OntoNotes dataset~\cite{onto} and Wikipedia dataset~\cite{ling:2012}. Xiang Ren et al.~\cite{xiang:2015} has pre-processed the training sets of BBN and OntoNotes using DBpedia Spotlight$^1$\url{1}{github.com/dbpedia-spotlight/dbpedia-spotlight}. Entity mentions in the training set are automatically linked to a named entity in Freebase and assigned with the Freebase types of induced named entity. As shown in Table \ref{tab:dataset4}, BBN dataset contains 2.3K news articles of Wall Street Journal, which includes 109K entity mentions belonging to 47 types. OntoNotes contains 13.1K news articles and 223.3K entity mentions belonging to 89 entity types. The size of Wikipedia dataset is much larger than the other two with 2.69M entity mentions of 113 types extracted from 780.5K Wikipedia articles. Each dataset has a test set that is manually annotated for purpose of evaluation. To tune parameters such as the type inference threshold $t$ and trade-off parameter $\lambda$, we randomly sample 10\% instances from each testing set as the development sets and use the rest as evaluation sets.

\begin{table}[!h]
\small
\centering
\begin{tabular}{|l|l|} 
\hline
 Feature&Description\\
\hline
Tokens & Unigram words in the mentions\\

Head& Head word of the mention\\
Cluster& Brown Cluster IDs of the head word \\
POS Tag& POS tag of the mention \\
Character&Lower-cased character trigrams in the head word\\
Word Shape & The word shape of words in the mention\\
Context &Unigram/bigram words in context of the mention\\
Dependency& Dependency relations involving the head word\\
\hline
\end{tabular}
\caption{Feature Template}
\label{tab:features}
\end{table}

\begin{table}[!h]
\centering
\begin{tabular}{|l|l|l|l|l|l|} 
\hline
\multicolumn{2}{|c|}{Dataset} &Types&Documents&Sentences&Mentions\\
\hline
\multirow{2}{*}{BBN}&train & \multirow{2}{*}{47}&2.3K&48.8K&109K\\
&test&&459&6.4K&13.8K\\
\hline
\multirow{2}{*}{OntoNotes}&train&\multirow{2}{*}{89}&13.1K&147.7K&223.3K\\
&test&&76&1.3K&9.6K\\
\hline
\multirow{2}{*}{Wikipedia}&train&\multirow{2}{*}{113}&780.5K&1.15M&2.69M\\
&test&&-&434&563\\
\hline
\end{tabular}
\caption{Statistics of datasets}
\label{tab:dataset4}
\end{table}

\begin{figure}[!htb]
	\centering     \includegraphics[width=0.8\linewidth]{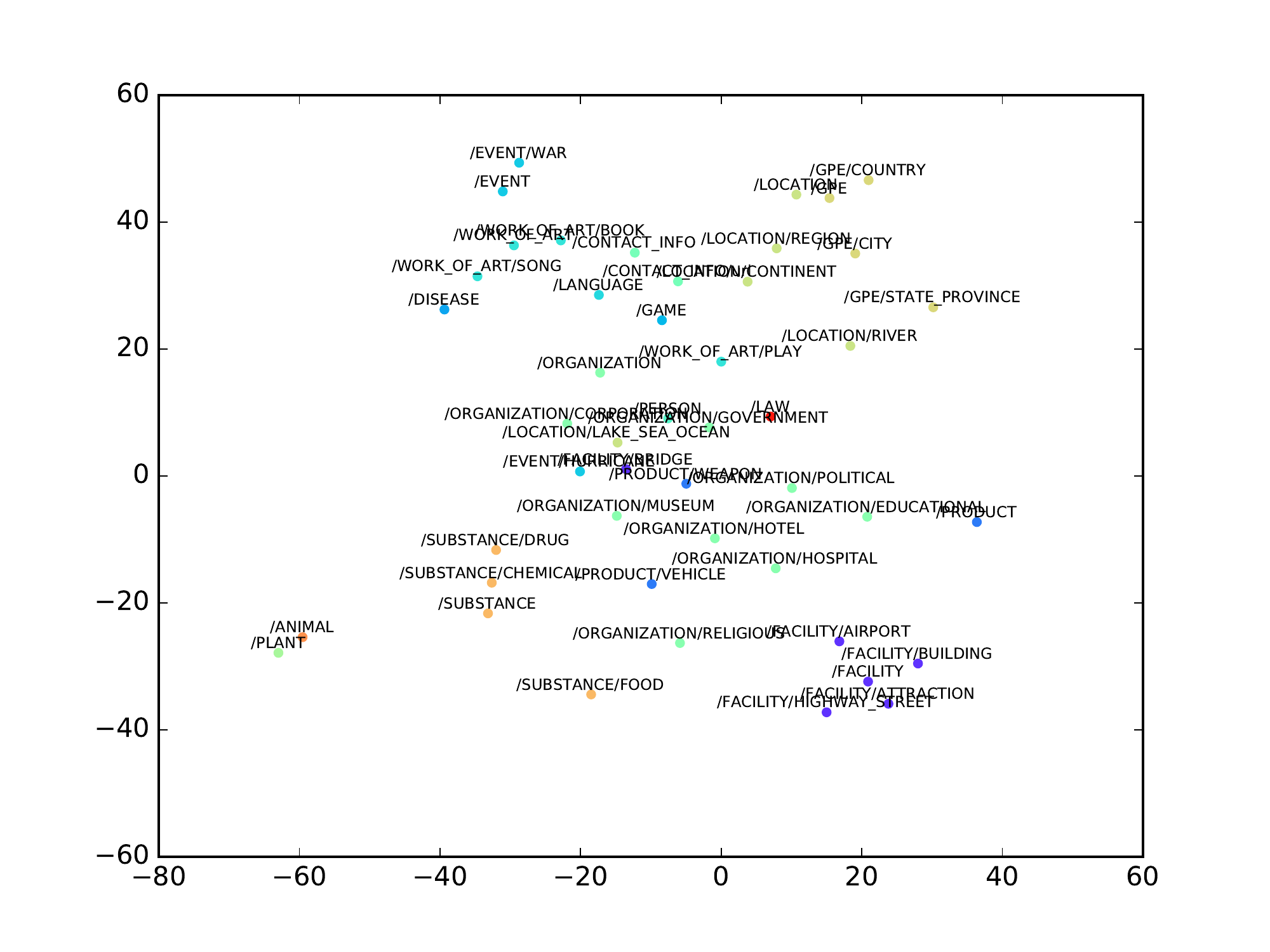}
    \caption{t-SNE visualization of the prototype-driven label embeddings for BBN dataset~\cite{maalab}}
    \label{fig:t-sne}
\end{figure}

Following prior works~\cite{ling:2012}, we evaluate our methods and baseline systems using both loose and strict metrics, i.e., Macro-F1, Micro-F1, and strict Accuracy (Acc.). Given the evaluation set $D$, we denote $Y_m$ as the ground truth types for entity mention $m\in D$ and $\widehat{Y}_m$ as the predicted labels. Strict accuracy (Acc) can be computed as:Acc$ = \frac{1}{D}\sum_{m\in D} \sigma(Y_m=\widehat{Y}_m)$, where $\sigma(\cdot)$ is an indicator function. Macro-F1 is based on Macro-Precision (Ma-P) and Micro-Recall (Ma-R), where Ma-P $= \frac{1}{|D|}\sum_{m\in D}\frac{|Y_m\cap \widehat{Y}_m|}{Y_m}$, and Ma-R$ =\frac{1}{|D|} \sum_{m\in D}\frac{|Y_m\cap \widehat{Y}_m|}{Y_m}$. And Micro-F1 is based on Micro-Precision (Mi-P) and Micro-Recall (Mi-R), where Mi-P$=\frac{\sum_{m\in D}|Y_m\cap \widehat{Y}_m|}{\sum_{m\in D}\widehat{Y}_m}$, and Mi-R$=\frac{\sum_{m\in D}|Y_m\cap \widehat{Y}_m|}{\sum_{m\in D}Y_m}$. 

\subsection{Generating ProtoLE}
Our ProtoLE embeddings use Continuous-Bag-of-Words (CBOW) word embedding model~\cite{mikolov:distributed2013} trained on Wikipedia dump using a window of 2 words to both directions. We use 300 dimensions for all embedding methods except HLE. Table \ref{tab:prototypes} illustrates examples of prototypes learned for types in BBN dataset. It can be observed that most of the top-ranked mentions are correctly linked to types, even though there are still some noises, e.g., 
north\_american for \texttt{/LOCATION/CONTINENT}. It also shows that prototypes of related types such as \texttt{/LOCATION} and \texttt{/GPE} are also semantically related. Figure\ref{fig:t-sne} visualizes the prototype-driven label embeddings for BBN dataset using -Distributed Stochastic Neighbor Embedding (t-SNE)~\cite{t-sne}. It can be easily observed that semantic related types are close to each other in the new space, which proves that prototype-driven label embeddings can capture the semantic correlation between labels. 

\begin{table}[h]
\small
\centering
\begin{tabular}{|l|l|l|} 

\hline
Type &Prototypes\\
\hline
/LOCATION & areas connaught earth lane brooklyn\\
/LOCATION/CONTINENT & north\_america europe africa north\_american asia\\
/LOCATION/LAKE\_SEA\_OCEAN& big\_bear lake\_erie champ lake\_geneva fujisawa\\
/LOCATION/RIVER& hudson thompson mississippi\_river james\_river tana\\
/GPE &soviet edisto canada china france\\
/GPE/STATE\_PROVINCE& california texas ohio arizona jersey\\
\hline
\end{tabular}
\caption{Example prototypes learned by PMI for types in BBN dataset }
\label{tab:prototypes}
\end{table}

\begin{figure}
\centering
  \begin{subfigure}
	\centering
    \includegraphics[width=0.8\linewidth]{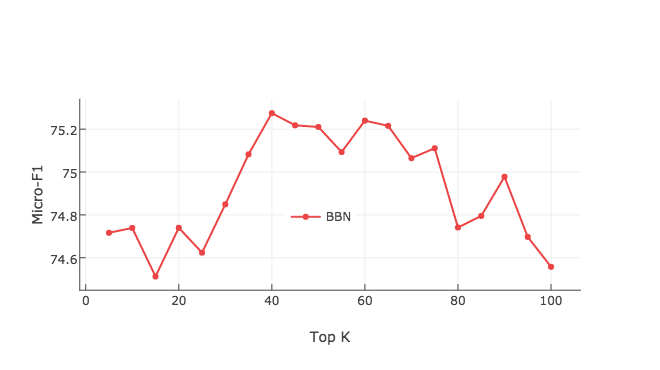}
    \caption{BNN Dataset}
    \label{fig:bnn}
  \end{subfigure}
  ~ 
  \begin{subfigure}
	\centering

    \includegraphics[width=0.8\linewidth]{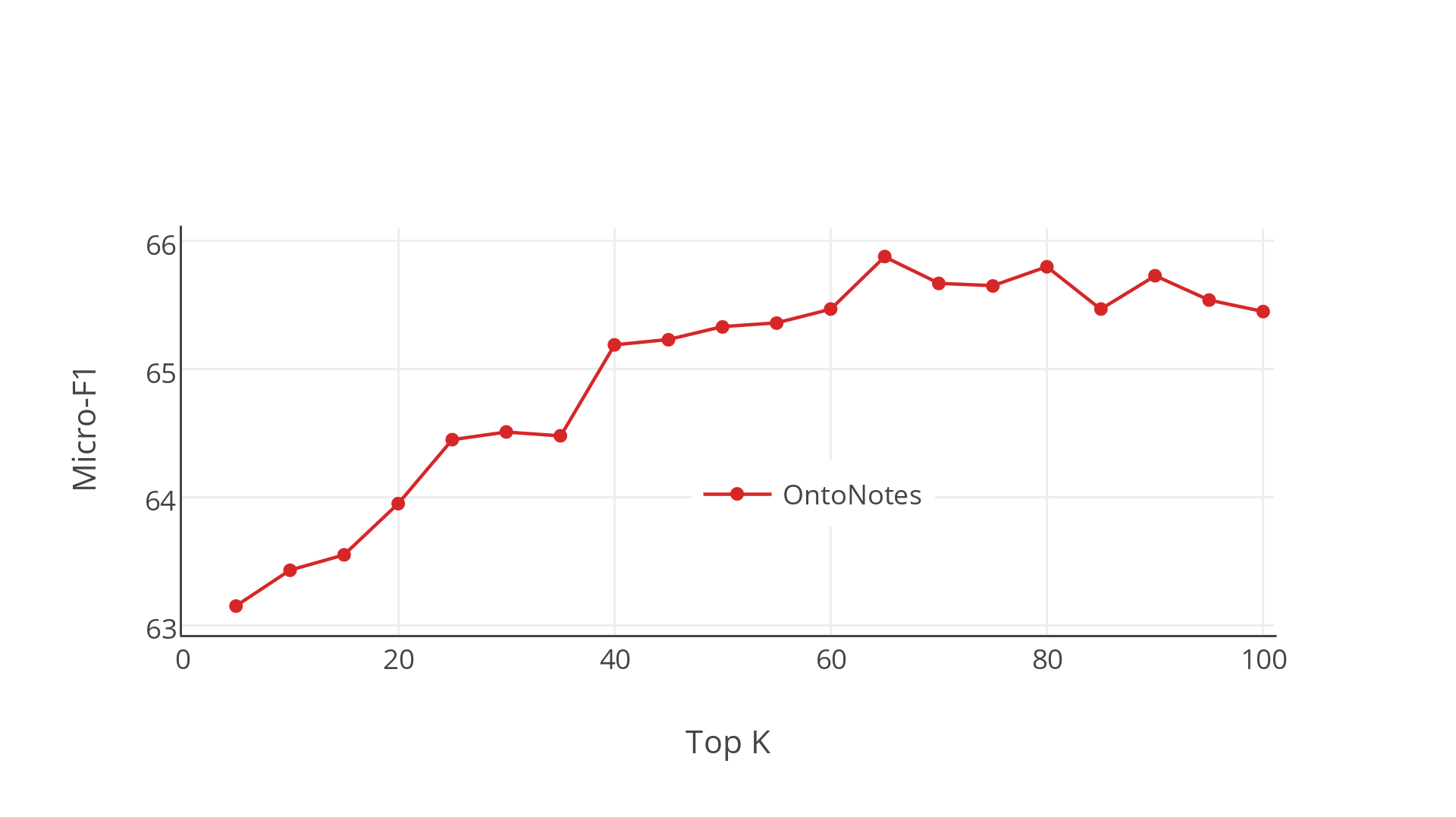}
    \caption{OntoNotes Dataset}
    \label{fig:onto}
  \end{subfigure}
   \begin{subfigure}
            	\centering

    \includegraphics[width=0.8\linewidth]{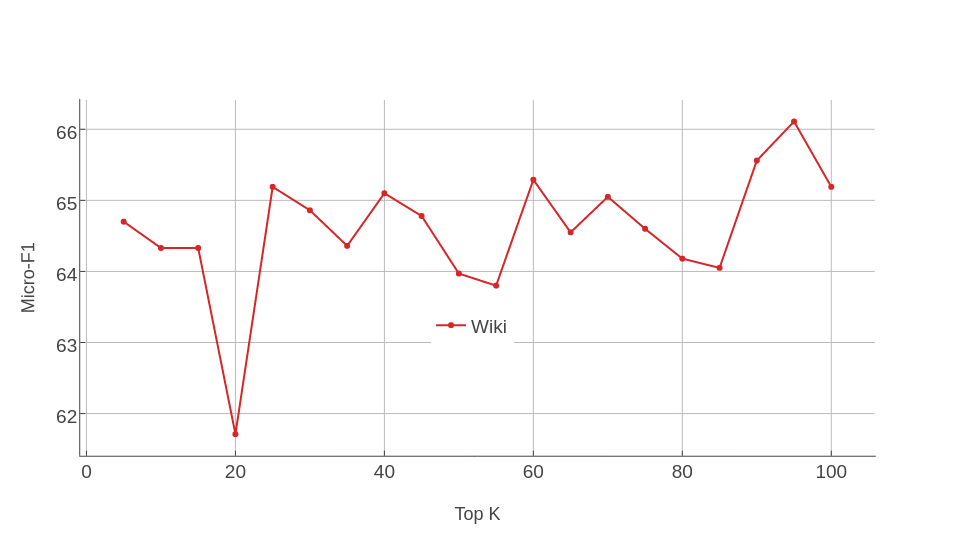}
    \caption{OntoNotes Dataset}
    \label{fig:wiki}
  \end{subfigure}
   
\caption{Performance changes on the development set with regard to the sizes of prototype list~\cite{maalab}}
\label{fig:topk}
\end{figure}

Figure~\ref{fig:topk} shows the Micro-F1 score of FNET with regard to the number of PMI prototypes used by ProtoLE. It shows that the Micro-F1 score does not change significantly on BBN and Wikipedia dataset, whereas using fewer prototypes per type ($\leq 40$) results in a drop of Micro-F1. Since the definitions of several types, especially the coarse-grained types, are actually very general, it may introduce bias into the label embeddings if using too few prototypes. We use $K=60$ for all our experiments for that it achieves decent performance on all three datasets.

\subsection{Few-shots Fine-grained Entity Typing}
In this section, we compare performances of FNET methods in the setting of few-shots FNET where the training set covers all types. Methods compared in this section are trained using the entire type set. We use evaluation metrics for our experiments: macro-F1, micro-F1 and accuracy. As in section \ref{bg:zero}, we train our label embeddings in two different ways: 1) non-adaptive training where label embeddings are fixed during training; and 2) adaptive training where label embeddings are also updated. Table \ref{tab:few-shots} shows the comparison with the state-of-the-art FNET methods: FIGER~\cite{ling:2012}, HYENA~\cite{hyena:2012} and WSABIE~\cite{yogatama:2015}. We make several findings from the results. 

Firstly, embedding methods with WARP loss function consistently outperform non-embedding methods (i.e., FIGER and HYENA) on all three datasets. The performance gaps are huge for BBN and OntoNotes, where the best embedding method achieves 10\%-20\% absolute improvement over the best non-embedding method (FIGER). However, the gap is much smaller on Wikipedia dataset whose size is significantly larger than the other two. 

Secondly, non-adaptive embedding methods always outperform their adaptive versions except HLE on Wikipedia dataset. Performance of adaptive label embeddings are all close to WSABIE, which suggests that adaptive label embeddings might suffer from same label noise problem as WSABIE does.\\

Thirdly, our ProtoLE and its combination with HLE consistently outperform both non-embedding and embedding baselines. Using the prototype information and non-adaptive framework results in absolute 3\%-5\% improvement with both loose and strict evaluation metrics. Non-adaptive HLE performs poorer than other embedding methods, which is most likely due to its sparsity in representing labels. However, Proto-HLE performs very close to ProtoLE on BBN and Wiki, while it improves all three measures by another absolute $\approx$2.5\% on OntoNotes.

 \begin{table}[!htb]\small
 \centering
 \begin{tabular}{|c|c|c|c|c|}
 \hline
\multirow{2}{*}{Method} &\multirow{2}{*}{Adapt}&\multicolumn{3}{c|}{BBN}\\
\cline{3-5}
&&Ma-F1&Mi-F1&Acc.\\
\hline
FIGER&NA&67.28&60.70&46.92\\
HYENA&NA&51.38&52.85&45.01\\
WSABIE&NA&71.28&72.08&66.22\\
\multirow{2}{*}{HLE}&Y&70.84&71.61&65.74\\
&N&68.86&70.00&63.32\\

\multirow{2}{*}{ProtoLE}&Y&72.67&73.54&67.58\\
&N&\textbf{75.78}&\textbf{76.50}&\textbf{70.43}\\
\multirow{2}{*}{Proto-HLE}&Y&71.97&72.89&67.05\\
&N&74.54&74.38&69.46\\
\hline
& &\multicolumn{3}{c|}{OntoNotes}\\
\hline
FIGER&NA&58.77&52.37&38.01\\
HYENA&NA&47.65&43.97&26.56\\
WSABIE&NA&62.03&55.83&43.61\\
\multirow{2}{*}{HLE}&Y&61.54&49.16&43.25\\
&N&59.52&54.01&41.60\\

\multirow{2}{*}{ProtoLE}&Y&60.90&54.68&42.82\\
&N&65.91&59.08&46.94\\
\multirow{2}{*}{Proto-HLE}&Y&62.71&56.64&44.81\\
&N&\textbf{68.23}&\textbf{61.27}&\textbf{49.30}\\
\hline
 & &\multicolumn{3}{c|}{Wiki}\\
\hline
FIGER&NA&\textbf{68.28}&64.71&47.37\\
HYENA&NA&45.51&43.80&30.67\\
WSABIE&NA&67.97&64.49&48.28\\
\multirow{2}{*}{HLE}&Y&67.09&65.65&47.01\\
&N&65.29&62.53&45.19\\

\multirow{2}{*}{ProtoLE}&Y&66.96&65.78&49.18\\
&N&68.06&\textbf{66.53}&\textbf{53.54}\\
\multirow{2}{*}{Proto-HLE}&Y&67.85&65.74&50.27\\
&N&66.61&65.29&50.45\\
\hline
 \end{tabular}
 \caption{Performance of FNET in a few-shots learning on 3 benchmark datasets }
\label{tab:few-shots}
\end{table}

\subsection{Zero-shot Fine-grained Entity Typing}

In this section, we evaluate our method's capability recognizing mentions of unseen fine-grained types. We assume that the training set contains only coarse-grained types (i.e., Level-1), and Level-2 types are unseen types to be removed from the training set. Table \ref{tab:zero-shots} shows the Micro-Precision for Level-1 and Level-2 types using top $k$ type candidates for type inference. NPMI is computed for Level-1 types. We manually build prototype lists for unseen types by choosing from a randomly sampled list of entity mentions. Level-3 types are ignored for OntoNotes as Level-3 types never show in top-10 list produced by all methods. As the prediction for coarse-grained types are the same with regard to $k$, we only list the results using $k=3$. 

One interesting finding on all three datasets is that combining hierarchical and prototypical information results in better classification of coarse-grained types. It suggests that embeddings of unseen fine-grained types contains information complementary to the embeddings of coarse-grained types. Since HLE actually produces random prediction on Level-2 types due to its sparse representation, HLE perform poorly on Level-2 types.\\ 

\begin{table}[!htb]
\centering
\begin{tabular}{|c|c|c|c|c|c|} 
\hline
 \multirow{3 }{*}{ Dataset}& \multirow{3 }{*}{ Method}& Micro-Precision @k 	&\multicolumn{3}{c|}{Micro-Precision @k }\\
 &&Level 1&\multicolumn{3}{c|}{Level 2} \\\cline{3-6}
 &&3&3&5&10\\
 \hline
 \multirow{3}{*}{BBN}&ProtoLE &76.71&\textbf{42.95}&\textbf{36.61}&\textbf{42.34}\\
 &HLE &70.44& 13.08&13.16&12.82\\
 &Proto-HLE &\textbf{76.89}&42.35&35.18&30.16\\\cline{2-6}

 \hline
 \multirow{3}{*}{OntoNotes} &ProtoLE &73.26&\textbf{21.01}&\textbf{13.72}&\textbf{12.22}\\
 &HLE &66.96&7.13&6.14&6.23\\
 &Proto-HLE &\textbf{76.33}&7.09&11.43&9.91\\\cline{2-6}

\hline
 \multirow{3}{*}{Wiki} &ProtoLE &65.52 &12.50&21.28&17.91\\
 &HLE &65.13& 0.00&8.82&8.99\\
 &Proto-HLE &\textbf{67.41}&\textbf{20.01}&\textbf{31.25}&\textbf{24.24}\\\cline{2-6}

 \hline
\end{tabular}
\caption{Performance of zero-shot entity typing}
\label{tab:zero-shots}
\end{table}

ProtoLE outperforms HLE by 100\%-300\% in terms of Micro-Precision. However, again the combination of prototypes and hierarchy achieves similar or better results than ProtoLE on BBN and Wikipedia dataset. The drop of precision of Proto-HLE on OntoNotes is likely due to a different nature of annotation. It is more prevalent in the test set of OntoNotes that one entity mention is annotated with multiple Level-1 types, and the presence of fine-grained types are less constrained by the label hierarchy. In such case, hierarchical constrains enforced by Proto-HLE might have negative impacts on type inference.

\section{Summary}
\label{conclusion}
In this paper, we present a prototype-driven label embedding method for fine-grained named entity typing (FNET). It shows that our method outperforms state-of-the-art embedding-based FNET methods in both few-shots and zero-shots settings. It also shows that combining prototype-driven label embeddings and type hierarchy can improve the prediction of coarse-grained types. 

\chapter{Binary Entity Embedding for Reranking ASR Hypotheses}
\label{tag:chap5}
In this chapter, we address the problem of reranking ASR hypotheses via binary embeddings of entities and words.
\section{Introduction}


Reranking models have been shown effective for reducing errors in a variety of Natural Language Processing tasks such as Named Entity Recognition~\cite{collins:2002,maalab}, Syntactic Parsing~\cite{collins:2005:drp,koo:2005} and Statistical Machine Translation~\cite{dtrans:2009}. A reranking model typically treats the baseline system as a black box and is trained to rank the competing hypotheses based on more complex or global information.

In an ASR system, discriminative language model (DLM) is first introduced by Roark et al.~\cite{roark:2004} for reranking ASR hypotheses. They adopt a single perceptron to modify the confidence scores of hypotheses generated by a baseline ASR system. By using only n-gram features, their reranking model is shown capable of reducing the Word Error Rate (WER) of an ASR system. His work is followed by several variants with a variety of feature choices such as syntactic features~\cite{Collins:2005dsl,Lambert:2013dp}, which try to capture correlation between simple features on the feature level. However, existing DLMs still suffer from poor generalization power and are vulnerable to a shortage of training data, because most of them rely on linear or log-linear models that fail to take into consideration the correlation of input features on the model level. 
Apart from feature engineering, using hidden variables encoding semantic information helps to improve the generalization power. 

Koo et al.~\cite{koo:2005} proposes a hidden-variable model to rerank syntactic parsing trees. By linking input features to hidden states corresponding to word senses or classes, they achieve improved accuracy over a linear baseline.
Inspired by the success of Koo et al., we propose to use the computational structure of Restricted Boltzmann Machine (RBM)~\cite{rbm:1986} for the task of ASR hypotheses reranking. RBM is a neural network composed of one hidden layer and one input layer. 
The hidden layer of RBM can be viewed as embeddings of the input and has been shown capable of capturing high-order correlation and semantic information in the context of language modeling~\cite{rbmlm,rbmlm2}. These approaches model the probability of a fixed length of word sequences, i.e., $N$-grams, using only local information, and are trained with a generative objective function. However, RBM cannot be directly used for ASR reranking due to its generative training manner. 

In this chapter, we propose two modifications to train RBM in a discriminative manner. We modify the energy function of RBM to incorporate the ASR confidence score, which has been proved critical for reranking by previous DLMs~\cite{roark:2004,Collins:2005dsl,Lambert:2013dp}. We then propose a novel discriminative objective function for training RBM with $N$-best lists of ASR hypotheses. Our method differs from existing RBM-based language models~\cite{rbmlm,rbmlm2} in two major aspects. Firstly, the proposed RBM reranker is trained discriminatively. Secondly, RBM in our method represents sentences of variable length as global feature vectors. The most attractive property of RBM is that the computational structure is flexible enough to incorporate prior knowledge~\cite{aspect:2015}. As function words have little meanings and are less important for language understanding~\cite{Fries:1952}, we decide to focus more on content words. 
In order to allow the reranking model to have a better knowledge of content words, we propose to fuse the hidden layer with information about a particular class of natural language concepts, i.e., named entities. For technical simplicity, we constrain the hidden layer of RBM to take only binary values. Therefore, the hidden layer can be seen as a binary embedding of the input. In this regard, the resulting binary embedding encodes the fusion of both word-level and concept-level information.

To our knowledge, this work is the first to consider using hidden layer and fusion of knowledge from both concept and word levels in the context of ASR hypotheses reranking. The remainder of this chapter is structured as follows: Section 2 describes in detail the proposed work; Section 3 shows the empirical results as well as analyses; finally, Section 4 concludes this work and provide more discussion on future work.

\section{Training RBM for ASR Reranking}
\subsection{Restricted Boltzmann Machine}
\begin{figure}[!htb]
\centering
\includegraphics[width=1.0\linewidth]{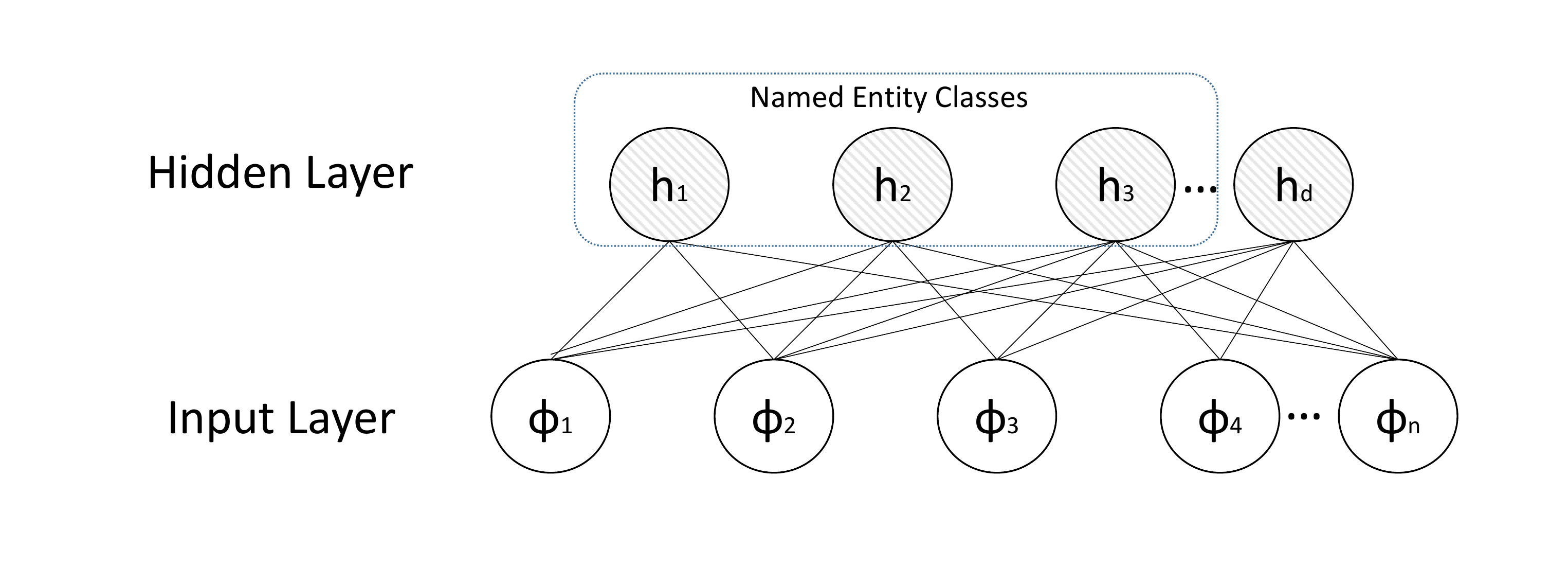}
\caption{Structure of RBM with Entity-related Prior~\cite{maasr}}
\label{fig:rbm}
\end{figure}

A Restricted Boltzmann Machine~\cite{rbm:1986} (see Figure~\ref{fig:rbm}) is a neural network composed of : one $n$-dimension input feature layer $\mathbf{\phi(t)} = [\phi_1(t),\phi_2(t),\cdots,\phi_n(t)]$, which is a global feature vector extracted for a raw input $t$, and one $d$-dimension binary hidden layer $\mathbf{h}=[h_ 1,h_2,\cdots,h_d]$. The joint probability $P_{\text{RBM}}(t,h)$ of hidden variables and raw input is defined as
\begin{align*}
P_{\text{RBM}}(t,h)& = \frac{e^{-E_{\text{RBM}}(t,h)}}{\sum_{t,h}e^{-E_{\text{RBM}}(t,h)}}\\
E_{\text{RBM}}(t,h)&=-\phi(t)^{T}Wh -b^T\phi(t)-c^Th,
\end{align*}
where $W\in R^{n\times d}$ is the matrix specifying the weights of connections between hidden and input layer, and $b \in R^{n}$ and $c \in R^{d}$ are the bias vectors of the two layers. $E_{\text{RBM}}(t,h)$ is called the energy function of RBM. The probability of a raw input $t$ is then defined as the marginal probability of $t$
$$P_{\text{RBM}}(t) = \sum_{h}P_{\text{RBM}}(t,h)$$
, and the training objective is to maximize the log likelihood of training data $D$
$$\sum_{t\in D}\ln P_{\text{RBM}}(t)$$



\subsection{Maximum Margin Training for RBM-based Reranker}
The goal of generative training of RBM is to learn a probability distribution, which is not necessary for choosing correct ASR hypotheses. Instead, the discriminative training allows the model to explicitly select ASR hypotheses containing fewer errors. In this section, we describe our discriminatively trained RBM-based reranking model, denoted as \textbf{dRBM}.

Before introducing the training objective function, we first introduce the energy function of RBM model. ASR posterior probabilities produced by the baseline ASR system have been shown useful for reranking in previous works on DLM~\cite{roark:2004,Collins:2005dsl,Lambert:2013dp}. We hence add ASR posterior to the energy function of RBM. The modified energy function is expressed as
\begin{align*}
E_{\text{dRBM}}(t,h) &= E_{\text{RBM}}(t,h) + E_{asr}(t)\\
E_{asr}(t) &= -w_0\ln(P(t|a)),
\end{align*}
where $P(t|a)$ is the posterior probability of a given ASR hypothesis $t$ given the acoustic input $a$, and $w_0$ is the weight of ASR confidence score fixed during training. We represent each hypothesis as a global feature vector $\phi(t)$ using a predefined set of feature functions. In this chapter, we mainly consider using unigram features, yet using more complicated features does not need to change the model.

Inspired by the maximum margin training for Bayesian Networks~\cite{Pernkopf:2012}, we adopt a discriminative objective function $L$ using likelihood ratio,
$$L=\frac{1}{|D|}\sum_{a \in D} \sum_{t'\in \text{GEN}(a)}\max(1-\ln \frac{P_{\text{dRBM}}(\hat{t})}{P_{\text{dRBM}}(t')},0),$$
where $D$ is the training set for the discriminative training of RBM and $|D|$ denotes the number of utterances in training set. GEN($a$) refers to the list of $N$-best hypotheses generated by the baseline ASR system for the acoustic input $a$, while $\hat{t}$ is the oracle-best in the $N$-best list of $t$. Intuitively, the learning process finds the parameter setting maximizing the margin between the oracle-best hypotheses and other hypotheses in the $N$-best list. 
The subgradient of the objective function is

\[\frac{\partial L}{\partial{\theta}}
= \sum_{a \in D}\sum_{t'\in \text{GEN}(a)}\mathbb{I}(\mathcal{F}(\hat{t})-\mathcal{F}(t')<1)( \frac{\partial{\mathcal{F}(t')}}{\partial{\theta}}-\frac{\partial\mathcal{F}(\hat{t})}{\partial{\theta}}),\]
where $\mathcal{F}(\cdot)$ is the free energy of RBM defined as
$$ \mathcal{F}(t) = -\ln\sum_he^{-E_{\text{dRBM}}(t,h)}$$ 
\begin{megaalgorithm}
 \DontPrintSemicolon
 \KwIn{\\
 	$D$: the training dataset \\
 	$\text{GEN}(a)$: $N$-best list for an utterance $t$ in the reference\\
 	$\lambda$: learning rate
}
 \For{k=1:K}{
 \For{$a \in D$}
 {
 Positive:\\ $\hat{t}= \mathtt{argmin}_{t' \in \text{GEN}(a)}\text{WER}(t')$ \\
 Negative:\\ $T^{-}=\{t'|1+Score(t')>Score(\hat{t})\},t' \in \text{GEN}(t)\}$\\
 \For{$t' \in T^-$}{
	$\theta \leftarrow \theta + \lambda\frac{\partial \mathcal{-F}(\hat{t})}{\partial \theta} $\\
	$\theta \leftarrow \theta - \lambda\frac{\partial \mathcal{-F}(t')}{\partial \theta}$
 }
 }
 }
 \caption{Discriminative training for RBM}
\end{megaalgorithm}

The training algorithm is described in Algorithm 1. For each acoustic input in training set, we select a set of hypotheses $T^{-}$, which are ranked higher than the oracle best hypothesis. Based on our analysis of the loss function, we boost the score of oracle best with its derivate of negative free energy and penalize hypothesis in $T^{-}$ with their derivates of negative free energy. Note that, as compared with standard RBM training~\cite{hinton:2002}, which iterates over input space or samples of input space, our discriminative training needs only to iterate over the $N$-best list which grows linearly with the size of training data and $N$-best list.

\subsection{Training with Prior Knowledge}
\label{prior}
The binary hidden layer of RBM allows for easily incorporating prior knowledge into the reranking model. We consider using two types of prior knowledge: named entity labels and pre-trained latent layer from texts. Firstly, to improve the capability of recognizing content words, we capture prior of a special class of content words -- named entities. As entity related prior also encodes information about word classes, it helps improving the generalization power of language models~\cite{WPE} as well.

Specifically, we extract pairs of named entity words and their classes from texts using a named entity tagger, which annotates the text with 3 widely-adopted named entity classes, i.e., LOCATION, ORGANIZATION and PERSON. As show in Fig.~\ref{fig:rbm}, 3 variables in the hidden layer of RBM are used to represent named entity classes. For purpose of reducing ambiguity, we remove words belonging to multiple entity classes. We denote the list of entity-class pairs as $G=\{w,e\}$, where $w$ is an index of the unigram feature in the input layer, and $e$ an index of entity-class variable in the hidden layer. The objective function is then augmented with an entity-related regularizer, 
\begin{align*}
L -\lambda\ln\prod_{w,e \in G}\prod (P(h_e=1|\phi_w)-1)^2,\\
P(h_e|\phi_w)=\sigma(c_e+W_{e,w}\phi_w).
\end{align*}
As introduced in Wang et al.~\cite{aspect:2015}, $P(h_e|\phi_w)$ denotes the probability of a hidden variable $h_e$ being activated by a given input feature $\phi_w$. 

To handle the data sparsity, we initiate connection matrix $W$ of RBM with values pre-trained using a large text corpus and the generative training. The pretraining captures the distributional semantics of input features~\cite{salakhutdinov:2009}.

\subsection{Scoring ASR Hypotheses}
To score a given hypothesis, we propose two scoring functions using our RBM-based reranker and its combination with SLP. First of all, the RBM-based reranking score $S_{\text{RBM}}$ is defined as the logarithm of the unnormalized probability $\tilde{P}_{\text{dRBM}}(t)$ assigned by the RBM-based reranker solely,\\

\begin{align*}
S_\text{RBM}(t) &=\ln\tilde{P}_{\text{dRBM}}(t) \\
&=\underbrace{w_0\ln(P(t|a))}_{\textbf{ASR posterior}}+（\underbrace{\sum_i^{n}{b_i\phi_i(t)}}_{\textbf{linear part}} + \underbrace{\sum_j^{d}\ln(1+e^{(c_i+W_i\phi(t))})}_{\textbf{hidden variable part}}.\\
\end{align*}
As shown above, the re-scoring function is composed of the original ASR posterior, a linear bias,  and the hidden variable component. In addition, SLP and RBM are likely to have encoded information complementary to each other due to their different structures and training methods. Therefore, we propose a late fusion of the two methods, which combines their confidence scores in the testing phase. The combined reranking score is
$$S(t) = S_{\text{RBM}}(t)+ \alpha S_{\text{SLP}}(t),$$
where $S_{\text{SLP}}(t)$ is the single perceptron based confidence score weighted by $\alpha$.

\section{Related Work}
DLM has been first introduced by Roark et al. in~\cite{roark:2004}, where simple features like $N$-gram is shown able to effectively reduce WER. 
This previous work is using a Single Layer Perceptron (SLP) to modify the original posterior probabilities of the outputs of a baseline ASR system using a linear function,
$$\log P(t|a) + \sum_{i}w_if_i(t),$$ 
where $\log P(t|a)$ is the log probability of a word sequence $t$ given the acoustic signal $a$, and $\{f_i(\cdot)\}$ are the set of feature functions of an utterance weighted by $\{w_i\}$. 
Different types of features extracted from syntactic trees~\cite{Collins:2005dsl} and dependency trees~\cite{Lambert:2013dp} have also been used to enrich the feature set.

Apart from feature engineering and using a linear combination of feature functions, inferring hidden variables from the observed input captures semantic information related to word classes and word senses. Our work is closely related to Koo et al.~\cite{koo:2005} who proposed a hidden-variable model to rerank syntactic parsing trees. For the tractability of their model, they put constrains on the connections between latent variables and visible variables (i.e., input layer) by splitting features into two sets. However, the way they divide features is specific to syntactic parsing and thus is not applicable to our task. Our model differs from Koo et al.~\cite{koo:2005} in the sense that the connection is not constrained by their feature type, but instead relying on the structure of RBM to build connections between input and hidden layer. 

RBM-based models~\cite{rbmlm,rbmlm2} have been explored for language modeling. Both approaches model the probability of a fixed length of word sequences, i.e., $N$-grams, and trained with a generative objective function. Our method differs from these methods in two major aspects: the training of proposed RBM reranker is discriminative, and it represents sentences of variable length as global feature vectors.


 \section{Experiment}
 \subsection{Dataset}
 We evaluate our work on the latest release of TedLium Corpus~\cite{tedlium:2014} which is a set of audio and manually transcribed texts of Ted talks. As shown in Table~\ref{tab:dataset5}. We split the training set of TedLium Version 2 into two parts: former Tedlium Training set Version 1 and the rest. The Version 1 part is a set of 774 Ted talks consisting of 56,800 utterances and more than 1.7 million words, while the remaining of the TedLium training set contains another 718 talks. The evaluation set of our experiment is the testing set of TedLium corpus, which is composed of 11 talks. Our text corpus is the ukWaC corpus~\cite{ukwac}, which is a collection of texts containing about 1.8 billion words.

 \begin{table}[!htb]
\centering
\begin{tabular}{|l | l | l |l|} 
\hline

 &Utterances & Talks& Words\\
 \hline
 ASR AM Train & 56.8K &774&1.7M \\
 Reranking Train(speech)&36.2K&718&0.9M\\
 Reranking Train(text) & 24M & -& 1.8B\\
 Reranking Test &1.15K &11 & 29K\\
 

\hline
\end{tabular}
\caption{Characteristics of the datasets used in experiments}
\label{tab:dataset5}
\end{table}

 \subsection{Baseline}
 
The baseline ASR system is based on KALDI\footnote{http://kaldi.sourceforge.net/} toolkit~\cite{kaldi} including a DNN-based acoustic model. It uses a pre-trained language model\footnote{http://cmusphinx.sourceforge.net/2013/01/a-new-english-language-model-release/}, which is released as part of Sphinx project~\cite{sphinx:2004} and has achieved a perplexity of 158.3 on a corpus of Ted Talks. The acoustic model is trained using the training set of TedLium Version 1. The rest of training set of TedLium Version 2 is used for training reranking models.

 The baseline SLP reranker is trained by following the work of Lambert et al.~\cite{Lambert:2013dp} that randomly selects $K$ pairs of hypotheses from the $N$-best lists. Specifically, we randomly select 100 pairs from the 100-best list. We ran the training process is for 10 iterations, as we cannot observe further WER reduction with more iterations. 
\subsection{RBM Setup}
We refer to the system integrating prior knowledge with dRBM as p-dRBM. Both dRBM and p-dRBM use 200 hidden units and are trained using the same dataset and 100-best hypotheses as SLP reranker. Since our focus is not on feature engineering, and for simplicity of interpreting our experiment results, we use only unigram features (i.e., single words) in our experiments, which have also been shown as the most effective features by previous works~\cite{Lambert:2013dp}. For training p-dRBM, we first crawled down a set of text summaries of ted talks from Ted website. We then create a list of 20 words for each entity category by tagging the collected text summaries with Stanford named entity recognition tool~\cite{stanfordner} following description in section~\ref{prior}. A basic RBM is trained using ukWaC corpus and used for initiating the connection matrix $W$ of p-dRBM. The late fusion of SLP and our proposed methods are denoted as SLP + dRBM and SLP + p-dRBM. We use $\lambda=0.01$ and $\alpha=1.0$ as weights of entity-related regularizer and SLP scores in late fusion.

\subsection{Evaluation}

First of all, we analyze the behavior of p-dRBM by computing the most-activated words by p-dRBM as shown in Table~\ref{tab:modelwords}. 
As shown in the previous section, the scoring function used by our method is a combination of ASR confidence score, a linear component (denoted as p-dRBM-L) and a hidden-variable component (denoted as p-dRBM-H). p-dRBM then takes as input one-hot vectors of words to compute their reranking scores. It shows that the linear component is mainly accounting for the function words, while the hidden component favors content words that are mostly nouns and adjectives. The final scoring function is a trade-off between function words and content words through a combination of the two components.

 \begin{table}[!htb]
\centering

\begin{tabular}{|l|l|l|l|l|l|} 
\hline
\textbf{p-dRBM-H}& \textbf{p-dRBM-L} & \textbf{p-dRBM}\\
\hline
integrated&	of&	integrated\\
demeanor&	and	&demeanor\\
	disgust	&the&	disgust\\
tattoo	&to	&tattoo\\
formula&	a	&formula\\
\hline
\end{tabular}
\caption{Most activated word by p-dRBM}
\label{tab:modelwords}
\end{table}

To investigate what is captured by RBM and potentially effective for improving the Word Error Rate (WER), we represent each hidden variable as a vector of words. These vectors represent how much a word is activated by a given hidden variable. Table~\ref{tab:topics} shows a selected set of hidden variables that can be seen as a set of topics. We found that RBM can capture meaningful topics by using only sentence-level co-occurrence. 
We then represent each word as a vector of hidden variables by taking the rows of the matrix $W \in R^{|V|\times d}$ of p-dRBM. 
 \begin{table}[!htb]
\centering
\begin{tabular}{|l|l|l|l|l|} 
\hline
 working&media & higher education& entertainment\\
 \hline
security &news &cambridge&scene \\
 services & forum&mary &story\\
 office&business&professor&tv\\
 home &press & william&songs\\
 for & new & royal & moving\\
\hline
\end{tabular}
\caption{Example topics learned by p-dRBM}
\label{tab:topics}
\end{table}

We rank words based on their cosine similarity with the queries and select the top 5 words for four query words. As shown in Table~\ref{tab:embd}, the top-ranked words all seem very relevant to the query words. Since our RBM is trained with sentence-level concurrence, which is different from the window-based methods, the word `similarity' seems more like a topical relatedness rather than syntactical similarity. In general, we can conclude that the resultant RBM-based reranking model to some extent captures the distributional semantics related to the topics of words.

\begin{table}[!htb]
\centering
\begin{tabular}{|l|l|l|l|l|} 
\hline

 japan&film & bible& computer\\
 \hline
india &story &greatest&software \\
 italy & music&holy &database\\
 asia&beautiful&truth&digital\\
 germany &famous & gospel&user\\
 china & classic & spirit & server\\

\hline
\end{tabular}
\caption{Most-similar words for queries using p-dRBM word embeddings}
\label{tab:embd}
\end{table}

As shown in Table~\ref{tab:wer}, we evaluate WER of reranking systems. It shows that the proposed discriminatively trained RBM produces greater WER reduction than baseline SLP rerankers. Effectiveness of using prior knowledge is validated by further improving WER over dRBM. The greatest absolute WER reduction (1.3\%) is achieved by the late fusion of SLP and p-dRBM, which confirms that our reranker captures information complementary to SLP.

\begin{table}[!htb]
\centering
\begin{tabular}{|l|c|c|} 
\hline
&\textbf{WER} & \textbf{WER (TF-IDF$\geq 3$)}\\
 \hline
ASR 1-best& 18.23 & 46.9\\
Oracle 1-best&11.42 &36.1\\
\hline
SLP & 17.76& 46.3\\
dRBM & 17.51& 44.6\\
p-dRBM & 17.36 & \textbf{43.8}\\
SLP + dRBM & 17.11 & 45.2\\
SLP + p-dRBM & \textbf{16.91} & 44.2\\
\hline
\end{tabular}
\caption{Performance of reranking model on TedLium corpus}
\label{tab:wer}
\end{table}
Since the latent layer of p-dRBM incorporates prior knowledge related to content words (e.g., named entities), it is desirable that the proposed method can better recognize content words, which are more critical for downstream applications such as spoken language understanding. To evaluate the performance of our proposed methods on recognizing content words in a more general way, we words that have higher TF-IDF scores are more likely to be content words. We hence assign more weight to errors involving a set of keywords with high TF-IDF instead of treating all words equally. Specifically, the list of keywords are chosen based on TF-IDF scores ($\geq 3.0$) computed from whole TedLium corpus. We use the weighted-word-scoring implementation in NIST SCLITE tool~\footnote{http://www1.icsi.berkeley.edu/Speech/docs/sctk-1.2/sclite.htm} by aggressively assigning weight $1.0$ to words on the list and $0.0$ to the rest.

\begin{figure}[!htb]
\centering
\includegraphics[width=10cm]{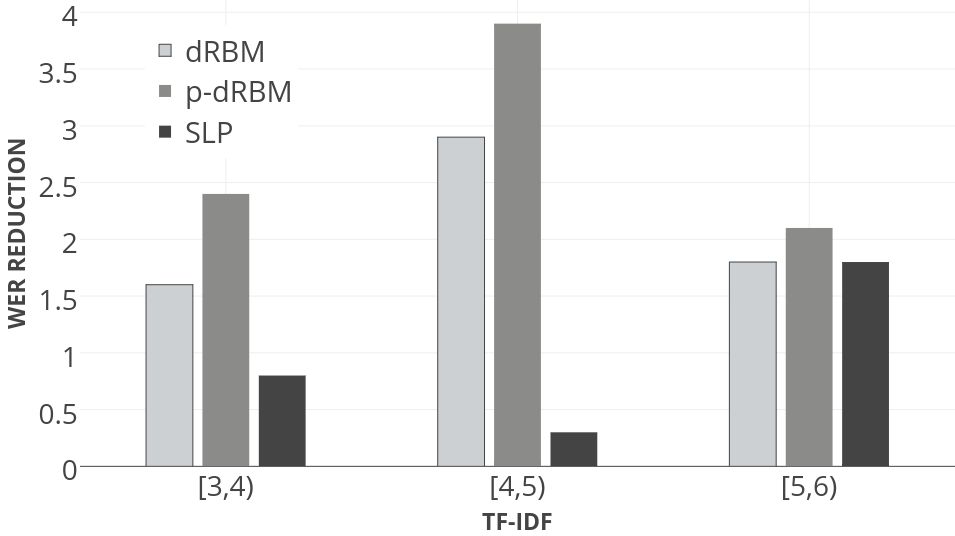}
\caption{WER reduction for words versus TF-IDF scores~\cite{maasr}}
\label{fig:wer}
\end{figure}

Table~\ref{tab:wer} clearly shows that baseline reranking systems (SLP) fail to reduce much WER for selected keywords. In comparison, proposed RBM rerankers, especially p-dRBM, have reduced more errors on chosen keywords without sacrificing overall performance. We further break down the TF-IDF scores into 3 bins. Figure~\ref{fig:wer} shows the WER reduction by all three approaches. Thanks to hidden variables, our methods are capable of better capturing the discriminative information for most content words. In particular, p-dRBM is shown working significantly better than other methods on words with medium TF-IDF scores ($<$5), which is a result of injecting named entity words, e.g., washington (TF-IDF$=3.87$), that mostly have TF-IDF between 3.0 and 5.0.

 \section{Summary}

In this chapter, we proposed an RBM-based reranking model that is discriminatively trained for reranking ASR hypotheses. The proposed reranking learns a binary embedding of words by fusing information of concept-level and word-level in its hidden layer. In comparison with single perceptron based reranker, our proposed approach achieves higher WER reduction. The success of combining single perceptron and RBM-based reranker suggests that two models actually capture complementary information useful for selecting less erroneous ASR hypotheses. In addition, we found that introducing concept-level knowledge to RBM-based reranker results in a better recognition of content words. In the future, we would like to explore the use of lexical knowledge obtained from different resources, e.g., WordNet~\cite{felwor} or SenticNet~\cite{camnt5}, as additional prior knowledge for the proposed model.


\chapter{Embedding Concepts for Target-based Sentiment Analysis}
\graphicspath{{Chapter6/fig/EPS/}{Chapter6/fig/}}
\label{tag:chap6}
In this chapter, we focus on leveraging commonsense concepts for target-based sentiment analysis. 

\section{Introduction}
In the past decade, sentiment analysis~\cite{campra} has become increasingly popular for processing social media data on online communities, blogs, wikis, microblogging platforms, and other online collaborative media. Sentiment analysis is a branch of affective computing research~\cite{porrev} that aims to classify text into either positive or negative, but sometimes also neutral~\cite{chasub}. 
Most of the literature is on English language but recently an increasing number of publications is tackling the multilinguality issue~\cite{loomul}.

While most works approach it as a simple categorization problem, sentiment analysis is actually a suitcase research problem~\cite{camsui} that requires tackling many NLP tasks, including named entity recognition~\cite{maalab}, word polarity disambiguation~\cite{xiawor}, personality recognition~\cite{majdee}, sarcasm detection~\cite{porloo}, and aspect extraction. The last one, in particular, is an extremely important subtask that, if ignored, can consistently reduce the accuracy of sentiment classification in the presence of multiple opinion targets.

Hence, aspect-based sentiment analysis (ABSA)~\cite{pontiki2014semeval,pontiki2016semeval,porasp} extends the typical setting of sentiment analysis with a more realistic assumption that polarity is associated with specific aspects (or product features) rather than the whole text unit. For example, in the sentence ``The design of the space is good but the service is horrible", the sentiment expressed towards the two aspects (``space" and ``service") is completely opposite. Through aggregating sentiment analysis with aspects, ABSA allows the model to produce a fine-grained understanding of people's opinion towards a particular product~\cite{camacsa}. 

Targeted (or target-dependent) sentiment classification~\cite{tang2015effective,dong2014adaptive,wang2017td}, instead, resolves the sentiment polarity of a given target in its context, assuming that a sentence might express different opinions towards different targeted entities. 
For instance, in the sentence ``I just log on my [facebook]. [Transformers] is boring", the sentiment expressed towards [Transformers] is negative, while there is no clear sentiment for [facebook]. Recently, targeted ABSA~\cite{saeidi:2016} has attempted to tackle the challenges of both ABSA and targeted sentiment analysis. The task is to jointly detect the aspect category and resolve the polarity of aspects with respect to a given target. 

Deep learning methods~\cite{nguyen2015phrasernn,wangattention,tang2015effective,tang2016aspect,wang2017td} have achieved great accuracy when applied to ABSA and targeted sentiment analysis. Especially, neural sequential models, such as LSTM networks~\cite{hochreiter1997long}, are of growing interest for their capacity of representing sequential information. Moreover, most of these sequence-based methods incorporate the attention mechanism, which has its root in the alignment model of machine translation~\cite{bahdanau2014neural}. Such mechanism takes an external memory and representations of a sequence as input and produces a probability distribution quantifying the concerns in each position of the sequence.
\begin{figure}[b]
\centering
 \includegraphics[width=\linewidth]{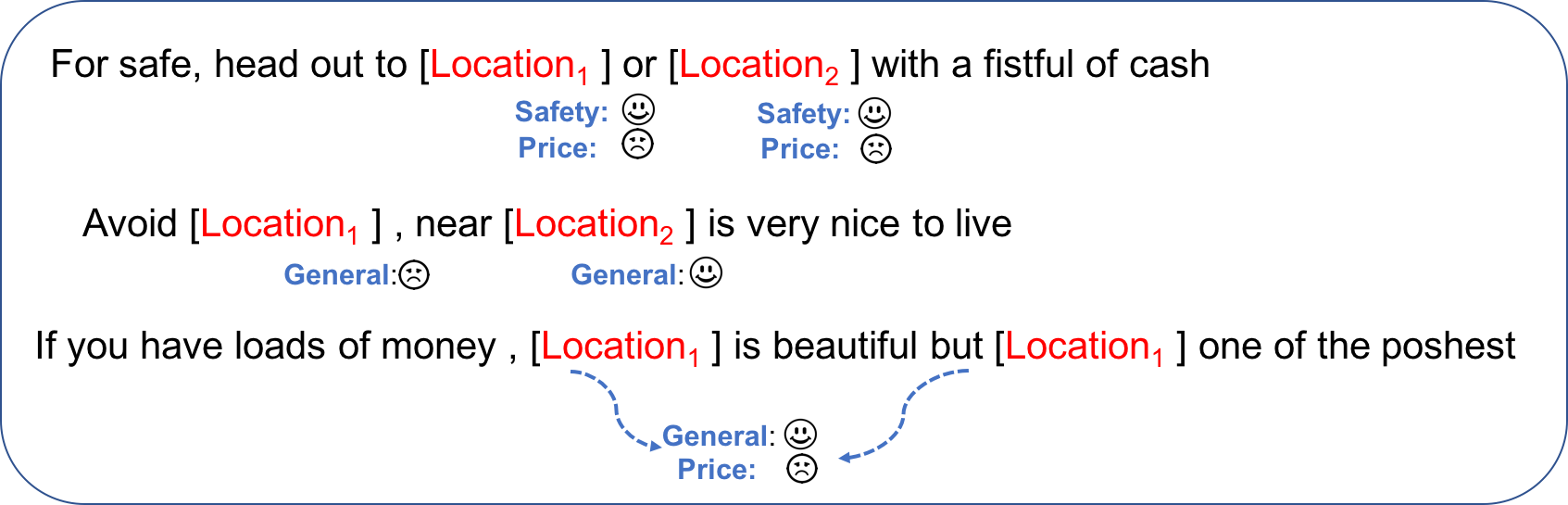}
\caption{Example of Sentihood data set}
\label{fig:chap6-example}
\end{figure}

Despite these advances in sentiment analysis, we identify three problems remaining unsolved in current state-of-the-art methods. Firstly, a given target might consist of multiple instances (mentions of the same target) or multiple words in a sentence, existing research assumes all instances are of equal importance and simply computes an average vector over such instances. 
This oversimplification conflicts with the fact that one or more instances of the target are often more tightly tied to sentiment than others. Secondly, hierarchical attention exploited by existing methods only implicitly models the process of inferring the sentiment-bearing words related to the given target and aspect as black-box.
Last but not least, existing research falls short in effectively incorporating into the deep neural network external knowledge, e.g., affective or commonsense knowledge, that could directly contribute to the identification of aspects and sentiment polarity. Without any constraints, moreover, the global attention model might tend to encode task-irrelevant information. 
To address these problems, our method simultaneously learns a target-specific instance attention as well as a global attention. In particular, our contribution is three-fold: 
\begin{enumerate}
\item We propose a hierarchical attention model that explicitly attends to first the targets and then the whole sentence;
\item We extend the classic LSTM cell with components accounting for integration with external knowledge;
\item We fuse affective commonsense concepts with word-level information in a deep neural network.
\end{enumerate}

\section{Methodology}
In this section, we describe the proposed attention-based neural architecture in detail: we first proposed the task definition of targeted ABSA, followed by an overview of the whole neural architecture; afterwards, we describe instance attention and global attention model; lastly, we describe the proposed knowledge-embedded extension of LSTM cell.
\begin{figure}[b]
\centering
 \includegraphics[width=\linewidth]{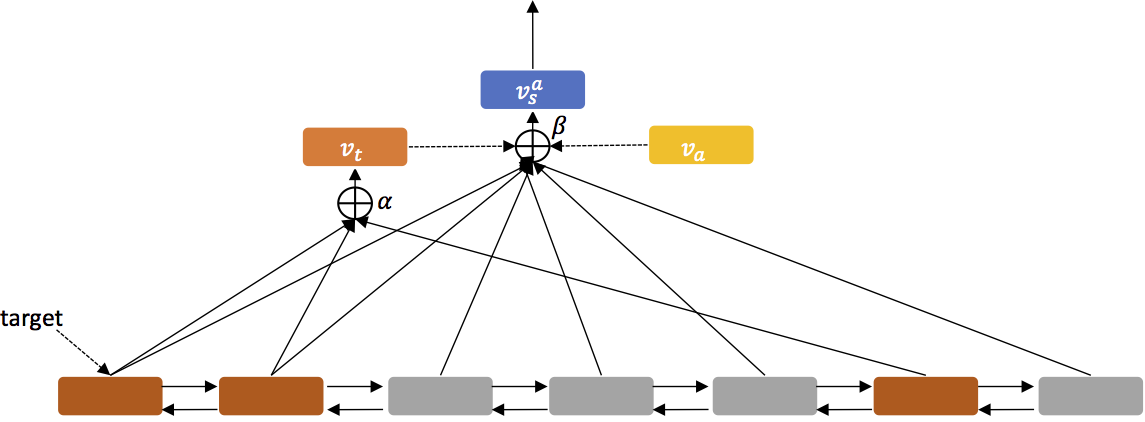}
\caption{Overview of the attentive neural architecture}
\label{fig:chap6-overview}
\end{figure}

\subsection{Task Definition}
A sentence $s$ consists of a sequence of words. Similar to~\cite{wang2017td}, we consider all mentions of the same target as a single target. A target $t$ composed of $m$ words in sentence $s$, denoted as $T=\{t_1,t_2,\cdots,t_i, \cdots,t_m\}$ with $t_i$ referring to the position of $i$th word in the target expression, the task of targeted ABSA can be divided into two subtasks. Firstly, it resolves the aspect categories of $t$ belonging to a predefined set. Secondly, it classifies the sentiment polarity with respect to each aspect category associated with $t$. 

For example, the sentence ``I live in [West London] for years. I like it and it is safe to live in much of [west London]. Except [Brent] maybe. " contains two targets, [$West London$] and [$Brent$]. Our objective is to detect the aspects and classify the sentiment polarity. The desired output for [$West London$] is [\textit{general}:\textbf{positive}; \textit{safety}:\textbf{positive}], while output for [$Brent$] should be [\textit{general}:\textbf{negative}; \textit{safety}:\textbf{negative}].

\subsection{Overview}

In this section, we provide an overview of the proposed method.
Our neural architecture consists of two components: the sequence encoder and a hierarchical attention component.

Fig.~\ref{fig:chap6-overview} illustrates how the neural architecture works. Given a sentence $s= \{w_1,w_2,\cdots,w_L\}$,
a look-up operation is first performed to convert input words into word embeddings $\{v_{w_1},v_{w_2},\cdots,v_{w_L}\}$. The sequence encoder, which is based on a bidirectional LSTM, transforms the word embeddings into a sequence of hidden outputs. The attention component is built on top of the hidden outputs. The target-level attention takes as input the hidden outputs at the positions of target expression (highlighted in brown) and computes a self-attention vector over these words. 

The output of target-level attention component is a representation of the target. Afterwards, the target representation together with the aspect embeddings is used for computing a sentence-level attention transforming the whole sentence into a vector. The sentence-level attention component returns one sentence vector for each aspect and target pair. The aspect-based sentence vector is then fed into the corresponding multi-class (e.g., None, Neural, Negative, and Positive for a 4-class setting; or None, Negative, and Positive for a 3-class setting) classifier to resolve the sentiment polarity. 

\subsection{Long Short-Term Memory Network}
The sentence is encoded using an extension of RNN~\cite{schuster1997bidirectional}, termed LSTM~\cite{hochreiter1997long}, which is firstly introduced by~\cite{hochreiter1997long} to solve the vanishing and exploding gradient problem faced by the vanilla RNN. A typical LSTM cell contains three gates: forget gate, input gate and output gate. These gates determine the information to flow in and flow out at the current time step. The mathematical representations of the cell are as follows:
\begin{equation}
\label{eq:lstm}
\begin{aligned}
f_{i} &=\sigma (W_{f} [x_i,h_{i-1}] + b_{f}) \\
I_{i} &=\sigma (W_{I} [x_i, h_{i-1}] + b_{I})\\
\widetilde{C}_{i} &=tanh (W_{C} [x_i,h_{i-1}] + b_{C})\\
C_{i} &= f_{i}* C_{i-1} +I_{i}* \widetilde{C}_{i} \\
o_{i} &=\sigma (W_{o} [x_i,h_{i-1}] + b_{o}) \\
h_{i} &= o_{i}* tanh({C}_{i}) \\
\end{aligned}
\end{equation}

where $f_{i}$, $I_{i}$ and $o_{i}$ are the forget gate, input gate and output gate, respectively. $W_{f}$, $W_{I}$, $W_{o}$, $b_{f}$, $b_{I}$ and $b_{o}$ are the weight matrix and bias scalar for each gate. $C_{i}$ is the cell state and $h_{i}$ is the hidden output. 
A single LSTM typically encodes the sequence from only one direction. However, two LSTMs can also be stacked to be used as a bidirectional encoder, referred to as bidirectional LSTM. For a sentence $s= \{w_1,w_2,\cdots,w_L\}$, bidirectional LSTM produces a sequence of hidden outputs, 

$$H = [h_1,h_2...h_L]= \begin{bmatrix}
  \overrightarrow{h_1} & \overrightarrow{h_2} & \cdots&\overrightarrow{h}_{L}   \\
  \overleftarrow{h_1} & \overleftarrow{h_2} & \cdots&\overleftarrow{h}_{L}
  \end{bmatrix}$$
where each element of $H$ is a concatenation of the corresponding hidden outputs of both forward and backward LSTM cells.

\subsection{Target-Level Attention}  
Based on the attention mechanism, we calculate an attention vector for a target expression. A target might consist of a consecutive or non-consecutive sequence of words, denoted as $T=\{t_1,t_2,\cdots,t_m\}$, where $t_i$ is the location of an individual word in a target expression. The hidden outputs corresponding to $T$ is denoted as $H' = \{h_{t_1},h_{t_2},\cdots,h_{t_m}\}$. We compute the vector representation of a target $t$ as
\begin{equation}
v_t = H' \alpha= \sum_{j} \alpha_j h_{t_j} \end{equation}
where the target attention vector $\alpha = \{\alpha_1,\alpha_2,\cdots,\alpha_m\}$ is distributed over target word sequence $T$. The attention vector $\alpha$ is a self-attention vector that takes nothing but the hidden output itself as input. The attention vector $\alpha$ of target expression is computed by feeding the hidden output into a bi-layer perceptron, as shown in Equation~\ref{eq:self-att}.
\begin{equation}
\label{eq:self-att}
\alpha = softmax(W_a^{(2)} tanh(W_a^{(1)}H'))
\end{equation}
where $W_a^{(1)} \in R^{d_m \times d_h}$ and $W_a^{(2)} \in R^{1 \times d_m}$ are parameters of the attention component.

\subsection{Sentence-Level Attention Model}

Following the target-level attention, our model learns a target-and-aspect-specific sentence attention over all the words of a sentence.
Given a sentence $s$ of length $L$, the hidden outputs are denoted as $H=[h_1,h_2,\cdots,h_L]$. An attention model computes a linear combination of the hidden vectors into a single vector, i.e.,
\begin{equation}
\label{eq:se}
v^a_{s,t} = H\beta = \sum_i \beta_i h_i
\end{equation}
where the vector $\beta = [\beta_1,\beta_2,\cdots,\beta_L]$ is called the sentence-level attention vector. Each element $\beta_i$ encodes the salience of the word $w_i$ in the sentence $s$ with respect to the aspect $a$ and target $T$. Existing research on targeted sentiment analysis or ABSA mostly uses targets or aspect terms as queries.

At first, each $h_i$ is transformed to a $d_m$ dimensional vector by a multi-layer neural network with a $tanh$ activation function, followed by a dense softmax layer to generate a probability distribution over the words in sentence $s$, i.e.,
\begin{equation}
\label{eq:sent-att}
\beta_a = softmax(v_a^{T} tanh(W_m(H'\odot v_t) ))
\end{equation}
where $v_a$ is the aspect embedding of aspect $a$, $H\odot v_t$ is the operation concatenating $v_t$ to each $h_i$; $W^{(1)}_m \in R^{d_m \times d_h}$ is the matrix mapping row vectors of $H$ to a $d_m$ dimensional space, and $W^{(2)}_m\in R^{1\times d_m}$ maps each new row vector to a unnormalized attention weight.

\subsection{Commonsense Knowledge}
\begin{figure}[t]
 \includegraphics[width=\linewidth]{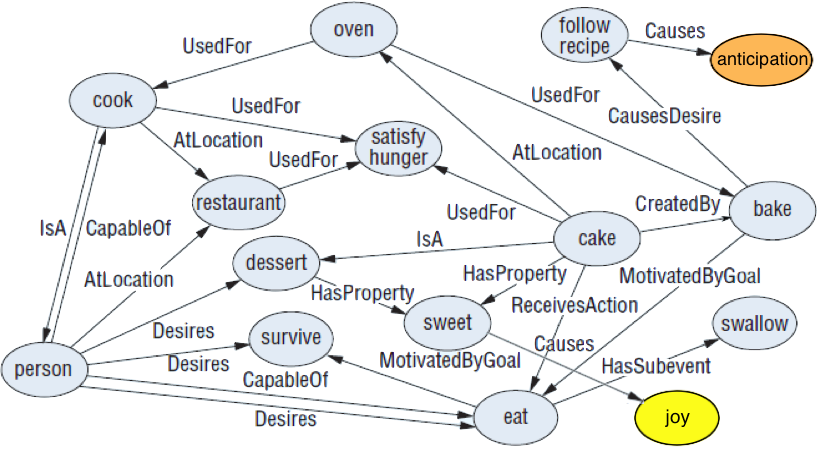}
 \centering
 \caption{A sketch of SenticNet semantic network~\cite{camaf2}}
 \label{fig:affnet}
\end{figure}

In order to improve the accuracy of sentiment classification, we use commonsense knowledge as our knowledge source to be embedded into the sequence encoder. In particular, we use SenticNet~\cite{camnt5}, a commonsense knowledge base that contains 100,000 concepts associated with a rich set of affective properties (Table~\ref{tab:affectnet}). These affective properties provide not only concept-level representation but also semantic links to the aspects and their sentiment. For example, the concept `rotten fish' has property ``KindOf-food'' that directly relates with aspects such as `restaurant' or `food quality', but also emotions, e.g., `joy', which can support polarity detection (Fig.~\ref{fig:affnet}).

\begin{table}[b]
\centering
 \caption{Example of SenticNet assertions}
 \label{tab:affectnet}
 \begin{tabular}{|l|c|c|c|c|}

 \toprule
   SenticNet&IsA-pet&KindOf-food&Arises-joy&...\\
 \midrule
  dog& 0.981& 0& 0.789& ...\\
cupcake& 0& 0.922& 0.910& ...\\
songbird &0.672& 0& 0.862 &...\\
gift& 0 &0 &0.899 &...\\
sandwich &0& 0.853& 0.768& ...\\
rotten fish& 0 &0.459 &0 &...\\
win lottery &0& 0 &0.991& ...\\
bunny &0.611& 0.892 &0.594 &...\\
police man &0& 0& 0 &...\\
cat &0.913 &0 &0.699 &...\\
rattlesnake &0.432 &0.235 &0 &...\\
 \bottomrule
 \end{tabular}
\end{table}

However, the high dimensionality of SenticNet hinders it from being used in deep neural models. AffectiveSpace~\cite{camaf2} has been built to map the concepts of SenticNet to continuous low-dimensional embeddings without losing the semantic and affective relatedness of the original space. We have included a brief introduction of efficiently generating low-dimensional concept embeddings using random projection in Section~\ref{sec:rp}. Based on this new space of concepts, we embed concept-level information into deep neural sequential models to better classify both aspects and sentiment in natural language text.

\subsection{Sentic LSTM}
In order to leverage SenticNet's affective commonsense knowledge efficiently, we propose an affective extension of LSTM, termed Sentic LSTM. It is reasonable to assume that SenticNet concepts contain information complementary to the textual word sequence as, by definition, commonsense knowledge is about concepts that are usually taken for granted and, hence, absent from the text. Sentic LSTM aims to entitle the concepts with two important roles: 1) assisting with the filtering of information flowing from one time step to the next and 2) providing complementary information to the memory cell. At each time step $i$, we assume that a set of knowledge concept candidates can be triggered and mapped to a $d_c$ dimensional space. We denote the set of $K$ concepts as $\{\mu_{i,1},\mu_{i,2},\cdots,\mu_{i,K}\}$. First, we combine the candidate embeddings into a single vector as follows:

\begin{equation}
\label{eq:average-mu}
\mu_i = \frac{1}{K}\sum_j \mu_{i,j} 
\end{equation}

As we realized that there are only up to 4 extracted concepts for each time step, we simply use the average vector (although a more sophisticated attention model can also be easily employed to replace the averaging function).

\begin{equation}
\label{eq:Sentic-LSTM}
\begin{aligned}
f_{i} &=\sigma (W_{f} [x_i,h_{i-1},\textcolor{red}{\mu_i}] + b_{f}) \\
I_{i} &=\sigma (W_{I} [x_i, h_{i-1},\textcolor{red}{\mu_i}] + b_{I})\\
\widetilde{C}_{i} &=tanh (W_{C} [x_i,h_{i-1}] + b_{C})\\
C_{i} &= f_{i}* C_{i-1} +I_{i}* \widetilde{C}_{i} \\
o_{i} &=\sigma (W_{o} [x_i,h_{i-1},\textcolor{red}{\mu_i}] + b_{o}) \\
\color{red}o^c_{i} &\color{red}=\sigma (W_{co} [x_i,h_{i-1},\mu_i] + b_{co}) \\
h_{i} &= o_{i}* tanh({C}_{i}) + \color{red}o^c_{i}*tanh(W_c \mu_i) \\
\end{aligned}
\end{equation}

Our affective extension of LSTM is illustrated in Equation~\ref{eq:Sentic-LSTM}. At first, we assume that affective concepts are meaningful cues to control the information of token-level information. For example, a multi-word concept `rotten fish' might indicate that the word `rotten' is a sentiment-related modifier of its next word `fish' and, hence, less information should be filtered out at next time step. We thus add knowledge concepts to the forget, input, and output gate of standard LSTM to help to filter the information. The presence of affective concepts in the input gate is expected to prevent the memory cell from being affected by input tokens conflicting with pre-existing knowledge. Similarly, the output gate uses such knowledge to filter out irrelevant information stored in the memory. 

Another important feature of Sentic LSTM is based on the assumption that the information from the concept-level output is complementary to the token level. Therefore, we extended the regular LSTM with an additional knowledge output gate $o^c_{i}$ to output concept-level knowledge complementary to the token-level memory. Since AffectiveSpace is learned independently, we leverage a transformation matrix $W_c \in R^{d_h \times d_\mu}$ to map it to the same space as the memory outputs. In other words, $o^c_{i}$ models the relative contributions of token level and concept level. 

Moreover, we notice that $o^c_{i}*tanh(W_c \mu_i)$ actually resembles the functionality of the sentinel vector used by~\cite{yang-mitchell:2017}, which allows the model to choose whether to use affective knowledge or not. 

\subsection{Prediction and Parameter Learning}
The objective to train our classier is defined as minimizing the sum of the cross-entropy losses of prediction on each target-aspect pair, i.e., 
$$\mathcal{L}_s = \frac{1}{|D|} \sum_{s \in D}\sum_{t \in s}\sum_{a\in A} \log{p^a_{c,t}} $$
where $A$ is the set of predefined aspects, and $p^a_{c,t}$ is the probability of the gold-standard polarity class $c$ given target $t$ with respect to a sentiment category $a$, which is defined by a softmax function,
$$ p^a_{c,t} = softmax(W^p v^a_{s,t}+ b^a_s)$$
where $W^p$ and $b^a_s$ are the parameters to map the vector representation of target $t$ to the polarity label of aspect $a$.
To avoid overfitting, we add a dropout layer with dropout probability of 0.5 after the embedding layer. We stop the training process of our model after 10 epochs and select the model that achieves the best performance on the development set.

\section{Experiments}
\subsection{Dataset and Resources}
We evaluate our method on two datasets: SentiHood~\cite{saeidi:2016} and a subset of Semeval 2015~\cite{pontiki-EtAl:2015:SemEval}. SentiHood is built by querying Yahoo! Answers with location names of London city. Table~\ref{tab:dataset6} shows statistics of SentiHood. The whole dataset is split into train, test, and development set by the authors. Overall, the entire dataset contains 5,215 sentences, with 3,862 sentences containing a single target and 1,353 sentences containing multiple targets. It also shows that there are approximately two third of targets annotated with aspect-based sentiment polarity (train set: 2476 out of 2977; test set:1241 out of 1898; development set: 619 out of 955). On average, each sentiment-bearing target has been annotated with 1.37 aspects. To show the generalizability of our methods, we build a subset of the dataset used by Semeval-2015. We remove sentences containing no targets as well as $NULL$ targets. To be comparable with SentiHood, we combine targets with the same surface form within the same sentence as mentions of the same target. In total, we have 1,197 targets left in the training set and 542 targets left in the testing set. On average, each target has 1.06 aspects.

\begin{table}[b]
\centering
 \caption{SentiHood dataset}
 \label{tab:dataset6}
 \begin{tabular}{lccc}
 \toprule
  &Train & Dev& Test\\
 \midrule
 Targets& 3,806&955&1,898\\
 Targets w/ Sentiment&2,476&619&1,241\\
 Aspects per Target(w/ Sentiment) &1.37&1.37& 1.35\\
 \bottomrule
 \end{tabular}
\end{table}

To inject the commonsense knowledge, we use a syntax-based concept parser\url{github.com/senticnet} to extract a set of concept candidates at each time step, and use AffectiveSpace\url{sentic.net/downloads} as the concept embeddings. In case no concepts are extracted, a zero vector is used as the concept input.

\subsection{Experiment Setting}
We evaluate our method on two sub-tasks of targeted ABSA: 1) aspect categorization and 2) aspect-based sentiment classification.
Following Saeidi et al.~\cite{saeidi:2016}, we treat the outputs of aspect-based classification as hierarchical classes. For aspect categorization, we output the label (e.g., in the 3-class setting, it outputs `Positive', `Negative', or `None') with the highest probability for each aspect. For aspect-based sentiment classification, we ignore the scores of `None'. For evaluating the aspect-based sentiment classification, we simply calculate the accuracy averaged over aspects. We evaluate aspect categorization as a multi-label classification problem so that results are averaged over targets instead of aspects. 

We evaluate our methods and baseline systems using both loose and strict metrics. We report scores of three widely used evaluation metrics of multi-label classifier: Macro-F1, Micro-F1, and strict Accuracy. 
Given the dataset $D$, the ground-truth aspect categories of the target $t\in D$ is denoted as $Y_t$, while the predicted aspect categories denoted as $\widehat{Y}_t$. The three metrics can be computed as
\begin{itemize}
\item Strict accuracy (Strict Acc.): $ \frac{1}{D}\sum_{t\in D} \sigma(Y_t=\widehat{Y}_t)$, where $\sigma(\cdot)$ is an indicator function.
\item Macro-F1 $= 2\frac{\text{Ma-P} \times \text{Ma-R}}{\text{Ma-P} + \text{Ma-R}}$, which is based on Macro-Precision (Ma-P) and Micro-Recall (Ma-R) with Ma-P $= \frac{1}{|D|}\sum_{t\in D}\frac{|Y_t\cap \widehat{Y}_t|}{\widehat{Y_t}}$, and Ma-R$ =\frac{1}{|D|} \sum_{t\in D}\frac{|Y_t\cap \widehat{Y}_t|}{Y_t}$.
\item Micro-F1 $= 2\frac{\text{Mi-P} \times \text{Mi-R}}{\text{Mi-P} + \text{Mi-R}}$, which is based on Micro-Precision (Mi-P) and Micro-Recall (Mi-R), where Mi-P$=\frac{\sum_{t\in D}|Y_t\cap \widehat{Y}_t|}{\sum_{t\in D}\widehat{Y}_t}$, and Mi-R$=\frac{\sum_{t\in D}|Y_t\cap \widehat{Y}_t|}{\sum_{t\in D}Y_t}$. 
\end{itemize}

\begin{table*}
\centering
\small
 \caption{System performance on SentiHood dataset}
 \label{tab:perf}
 \begin{tabular}{rcc}
 \toprule
 	 & \multicolumn{2}{c}{Sentiment} \\

  \multicolumn{2}{c}{Sentiment Acc. (\%)}\\
 dev&test\\
  \midrule
 TDLSTM&82.60&81.82\\

 LSTM + TA &83.80&84.29\\
 
 LSTM + TA + SA & 	86.00&86.75\\
 
LSTM + TA + DMN SA&84.80& 83.36\\
\midrule
 LSTM + TA + SA + KB Feat& 		87.00&	 88.70\\
 LSTM + TA + SA + KBA &	87.40	&		87.98\\
 Recall LSTM + TA + SA & 86.80&				86.85\\
Sentic LSTM + TA + SA&	\textbf{88.80}&				\textbf{89.32}\\

\bottomrule
\end{tabular}
\begin{tabular}{rcccccc}
 \toprule
 & \multicolumn{6}{c}{Aspect Categorization}	\\

 &\multicolumn{2}{c}{Strict Acc. (\%)} &\multicolumn{2}{c}{Macro F1 (\%)} & \multicolumn{2}{c}{Micro F1 (\%)} \\
 
 	&dev&test&dev&test&dev&test\\
  \midrule
 TDLSTM&50.27&50.83&	59.03&58.17&	55.72&55.78\\

 LSTM + TA & 54.17&52.02&	62.90&61.07&	60.56&59.02	\\
 
 LSTM + TA + SA & 68.83&66.42&	79.36&76.69&	79.14&76.64\\
 
LSTM + TA + DMN SA&60.66&60.14&68.89&70.19&67.28&	68.37\\
\midrule
 LSTM + TA + SA + KB Feat& \textbf{69.38}	&64.76&\textbf{80.00} & 76.33&	\textbf{79.79}&76.08\\
 LSTM + TA + SA + KBA & 68.08&65.12&	78.68&76.40&	78.73&	76.46\\
 Recall LSTM + TA + SA & 68.64&	64.66&78.44&75.61&	78.53&75.91\\
Sentic LSTM + TA + SA& 69.20&\textbf{67.43}&	78.84&\textbf{78.18}&	79.09&\textbf{77.66}\\

\bottomrule
\end{tabular}
\end{table*}

\subsection{Performance Comparison}

We compare our proposed method with the methods that have been proposed for targeted ABSA as well as methods proposed for ABSA or targeted sentiment analysis but applicable to targeted ABSA. 

Furthermore, we also compare the performances of several variants of our proposed method in order to highlight our technical contribution. We run the model for multiple times and report the results that perform best in the development set. For Semeval-2015 dataset, we report the results of the final epoch.\\

\begin{table*}
\centering
\small
 \caption{System performance on Semeval-2015 dataset}
 \label{tab:perf-sem}
 \begin{tabular}{rcccc}
 \toprule
 & \multicolumn{3}{c}{Aspect Categorization}	 & Sentiment \\

 &Strict Acc.&Macro F1 & Micro F1 & Sentiment Acc. \\
  \midrule

 TDLSTM&65.49&	70.56&	69.00	&68.57\\

 LSTM+TA &66.42&	71.71&	70.06&	69.24 \\
 
 LSTM+TA+SA &63.46&	70.73&	66.18&	74.28\\
 
LSTM+TA+DMN SA&48.33&	52.73&	51.39&	69.07\\
\midrule
 LSTM+TA+SA+KB Feat&65.68&	74.46&	70.71&	76.13 \\
 LSTM+TA+SA+KBA & \textbf{67.34}&	74.36&	71.78&	73.10\\
 Recall LSTM + TA + SA &66.05&72.90&	69.66&	74.11\\
 Sentic LSTM + TA + SA& 67.34&	\textbf{76.44}&	\textbf{73.82}&	\textbf{76.47}\\

\bottomrule
\end{tabular}
\end{table*}
\begin{itemize}

\item \textbf{TDLSTM}: TDLSTM~\cite{tang2015effective} adopts Bi-LSTM to encode the sequential structure of a sentence and represents a given target using a vector averaged on the hidden outputs of target instances.

\item \textbf{LSTM + TA}: Our method learns an instance attention on top of the outputs of LSTM to model the contribution of each instance. 

\item \textbf{LSTM + TA + SA}: In addition to target instance attention, we add a sentence-level attention to the model. 
\item \textbf{LSTM + TA + DMN SA}: The sentence-level attention is replaced by a dynamic memory network with multiple hops~\cite{tang2016aspect}. We run the memory network with different numbers of hops and report the result with 4 hops (best performance on development set of SentiHood). We exclude the case of zero hops as it corresponds to Bi-LSTM + TA + SA.
\item \textbf{LSTM + TA + SA + KB Feat}: Concepts are fed into the input layer as additional features.
\item \textbf{LSTM + TA + SA + KBA}: This is an integration of the method proposed by~\cite{yang-mitchell:2017}, which learns an attention over the concept embeddings (the embeddings are combined with the hidden output before being fed into the classifier). 

\item \textbf{Recall LSTM + TA + SA}: LSTM is extended with a recall knowledge gate as in~\cite{xu2016incorporating}.

\item \textbf{Sentic LSTM + TA + SA}: The encoder is replaced with the proposed knowledge-embedded LSTM. 

\end{itemize}

$ $\\

The word embedding of the input layer is initialized by a pre-trained skip-gram model~\cite{mikolov2013distributed} with 150 hidden units on a combination of Yelp\footnote{\url{yelp.com.sg/dataset/challenge}} and Amazon review dataset~\cite{he2016ups} and 50 hidden units for the bi-directional LSTM. 

\subsection{Results of Attention Model}
Table \ref{tab:perf} and Table~\ref{tab:perf-sem} show the performance on SentiHood and Semeval-15 dataset, respectively. In comparison with the non-attention baseline (Bi-LSTM+Avg.), we can find that our best attention-based model significantly improves aspect categorization (by more than 20\%) and sentiment classification (approximately 10\%) on SentiHood. However, it is notable that, on the Semeval-2015 dataset, the improvement is relatively smaller. We conjecture the reason is that SentiHood has masked the target as a special word ``LOCATION'', which resulted less informative than the full name of aspect targets that are used by Semeval-2015. 

Hence, using only the hidden outputs regarding the target does not suffice to represent the sentiment of the whole sentence in SentiHood. Compared with target averaging model, the target-level attention achieves some improvement (even though not significant), as the target attention is capable of identifying the part of target expressions with higher sentiment salience. On the other hand, it is notable that the two-step attention achieves significant improvement on both aspect categorization and sentiment classification, indicating that the target- and aspect-dependent sentence attention could retrieve information relevant to both tasks. 

To our surprise, using multiple hops in the sentence-level attention fails to bring in any improvement. The performance even falls down significantly on Semeval-2015 with a much smaller number of training instances but larger aspect set than SentiHood. We conjecture the reason is that using multi-hops increases the number of parameter to learn, which makes it less applicable to small and sparse datasets.

\subsection{Visualization of Attention}
\begin{figure}[b]
\centering
\includegraphics[width=\linewidth]{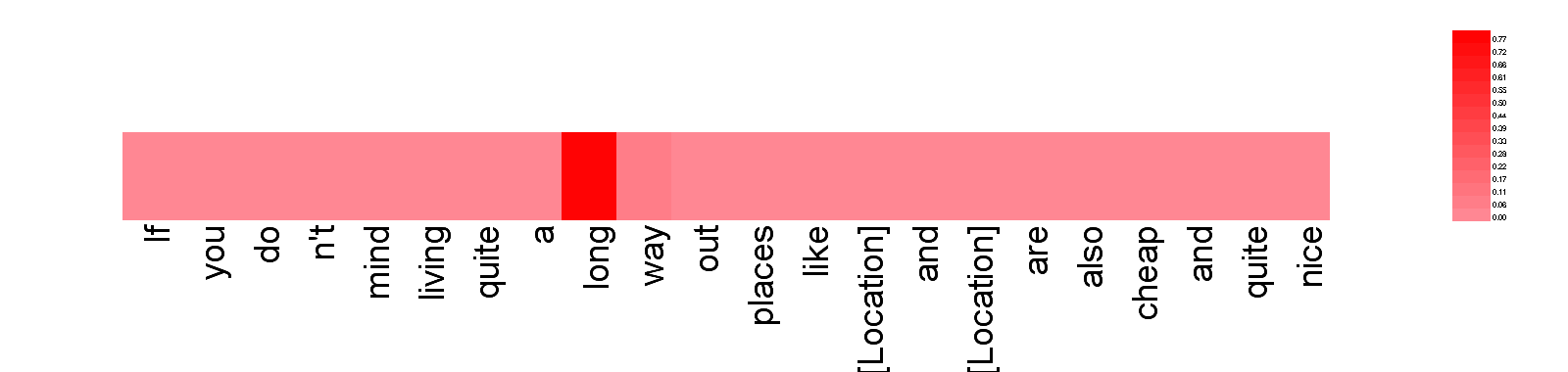}
\includegraphics[width=\linewidth]{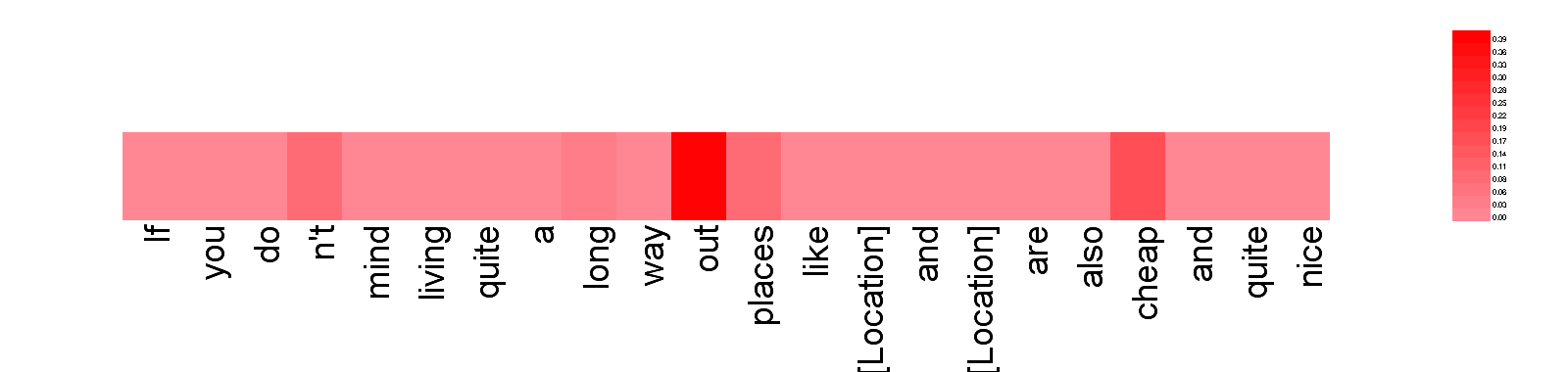}
\caption{Example of sentence-level attention}
\label{fig:ga_example}
\end{figure}

We visualize the attention vectors of sentence-level attention in Figure~\ref{fig:ga_example} with respect to ``Transition-location'' and ``Price'' aspects. The two attention vectors have encoded quite different concerns in the word sequence. 

\begin{figure}[t]
 \centering
 \includegraphics[width=\linewidth]{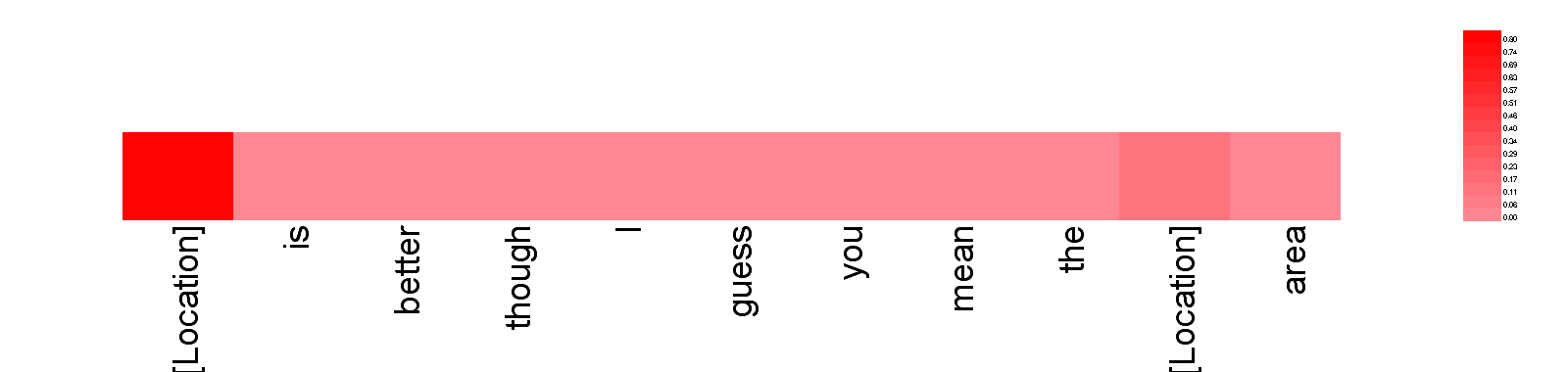}
\caption{Example of target-level attention}
 \label{fig:la_example}
\end{figure}

In the first example, the `Transition-location' attention attends to the word ``long", which is expressing a negative sentiment towards the target. In comparison, the `Price' attention attends more to the word `cheap', which is related to the aspect. That is to say, the two attention vectors are capable of distinguishing information related to different aspects. 
As visualized in Figure~\ref{fig:la_example}, the target-level attention is capable of selecting the part of target expression of which the aspect or sentiment is easier to be resolved. 


\subsection{Results of Knowledge-Embedded LSTM}
It can be seen from Table~\ref{tab:perf} and~\ref{tab:perf-sem} that injecting knowledge into the model improves the performance in general. Since AffectiveSpace encodes the information about affective properties that are semantically related to the aspects, it is reasonable to find out that it can improve performance on both tasks. The results show that our proposed Sentic LSTM outperforms baseline methods, even if not significantly. 

An important outcome of the experiments is that Sentic LSTM significantly outperforms a baseline (LSTM + TA + SA + KB feat) feeding the knowledge features to the input layer, which confirms the efficacy of using a knowledge output gate to control the flow of background knowledge. Furthermore, the superior performance of Sentic LSTM over Recall LSTM and KBA indicates that the activated knowledge concepts can also help filtering the information that conflicts with the background knowledge.

\subsection{AffectiveSpace v.s. SentiWordNet}
\begin{table}[!htb]
\centering
\small
 \caption{Comparison of systems using AffectiveSpace and SentiWordNet (SH stands for Sentihood, and SE stands for SemEval-15) }
 \label{tab:swn}
 \begin{tabular}{rcccccccc}
  \toprule
  & \multicolumn{6}{c}{Aspect Categorization}	 & \multicolumn{2}{c}{Sentiment} \\

  &\multicolumn{2}{c}{Strict Acc. (\%)} &\multicolumn{2}{c}{Macro F1 (\%)} & \multicolumn{2}{c}{Micro F1 (\%)} & \multicolumn{2}{c}{Sentiment Acc. (\%)}\\

 	&SH&SE&SH&SE&SH&SE&SH&SE\\
 AffectiveSpace &\textbf{67.71}&\textbf{70.47}&\textbf{78.05}&        	\textbf{77.90}&\textbf{78.19}&	\textbf{75.12}&\textbf{89.63}&	\textbf{78.31}\\
 SentiWordNet &60.23&66.42&70.03&73.60&69.48&71.18&85.01&75.79\\
\bottomrule
\end{tabular}
\end{table}
At last, we compare systems using different sentiment knowledge bases. SentiWordNet is a lexicon-based sentiment knowledge base consisting of word sense synsets annotated with sentiment polarity. However, it is notable that SentiWordNet contains neither commonsense concepts nor affective properties, which are the key features of AffectiveSpace. Consequently, we have to use randomly initialized embeddings for representing the SentiWordnet synsets. Word synsets are mapped to the same 100 dimension embeddings as AffectiveSpace does. Each word in the sentence is mapped to its word sense with the help of a Word Sense Disambiguation tool. We deliberately remove the neutral synsets (i.e., those having zero values for both Positive and Negative) to emphasize on the sentiment-bearing words. Table~\ref{tab:swn} shows the comparison of the two knowledge base. We report the results using our best-performed system (Sentic LSTM + TA + SA). It shows that using the AffectiveSpace achieves superior performance than using SentiWordNet. We conjecture the reason is that the word sense synsets is not as informative as affective properties. Moreover, probably because the link between word senses and aspects are not straightforward, we find the gap in sentiment classification is smaller than aspect categorization. 

\section{Summary}
In this chapter, we propose a neural architecture for the task of targeted ABSA. We explicitly modeled the attention as a two-step model which encodes targets and full sentence. The target-level attention learns to attend to the sentiment-salient part of a target expression and generates a more accurate representation of the target, while the sentence-level attention searches for the target- and aspect-dependent evidence over the full sentence. Moreover, we propose an extension of the LSTM cell so that it could more effectively incorporate affective commonsense knowledge when encoding the sequence into a vector. In the future, we would like to collectively analyze the sentiment of multiple targets co-occurring in the same sentence and investigate the role of commonsense knowledge in modeling the relation between targets.

\chapter{Summary and Future Work}
\label{tag:chap7}

In this final chapter, we summarize our contribution to the problem of leveraging concept-level information for natural language processing, followed by possible directions of research that we would like to explore in the future. 
\section{Summary of Proposed Methods}
Concepts are playing a critical role in natural languages, especially in the context of opinion understandings. In this thesis, we address the problem of leveraging concept-level information for concept-centered natural language processing tasks that are core modules in a pipeline opinion understanding system with a particular focus on fusing information from multiple levels. 
We identify our main contribution as follows:

\begin{itemize}
\item In Chapter~\ref{tag:chap3}, we demonstrate that incorporating entity-based features into the learning process of word embedding could better encode entity-related information in the word embedding space. We present a novel method to learn word embeddings in a multi-task setting, where entity-level information is fused with the word-level by setting the training objective to predicting the entity-related features. Each entity-related feature type is modeled as a separate subtask of prediction, while the word embedding layer is shared by all subtasks. We then demonstrate the effectiveness of the resulting word embedding by feeding word embeddings as features into a CRF-based NER tagger.

\item In Chapter~\ref{tag:chap4}, we capture the semantic correlation between fine-grained entity types. We propose a process to represent the entity labels with their prototpyes. The process can be implemented via either manual selection or an automated algorithm based on PMI. At the core of our proposed method are word embeddings learned from a large corpus. The embedding of each label is then defined as the average word embedding of its prototypes. In doing so, we able to project the entity labels into the word embedding space that has encoded distributional semantics. We validate the proposed label embedding with two different settings of fine-grained named entity typing: 1) few-shots learning, where all labels are assumed having presented in the training data; and 2) zero-shot learning, where the second level entity types are assumed absent from the training set. Our experiment shows that the resulting label embedding outperforms existing methods in both few-shots and zero-shots learning setting.

\item In Chapter~\ref{tag:chap5}, we study the problem of using entity-type information to help to rank ASR hypotheses. We propose an RBM based ASR reranking model that can be trained using discriminative objective function. The hidden layer of RBM are in fact binary embeddings of the input. We design a regularization term to softly constrain the binary embeddings such that it can encode a better knowledge of entity types. The resulting binary embedding could also be deemed as a fusion of entity-related prior information and word-level input. Our experiment shows that the proposed ASR reranking model can outperform the popularly used single-perceptron baseline thanks to a better knowledge of natural language concepts.

\item In Chapter~\ref{tag:chap6}, we address the problem of targeted sentiment analysis which is an important subtask of opinion understanding. Our solution leveraged commonsense concepts via an extended LSTM network. Our contribution also includes a two-layer attention neural architecture that jointly learns the representation of the targeted entity as well as a target-aware representation of the whole sentence. More specifically, our framework explicitly fuses the concept-level and word-level information via a separate output gate of LSTM cell. The experiment shows that the knowledge of AffectiveSpace provides complementary semantics to assist resolving both aspect categories as well as the corresponding sentiment polarity.
\end{itemize}
We further summarize the results of our work into Table~\ref{tab:perf-sum}.

\begin{table*}[!ht]
\centering
\small
 \caption{ Summarization of our proposed work }
 \label{tab:perf-sum}
 \begin{tabularx}{\textwidth}{l|X|X}
 \toprule
 Proposed Methods  &Major Contribution& Results\\
 \toprule
Feature Word Embedding & Encode information from multiple levels& Better representation of entity-related words; improve NER\\
\hline
Label Embedding of Entities& Encode the concept-level dependency using prototypes; combine the force of word-level and concept-level information.& Improve FNET at both few-shots and zero-shot setting.\\
\hline
RBM-based reranking model& Incorporate the concept-level prior knowledge into the binary embedding layer of RBM.&Improve ASR reranking algorithm using concept-level prior.\\
\hline
Sentic-LSTM & Utilize commonsense knowledge for targeted sentiment analysis; propose an effective extension of LSTM to incorporate the concept-level and word-level inputs.& Improve both aspect categorization and polarity detection of targeted sentiment analysis \\
\bottomrule 
\end{tabularx}
\end{table*}

At last, the results presented in this thesis suggests that the concept-level knowledge is generally helpful as long as concepts (including named entities) are playing a key roles in the task. On the other hand, as concepts are typically of high dimensionality, the embedding layer together with efficient computational structure could make the concept-level information more accessible.

\section{Future Work}

We would discuss possible directions we like to explore in the future. 
\subsection{Limitation of Our Work}

One limit of our solution to leveraging concepts as well as fusing concept-level and word-level information is that our usage of concepts is constrained to the original form of input data. In other words, the use of concepts does not change the form of inputs that are mostly fed into the model as a sequence of words. However, when considering the interaction between concepts, keeping to the original sequential order of words might become redundant or sometimes might even cause unnecessary confusion. For example, for a sentence 'this place is safe, but areas close to it are very dangerous', if the target is to analyze the sentiment towards the first place, the second clause might be cause confusion if being processed in its original order. In contrast, parsing the sentence into concepts and relations is able to shorten the sentence into a simple tuple (place,be,safe) which could not only help remove the redundant information but also create the dependency between a target and a correct sentiment-related modifier. Inspired by previous work reforming the sequential input into structures such as trees~\cite{treelstm} or graphs~\cite{graphlstm}, we would like to investigate exploiting non-sequential structure to better perform concept-level analysis of natural language. 

\subsection{Graph-based Concept Embedding}

Concepts are naturally organized in graphs. For example, ConceptNet defines a set of relations such as Isa relation between concepts. A graph can be built with concept being the nodes and relations being the edges. Similarly, AffectiveSpace can also be converted into a directed graph, even though it uses affective properties (relation + argument). So far, the concept embeddings~\cite{camaf2} are built by merely considering decomposition of the concept-property matrix. In other words, the result embedding of concepts are only affected by immediate neighbors. 

Recent work on embedding graphs~\cite{hamilton2017representation} has shed light upon propagating information over a graph so that the embedding of a node is able to also reflect the properties of its indirect neighbors. We could try to infer concept embeddings via the graph structure. Different from existing work, our method will also consider the relation types which are also important for modeling information propagation in AffectiveSpace.

\subsection{Concept-oriented Graph Model}
Since concepts are very informative and their interaction is key to properly understanding the meaning of natural language, our first proposal for future work is to extend the usage of concepts to reforming the input sequence. One way to do so is to transform the input to local graphs of concepts by considering their explicit or implicit relations. Explicit relations can be the semantic relation given by knowledge bases such as ConceptNet or SenticNet. Implicit relations, for example, can be the presence of concepts in each other's contexts. By converting the word sequence into concept graphs, the NLP model can explicitly avoid affected by redundant and misleading structures and take into account the interaction between concepts. 

In addition, another extension is to consider the global interaction of concepts. This can be fairly critical tor tasks such as sentiment analysis of geographical entities~\cite{saeidi:2016}~\footnote{Similar to named entities, we could also consider geographical entities as a type of concepts.}. For example, people might share the same safety concerns with neighboring locations. A graph built to represent the relation of concepts allows us to perform collective inference via approaches such as label propagation~\cite{zhu2002learning}.

\subsection{New Sentiment Analysis Application using Concepts}

In future, we also would like to explore new sentiment-related applications using concepts. It is known that sentiment is playing an important role in expressing opinions. In addition to decoding (or interpreting) people's opinion, it is also of our interest to generate sentiment-aware responses. Especially, in the context of e-commerce, automatically generating sentiment-aware responses to customers' reviews would largely improve the customer experience. Moreover, rather than generating a sequence of words, we are more interested in generating a mix of words and commonsense concepts. We believe that, by leveraging commonsense concepts as well as concept relations, it could help improve the consistency and comprehensiveness of generated responses.







\chapter*{Appendix A}
\section*{Authors Publications}
\begin{itemize}
\item \textbf{Yukun Ma}, Jung-jae Kim, Benjamin Bigot, and Tahir Muhammad Khan. ``Feature-enriched word embeddings for named entity recognition in open-domain conversations." In Acoustics, Speech and Signal Processing (ICASSP), 2016 IEEE International Conference on, pp. 6055-6059. IEEE, 2016.

\item Khan, Muhammad Tahir, \textbf{Yukun Ma}, and Jung-jae Kim.``Term Ranker: A Graph-Based Re-Ranking Approach." In FLAIRS Conference, pp. 310-315. 2016.

\item \textbf{Yukun Ma}, Erik Cambria, and Sa Gao. ``Label Embedding for Zero-shot Fine-grained Named Entity Typing." In COLING, pp. 171-180. 2016..

\item \textbf{Yukun Ma}, Erik Cambria, and Benjamin Bigot. ``ASR Hypothesis Reranking using Prior-informed Restricted Boltzmann Machine." In International Conference on
Computational Linguistics and Intelligent Text Processing (CiCLing), 2017

\item Sa Gao, Zhenchang Xing, \textbf{Yukun Ma}, Deheng Ye and Shang-Wei Lin, ``Enhancing Knowledge Sharing in Stack Overflow via Automatic External Web Resources Linking", International Conference on Engineering of Complex Computer Systems (ICECCS), Fukuoka, Japan, 2017.

\item \textbf{Yukun Ma}, Haiyun Peng, and Erik Cambria. ``Targeted Aspect-Based Sentiment Analysis via Embedding Commonsense Knowledge into an Attentive LSTM", In the Thirty-Second AAAI Conference on Artificial Intelligence (AAAI-18),pp. 5876-5883, 2018 
\item Haiyun Peng, \textbf{Yukun Ma}, Erik Cambria, Yang Li, ``Learning Multi-grained Aspect Target Sequence
for Chinese Sentiment Analysis", Vol. 1, pp.167-176, Knowledge-based Systems.

\end{itemize}

\bibliographystyle{IEEEtran}
\bibliography{reference}

%

\newpage
\addcontentsline{toc}{section}{\numberline{}\hspace{-.35in}{\bf
Author's Publications}}

\end{document}